\documentclass[preprint,12pt]{elsarticle}

\usepackage{amssymb}
\usepackage{amsmath}
\usepackage{bm}
\usepackage[flushleft]{threeparttable}
\usepackage{multirow}
\usepackage{array}

\usepackage{graphicx}%
\usepackage{multirow}%
\usepackage{amsmath,amssymb,amsfonts}%
\usepackage{amsthm}%
\usepackage{mathrsfs}%
\usepackage{xcolor}%
\usepackage{textcomp}%
\usepackage{manyfoot}%
\usepackage{booktabs}%
\usepackage{algorithm}%
\usepackage{algorithmicx}%
\usepackage{algpseudocode}%
\usepackage{listings}%

\begin{document}
\begin{frontmatter}
\title{Investigation of cardinality classification for bacterial colony counting using explainable artificial intelligence} 

\author[kcl-label]{Minghua Zheng\fnref{fn1}\corref{cor}}
\ead{minghua.zheng@kcl.ac.uk}
\author[uh-label]{Na Helian}
\author[uh-label]{Peter C. R. Lane}
\author[uh-label]{Yi Sun}
\author[synoptics-label]{Allen Donald}

\cortext[cor]{Corresponding author}
\fntext[fn1]{First author}

\affiliation[kcl-label]{organization={King's College London},
            addressline={St Thomas Street}, 
            city={London},
            postcode={SE1 1UL}, 
            country={United Kingdom}}

\affiliation[uh-label]{organization={University of Hertfordshire},
            addressline={College Lane}, 
            city={Hatfield},
            postcode={AL10 9AB}, 
            country={United Kingdom}}

\affiliation[synoptics-label]{organization={Synoptics Ltd},
            addressline={Beacon House, Nuffield Road}, 
            city={Cambridge},
            postcode={CB4 1TF}, 
            country={United Kingdom}}

\begin{abstract}
Automatic bacterial colony counting is a highly sought-after technology in modern biological laboratories because it eliminates manual counting effort. Previous work has observed that MicrobiaNet, currently the best-performing cardinality classification model for colony counting, has difficulty distinguishing colonies of three or more individuals. However, it is unclear if this is due to properties of the data together with inherent characteristics of the MicrobiaNet model. By analysing MicrobiaNet with explainable artificial intelligence (XAI), we demonstrate that XAI can provide insights into how data properties constrain cardinality classification performance in colony counting. Our results show that high visual similarity across classes is the key issue hindering further performance improvement, revising prior assertions about MicrobiaNet. These findings suggest future work should focus on models that explicitly incorporate visual similarity or explore density estimation approaches, with broader implications for neural network classifiers trained on imbalanced datasets.
\end{abstract}

\end{frontmatter}

\section{Introduction}
\label{sec-introduction}
Bacterial colony counting is widely used in biological laboratories to estimate the number of viable bacteria present in a test sample. The counting result is an important indicator of the cleanliness of a surface, the sterility of a product, or the presence of a bacterial infection. With the availability of large datasets and powerful computational resources, many researchers have attempted to apply artificial intelligence (AI) to count bacterial colonies.

Among these efforts, \citet{Ferrari2015, Ferrari2017} reformulated the counting task as a classification problem and proposed a novel deep learning-based algorithm, referred to as MicrobiaNet in this work for simplicity. To the best of our knowledge, MicrobiaNet currently represents the most effective cardinality classification model for colony counting. It accepts images with an undetermined number of colonies and assigns each image to a pre-defined class indicating the number of colonies (cardinality). These images are usually obtained by cropping regions from culture plates on which colonies have been grown after performing an upstream object detection step. The final colony count for a plate is calculated by adding up the predicted cardinalities of all images cropped from the plate.

Despite the successful application of deep learning on bacterial colony counting, MicrobiaNet has difficulty distinguishing colonies of three or more individuals~\citep{Ferrari2017}. It is unknown if this is caused by properties of the dataset together with inherent characteristics of the MicrobiaNet model. The dataset is characterised by high visual similarity across classes and an imbalanced class distribution. The former denotes that colonies of different cardinalities often share overlapping shapes, densities, and spatial arrangements. For example, a colony of three closely packed cells may appear almost identical to a colony of four, as the additional cell does not substantially alter the cluster’s overall appearance. The latter is often considered a contributor to misclassification. 

This work attempts to investigate MicrobiaNet using explainable artificial intelligence (XAI), aiming to identify factors that could inform strategies for further performance improvement. This is motivated by the success of XAI in uncovering the internal logic underlying algorithmic decisions. Additionally, the investigation aims to produce potential wider impacts on applications of convolutional neural network (CNN) classifiers trained on imbalanced datasets, with MicrobiaNet serving as a representative example.

Our contributions are twofold. First, we show that XAI can provide insights into how data properties constrain cardinality classification performance in colony counting, highlighting its potential as a general diagnostic tool beyond post hoc interpretability. Second, by applying XAI-driven analysis to MicrobiaNet, we empirically identify that the dominant limitation arises from high visual similarity among classes rather than class imbalance, which is further influenced by the model’s inherent properties. The latter contribution revises the assertion made by MicrobiaNet’s authors that mitigating class imbalance would lead to improved results.

The rest of this paper is organised as follows. Related work on bacterial colony counting is discussed in Section~\ref{sec-related-work}. Section~\ref{subsec-data} describes the data used in this study. Methods are introduced and explained in Section~\ref{sec-methods}. Section~\ref{sec-exp} presents experiments and results. Conclusions are provided in Section~\ref{sec-conclusions}. Finally, limitations and future work are discussed in Section~\ref{sec-limitations-future-work}.

\section{Related work}
\label{sec-related-work}
Automated bacterial colony counting has been an active research area for many years since the first attempt for automation made by~\citet{Mansberg1957}. Recent advances in AI have encouraged researchers to adopt machine learning techniques over traditional image processing methods for this task. This is because machine learning algorithms can learn from data without being explicitly programmed. Such capacity has enabled them to address complex tasks in computer vision~\citep{Simonyan2014}, natural language processing~\citep{collobert_natural_2011}, drug design~\citep{beck_predicting_2020}, etc. 

Machine learning algorithms are preferred for automation in bacterial colony counting. Traditional image processing methods, such as edge detection~\citep{Barber2001, Loukas2003, Choudhry2016} and contour detection~\citep{Niyazi2007, Wienert2012}, rely heavily on manually tuned parameters. Other commonly used techniques, including hough transform~\citep{DanaH.Ballard1981, Flaccavento2011, Geissmann2013, Matic2016} and threshholding~\citep{Zhang2007, Ates2009, Chen2009, Smith2014, Chiang2015, Khan2018}, similarly require users to manually adjust parameters to accommodate to the change of background colour, colony size, colony shape, colony species, etc. As a result, these traditional approaches are less suitable for fully automated and robust colony counting systems.

Despite the increasing adoption of machine learning algorithms, their black-box nature prevents people from understanding how these algorithms make decisions. This section reviews machine learning algorithms for bacterial colony counting and highlights the absence of studies applying XAI to this task.

\subsection{Machine learning algorithms for bacterial colony counting}
The design of machine learning algorithms for bacterial colony counting follows the detect-then-count strategy. The first step is to detect all bacterial colonies from the image. Then, the number of detections are added up to represent the total number of colonies in the image. A difficult challenge for detect-then-count approaches is that bacterial colonies are often overlapped and clustered, which increases the counting error.

Advances in general object detection algorithms have led researchers to investigate their applicability to bacterial colony counting. For example, \citet{majchrowska_deep_2021} conducted a benchmark study on bacterial colony counting using Faster R-CNN~\citep{Ren2017}, Cascade R-CNN~\citep{cai_cascade_2018}, Libra R-CNN~\citep{pang_libra_2019}, CBNetV2~\citep{liang_cbnet_2022}, YOLOv4~\citep{bochkovskiy_yolov4_2020}, EfficientDet-D2~\citep{tan_efficientdet_2020}, and Deformable DETR~\citep{zhu_deformable_2021}. However, none of these methods tackle clustered colonies or provide algorithmic explainability. \citet{mohammadkhan_automated_2021} also explored the use of Faster R-CNN for bacterial colony counting. Nevertheless, no attention was paid to clustered colonies and model explainability.

Several studies have focused on handling clustered colonies during the development of bacterial colony counting algorithms. For example,~\citet{Ferrari2015, Ferrari2017} proposed a support vector machine and a CNN to predict the number of colonies in an input image which may contain zero to seven colonies. This number is referred to as colony cardinality. The final colony count is calculated by adding up each predicted cardinality. Their best algorithm, a CNN referred to as MicrobiaNet in this work, has achieved 92.1\% accuracy and an F1 score of 0.81 on the dataset they created. However, their experimental results demonstrate that MicrobiaNet has difficulty distinguishing colonies of three or more individuals. 

The underlying cause of this limitation remains unclear. It may be attributed to properties of the dataset together with inherent characteristics of the MicrobiaNet model. The dataset is characterised by high visual similarity across classes and an imbalanced class distribution. While class imbalance is commonly considered a driver of misclassification, its influence relative to high visual similarity across classes is still uncertain. In addition, the explainability of MicrobiaNet has not yet been explored. Despite that, MicrobiaNet is chosen over algorithms investigated by~\citet{majchrowska_agar_2021, majchrowska_deep_2021} in this work due to its success in addressing clustered colonies.

\subsection{Explainable artificial intelligence for bacterial colony counting}
Explainable Artificial Intelligence (XAI) has gained increasing attention from the research community in recent years. It is a set of processes and methods that allows people to comprehend and trust the decisions made by machine learning algorithms. The growing scholarly and practical attention in this area is largely driven by societal, ethical~\citep{muller_ten_2021}, and legislative~\citep{schneeberger_european_2020} pressures. These pressures call for greater transparency and accountability in algorithmic systems. In particular, algorithm developers are increasingly expected to elucidate the internal logic underlying algorithmic decisions, especially in critical domains such as healthcare, finance, and security. 

According to the survey by~\citet{minh_explainable_2022}, XAI can be classified into three main categories: pre-modelling explainability, interpretable models, and post-modelling explainability. The first category includes methods that focus on data rather than the machine learning model itself. These methods emphasise data pre-processing and feature exploration to enhance users' understanding of the data used for model training. The second category encompasses machine learning algorithms that are inherently interpretable due to their simple internal structures. For example, linear regression, logistic regression, decision trees, and $k$-nearest neighbours have a very simple structure to explain the decision-making process. The third category comprises methods applied after model training to clarify how the model generates predictions for given inputs through visualisation techniques.

Despite the growing interest in XAI, its use in investigating machine learning algorithms for bacterial colony counting remains largely unexplored. The lack of attention to bacterial colony counting may be due to its perceived lower criticality compared to other healthcare applications, such as cancer prediction or tumour detection. However, it is problematic to overlook explainability of machine learning algorithms used for bacterial colony counting. This is because the count result is essential to inform the development of medicines and chemical products which hold significant scientific and industrial importance. Among the three XAI categories, this study employs post-modelling explainability methods to investigate MicrobiaNet with the goal of identifying factors that could inform strategies to overcome its limitation. This decision is motivated by the direct investigation into MicrobiaNet's decision-marking without altering the architecture.

\section{Data description}
\label{subsec-data}
A collection of 28418 segments created by~\citet{Ferrari2017} is used to investigate colony cardinality classification since it is the only labelled dataset to facilitate the investigation.\footnote{The original download URL is no longer available now, but a snapshot is available at https://web.archive.org/web/20220401001815/https://microbia.org/index.php/resources.} It is referred to as Microbia Dataset in this work for simplicity. A segment represents a colony cluster that may have multiple colonies. Each segment was cropped from a plate image into a square image whose dimension was determined by either the longer height or the longer width of the colony cluster, followed by a border extension of a fixed size and padding.

\begin{figure}[h!]
    \centering
    \includegraphics[width=.85\linewidth]{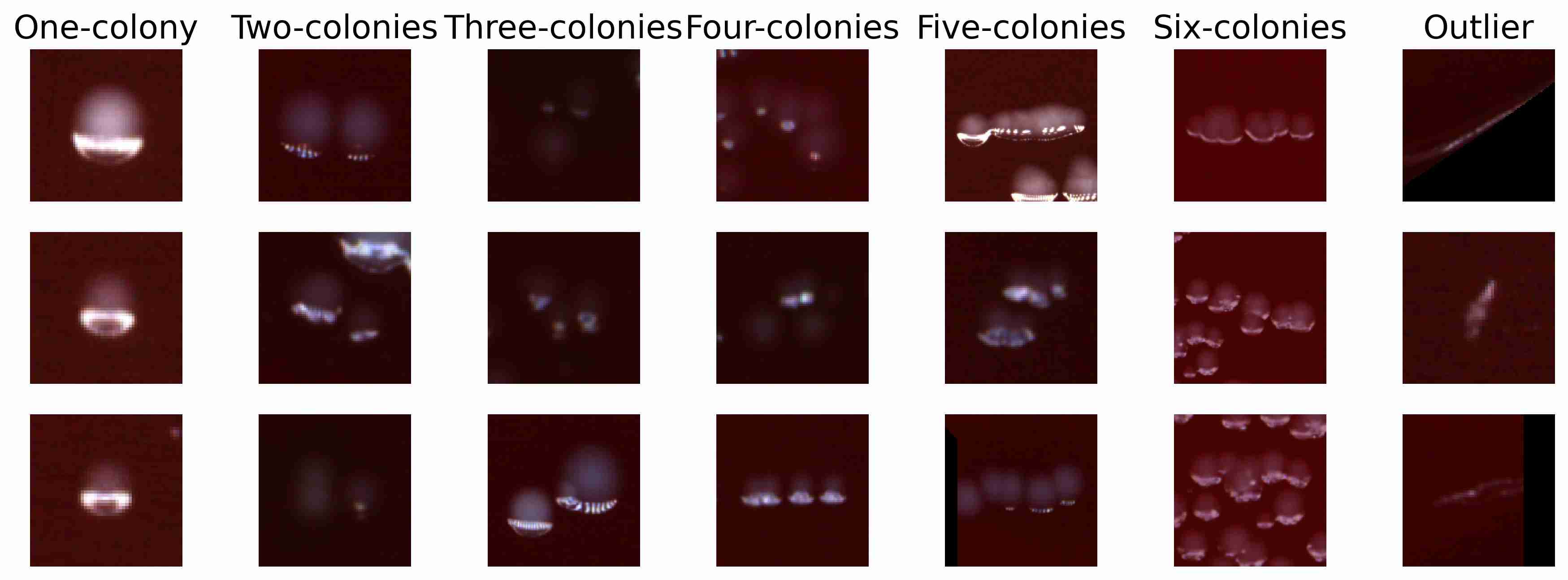}
    \caption{Example segments from seven different classes. Some segments, such as the first Five-colonies segment and the second Six-colonies segment, contain neighbouring colonies that may disrupt colony counting algorithms.}
    \label{fig:microbia_segments}
\end{figure}

The cropped segment was manually assigned a label by two experts to indicate the number of colonies in the cluster. Seven class labels are provided in Microbia Dataset: Outlier class, One-colony class, Two-colonies class, Three-colonies class, Four-colonies class, Five-colonies class, and Six-colonies class. Class names are treated as proper labels and are therefore capitalised (e.g. ``Two-colonies class''). When referring to individual samples, the same labels are used adjectivally (e.g. ``a Two-colonies image''). Among them, Outlier class means the segment is bubble, dust, or dirt, which has zero colony. Figure \ref{fig:microbia_segments} shows three example segments per class, illustrating that neighbouring colonies in the first Five-colonies segment and the second Six-colonies segment may disrupt colony counting algorithms. Moreover, the class distribution of Microbia Dataset is imbalanced as detailed in Table \ref{tab:class_distribution}.

\begin{table}[h!]
\centering
\caption{Class distribution in Microbia Dataset.}
\label{tab:class_distribution} 
\begin{tabular}{lcc}
\hline\noalign{\smallskip}
Class          & Count & Percentage (\%)  \\
\noalign{\smallskip}\hline\noalign{\smallskip}
        One-colony     & 14285 & 50.27    \\
        Two-colonies   & 5443  & 19.15    \\
        Three-colonies & 3634  & 12.79    \\
        Four-colonies  & 1836  & 6.46     \\
        Five-colonies  & 953   & 3.35     \\
        Six-colonies   & 1006  & 3.54     \\
        Outlier        & 1261  & 4.44     \\
\noalign{\smallskip}\hline
\end{tabular}
\end{table}

\begin{figure}[h!]
    \centering
    \includegraphics[width=.85\linewidth]{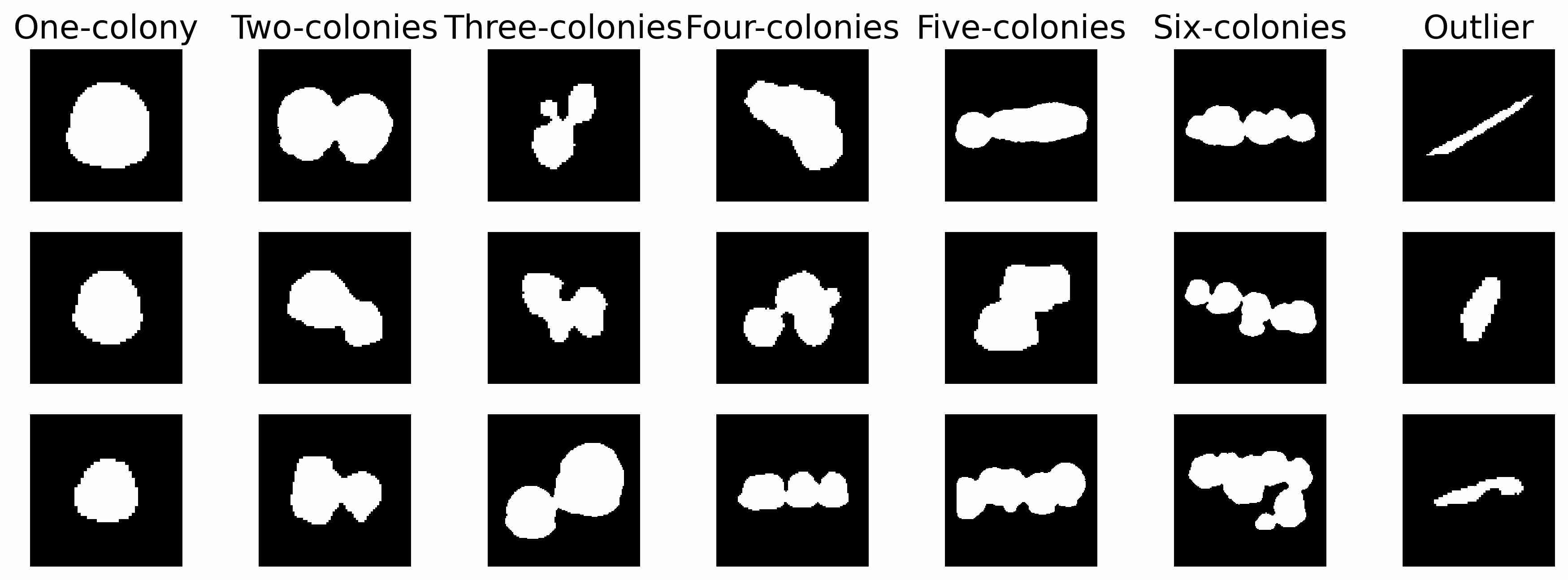}
    \caption{Masks for the segments in Figure \ref{fig:microbia_segments}.}
    \label{fig:microbia_masks}
\end{figure}
\begin{figure}[h!]
    \centering
    \includegraphics[width=.85\linewidth]{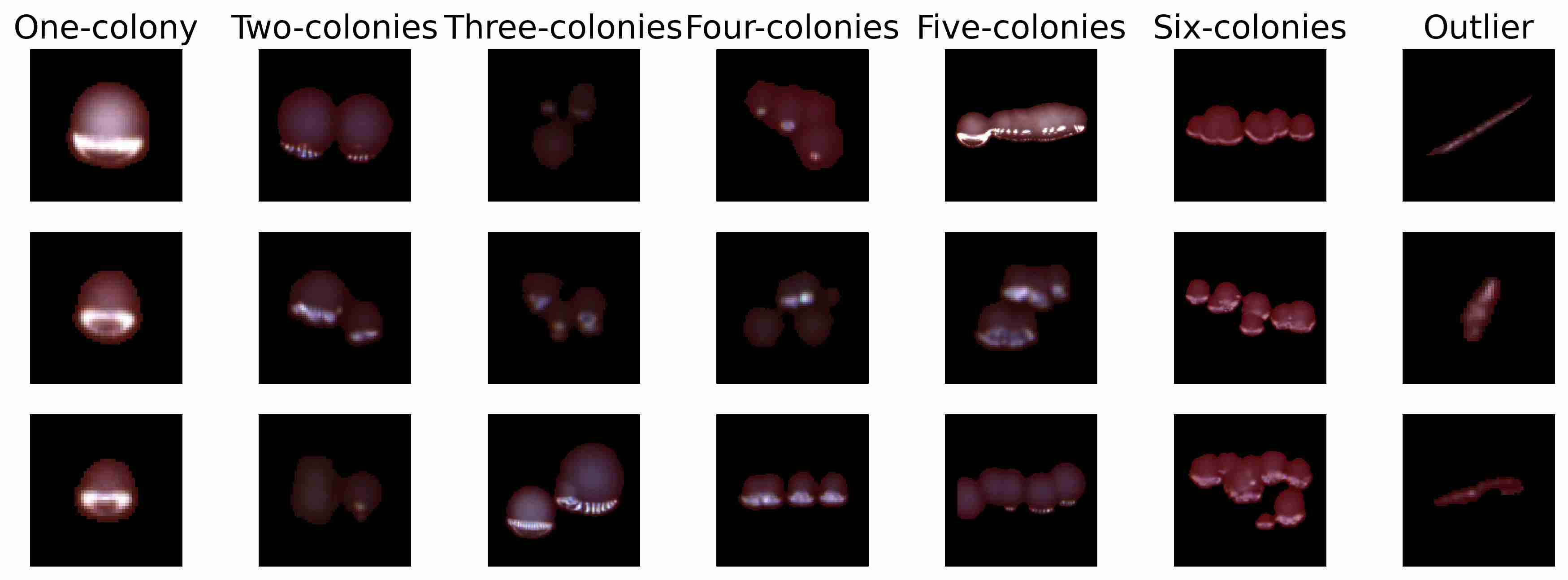}
    \caption{Masked segments generated based on Figure \ref{fig:microbia_segments} and Figure \ref{fig:microbia_masks}. Each colony image is now excluded from interference caused by neighbouring colonies.}
    \label{fig:microbia_masked_segments}
\end{figure}

In addition to providing labelled segments, Microbia Dataset includes 28418 corresponding masks for the cropped segments, enabling users to mitigate interference from neighbouring colonies. This can be accomplished by applying a bitwise AND operation on the segment with the mask. In other words, the pixel in the segment is assigned a value of zero if its corresponding pixel in the mask has a value of zero, and the pixel in the segment is unchanged if its corresponding pixel in the mask has a value that is greater than zero. The masks and masked segments of these seven classes corresponding to the segments in Figure \ref{fig:microbia_segments} are demonstrated in Figures \ref{fig:microbia_masks} and \ref{fig:microbia_masked_segments}.

Figure \ref{fig:microbia_masked_segments} shows that the first Five-colonies segment and the second Six-colonies segment become cleaner after removing interference from neighbouring colonies. It can also be observed that images containing three, four, five, and six colonies are visually similar and difficult to distinguish. In this work, this characteristic is referred to as high visual similarity across classes. Masked segments are used in this work to minimise interference from adjacent colonies that could otherwise disrupt analysis.

\section{Methods}
\label{sec-methods}
\subsection{The colony cardinality classification model}
\label{model}
\begin{figure}[h!]
    \centering
    \includegraphics[width=.85\linewidth]{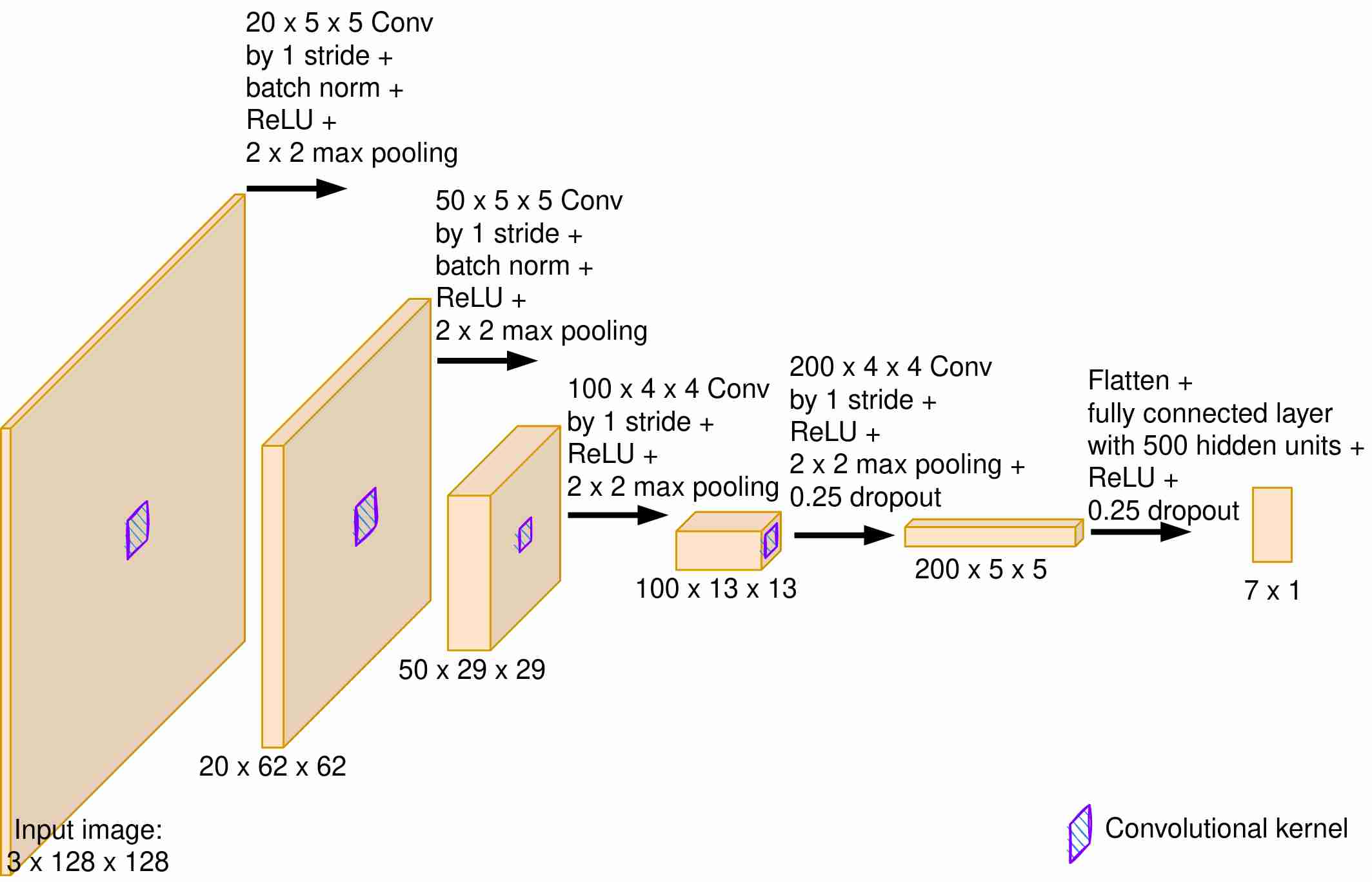}
   \caption{MicrobiaNet architecture.}
    \label{fig:microbia_network_architecture}
\end{figure}

MicrobiaNet is selected in this study to investigate its explainability, aiming to identify factors that could inform strategies for further performance improvement with XAI. To the best of our knowledge, MicrobiaNet currently represents the best-performing cardinality classification algorithm for colony counting. As illustrated in Figure \ref{fig:microbia_network_architecture}, MicrobiaNet consists of four convolutional layers followed by a fully connected layer. The input images have a size of $3 \times 128 \times 128$. The convolutional layers use kernel parameters (the number of kernels~$\times$~height~$\times$~width) of $20 \times 5 \times 5$, $50 \times 5 \times 5$, $100 \times 4 \times 4$, and $200 \times 4 \times 4$, respectively. All convolutional and fully connected layers employ Rectified Linear Unit (ReLU) activations~\citep{agarap_deep_2018}. The output of the first two convolutional layers is followed by batch normalisation~\citep{ioffe_batch_2015} and $2 \times 2$ max pooling.\footnote{The original normalisation method is local response normalisation attached to ReLU activation. It is replaced with batch normalisation followed by ReLU activation as it produces a better result based on our experimental results.}

For the remaining layers, the output of the last two convolutional layers is followed by $2 \times 2$ max pooling without batch normalisation. The final convolutional output is flattened and passed into a fully connected layer containing 500 hidden units. To mitigate overfitting, a dropout rate of 25\% is applied both before and after the fully connected layer.\footnote{The original dropout rate is 75\% as described by~\citet{Ferrari2017}. Based on our experimental results, 25\% is the dropout rate that reproduces a similar performance.} The model outputs seven class scores, each representing the likelihood that the input image belongs to one of the seven classes.

\subsection{Explainable artificial intelligence}
\label{subsec-xai}
This section introduces three complementary XAI approaches that focus on post-model explainability: network layer output visualisation, feature visualisation, and class activation map visualisation. These methods work together to provide a holistic view of the model’s behaviour and to investigate the difficulty MicrobiaNet has in distinguishing colonies comprising three or more individuals. Network layer output visualisation shows how information flows through the network. Feature visualisation reveals the patterns that activate neurons. Class activation map visualisation highlights which input regions drive the model’s predictions.

\subsubsection{Visualisation of network layer outputs}
\label{subsubsec-network-layer-out-vis}
Network layer output visualisation is commonly used to examine the internal information flows that drive a neural network’s decisions. The output of the final fully connected layer is typically visualised, as it contains the features the network uses for decision-making. Because these features are high-dimensional, dimensionality reduction techniques are applied to project them into a two-dimensional space suitable for plotting. These techniques are Principal Component Analysis (PCA) and t-distributed Stochastic Neighbour Embedding (t-SNE)~\citep{maaten_visualizing_2008}.

PCA is used to reduce dimensionality while preserving the directions of maximum variance in the data. It identifies a set of orthogonal vectors, i.e. principal components, that best match the description of the spread and direction of data across multiple dimensions. The top-$n$ best-describing principal components are subsequently selected to reduce the dimensionality of feature space in the dataset. The top-$2$ important principal components are used in this work to allow two-dimensional visualisation.

In addition to PCA, t-SNE is used to visualise high-dimensional neural network outputs, as it is particularly effective for visualising data that are not linearly separable. It remaps each data point in the high-dimensional space to a low-dimensional space where each point's nearby points are similar and its distant points are dissimilar. This is achieved by converting similarities between data points to joint probabilities, followed by minimising the Kullback-Leibler divergence between the joint probabilities of the mapped low-dimensional data and the high-dimensional data.

Perplexity is a hyper-parameter in t-SNE used to determine the number of close neighbours each point has, balancing the attention between regional and global data. According to t-SNE's authors, typical values for perplexity are between 5 and 50. As suggested by~\citet{wattenberg2016how}, a perplexity of 30 and 5000 iterations are used in this work. Additionally, the learning rate used to generate t-SNE plots is calculated by a formula suggested by~\citet{pedregosa_scikit-learn_2011}, defined as the greater of 50 and the dataset size divided by 48.

\subsubsection{Visualisation of learned features}
Feature visualisation is used to understand the patterns a neural network has learned by generating synthetic images that maximise neuron activations. Here, it focuses on visualising MicrobiaNet’s convolutional kernels to identify the input features that elicit the highest activations.

Visualisation is performed by optimisation. A $3 \times 128 \times 128$ image, matching MicrobiaNet’s input size, is initialised with random noise. It is input to the trained MicrobiaNet to reach the target convolutional kernel whose derivatives are back-propagated to the input image iteratively until it convergences. Following the approach of~\citet{olah_feature_2017}, eight images per kernel are generated over 512 iterations with a learning rate of 0.05. To simplify analysis, only the first and second kernels of each convolutional layer are visualised.

\subsubsection{Visualisation of class activation maps}
Class activation map (CAM) visualisation identifies the image regions most responsible for a model’s prediction~\citep{zhou_learning_2015}. CAM inserts a global average pooling layer between the last convolutional layer and the fully connected layer. The weights of the final network output layer with respect to a specific class are projected back onto each convolutional feature map to highlight influential image regions. However, because CAM requires architectural changes, it has largely been superseded by Gradient-weighted CAM (Grad-CAM)~\citep{selvaraju_gradcam_2017}.

Grad-CAM computes the gradient of a class score with respect to the final convolutional layer, followed by an average pooling operation to obtain the neuron importance weights. These weights are multiplied by the feature maps in the target convolutional layer, followed by a ReLU operation to keep positive influence. The last convolutional layer is considered here as it contains the richest semantic information.

Several other variants provide complementary insights. GradCAM++ uses the second order gradients~\citep{chattopadhay_gradcam_2018}. High-Resolution CAM (HiResCAM) multiplies gradients element-wise by the feature maps in the target convolutional layer~\citep{draelos_use_2021}. Axiom-based Grad-CAM (XGrad-CAM) normalises the feature maps before multiplying them by the gradients~\citep{fu_axiombased_2020}. Eigen-CAM focuses on the first principal component of the feature maps~\citep{muhammad_eigencam_2020}. Eigen Gradient-weighted CAM (EigenGrad-CAM) multiplies the first principal component of the feature maps by the gradients computed from the specified class score with respect to the convolutional layer.

In this work, activation maps for the true class of MicrobiaNet are visualised using randomly selected training images. The visualisations are implemented via an open-source library developed by~\citet{jacobgilpytorchcam}.

\section{Experiments}
\label{sec-exp}
The investigation of MicrobiaNet with XAI focuses on post-model explainability to directly examine its decision-making without altering the architecture. A baseline performance is first established, which serves as a reference point for subsequent analyses with XAI. This is followed by experiments assessing the impact of class imbalance and exploring whether XAI insights can improve counting performance. Overall, the experiments are organised into four case studies.

\subsection{Colony cardinality classification baseline performance}
\label{sec_colony_card_cls_baseline}
This case study establishes a baseline assessment of colony cardinality classification with MicrobiaNet, providing a reference for subsequent XAI analyses.

\subsubsection{Experimental setup}
\label{subsubsec_training_and_eval}
Microbia Dataset is stratified into training set, validation set, and test set with a ratio of 6:2:2.\footnote{The original Microbia Dataset does not contain fixed training set, validation set, and test set.} This ensures the class distribution remains the same in different sets to meet different needs. As listed in Table \ref{class_distribution_microbia_split}, training set, validation set, and test set have 17050, 5684, and 5684 images, respectively. Alternative split ratios, such as 8:1:1 and 7:2:1, are not used because of the relatively large dataset size of 28418. Microbia Dataset is randomly shuffled with five different seeds before splitting, enabling evaluation on diverse data and improving reliability. To avoid confusion, the five Microbia dataset variants, split with different random seeds, are called MicrobiaS1, MicrobiaS2, MicrobiaS3, MicrobiaS4, and MicrobiaS5, where S stands for the seed.

\begin{table}[h!]
       \centering
       \caption{Class distribution of the Microbia training, validation, and test sets.}
       \label{class_distribution_microbia_split}
       \begin{tabular}{p{0.25\textwidth}>{\centering}p{0.12\textwidth}>{\centering}p{0.12\textwidth}>{\centering}p{0.12\textwidth}>{\centering\arraybackslash}p{0.15\textwidth}}
               \hline
               \multirow{2}{*}{Class} & \multicolumn{3}{c}{Count} & \multirow{2}{*}{Percentage (\%)}                \\\cmidrule{2-4}
                                      & Training                  & Validation                       & Test &       \\ \hline
               One-colony             & 8571                      & 2857                             & 2857 & 50.27 \\
               Two-colonies           & 3265                      & 1089                             & 1089 & 19.15 \\
               Three-colonies         & 2180                      & 727                              & 727  & 12.79 \\
               Four-colonies          & 1102                      & 367                              & 367  & 6.46  \\
               Five-colonies          & 571                       & 191                              & 191  & 3.35  \\
               Six-colonies           & 604                       & 201                              & 201  & 3.54  \\
               Outlier                & 757                       & 252                              & 252  & 4.44  \\ \hline
       \end{tabular}
\end{table}

MicrobiaNet is trained and validated separately on each of the MicrobiaS1-S5 datasets, resulting in five independent training-evaluation cycles. This approach ensures more reliable evaluation than a single run and resembles $5$-fold cross-validation, which cannot be directly applied in XAI experiments. Validation results establish baseline performance, while the test set is withheld to prevent the model from being overly optimised. Training images are normalised using the mean and standard deviation of the training set RGB values, which are also applied to validation images. This accelerates convergence during model training~\citep{Krizhevsky2012, He2015}.

MicrobiaNet parameters are initialised with Kaiming initialisation~\citep{He2015} and optimised using Adaptive Moment Estimation (Adam)~\citep{kingma_adam_2014} (learning rate $10^{-2}$, batch size 64, 500 epochs, weight decay of $5 \times 10^{-4}$). These hyper-parameters match those used by~\citet{Ferrari2017}. Training is early terminated to avoid unnecessary computation if the validation F1 score does not improve by 1\% over 100 consecutive epochs. The F1 score is a weighted average used to mitigate the influence of class imbalance. Experiments are implemented in PyTorch~\citep{paszke_pytorch_2019} on a system with an AMD Ryzen Threadripper 3990X CPU, 2 Nvidia RTX 3090 GPUs, and 128GB RAM.

\subsubsection{Results}
The results presented in this work primarily focus on F1 score, which mitigates accuracy’s limitations on imbalanced class distributions by balancing precision and recall. Accuracy, precision, and recall are also reported for completeness and comparison with prior work.

\begin{table}[h!]
       \centering
       \caption{Overall evaluation results on the MicrobiaS1, MicrobiaS2, MicrobiaS3, MicrobiaS4, and MicrobiaS5 datasets.}
       \label{Microbia_model_results_with_diff_data_split}
       \begin{tabular}{p{0.01\textwidth}>{\centering}p{0.18\textwidth}>{\centering}p{0.05\textwidth}>{\centering}p{0.05\textwidth}>{\centering}p{0.05\textwidth}>{\centering}p{0.05\textwidth}>{\centering}p{0.05\textwidth}>{\centering}p{0.05\textwidth}>{\centering\arraybackslash}p{0.09\textwidth}}
               \hline
                                                                     & \multirow{2}{*}{Metric} & \multicolumn{5}{c}{Split with Different Seeds} & \multirow{2}{*}{Mean} & \multirow{2}{*}{Std}                                                 \\ \cmidrule{3-7}
                                                                     &                         & S1                                                           & S2                    & S3                        & S4        & S5        &           &           \\ \hline
               \multirow{4}{*}{\rotatebox[origin=c]{90}{Training}}   & Precision               & 0.85                                                         & 0.85                  & 0.84                      & 0.85      & 0.85      & 0.85      & 0.0010      \\
                                                                     & Recall                  & 0.86                                                         & 0.85                  & 0.85                      & 0.85      & 0.86      & 0.85      & 0.0039      \\
                                                                     & F1 score                & \bf{0.85}                                                    & \bf{0.85}             & \bf{0.84}                 & \bf{0.85} & \bf{0.85} & \bf{0.85} & \bf{0.0030} \\
                                                                     & Accuracy (\%)           & 85.56                                                        & 85.18                 & 84.63                     & 84.82     & 85.62     & 85.16     & 0.3900      \\ \cmidrule{1-9}
               \multirow{4}{*}{\rotatebox[origin=c]{90}{Validation}} & Precision               & 0.82                                                         & 0.83                  & 0.83                      & 0.83      & 0.81      & 0.82      & 0.0080      \\
                                                                     & Recall                  & 0.83                                                         & 0.84                  & 0.83                      & 0.83      & 0.82      & 0.83      & 0.0055      \\
                                                                     & F1 score                & \bf{0.82}                                                    & \bf{0.83}             & \bf{0.83}                 & \bf{0.83} & \bf{0.81} & \bf{0.82} & \bf{0.0063} \\
                                                                     & Accuracy (\%)           & 82.79                                                        & 83.52                 & 83.27                     & 83.39     & 82.00     & 82.99     & 0.5500      \\ \hline
       \end{tabular}
\end{table}

The overall classification performance of MicrobiaNet on the MicrobiaS1-S5 datasets is presented in Table \ref{Microbia_model_results_with_diff_data_split}.\footnote{Unless otherwise stated, the reported precision, recall, and F1 score in the overall classification results are weighted averages.} Across five data splits, MicrobiaNet achieves an average training F1 of 0.85 and validation F1 of 0.82, with very low standard deviations (0.0030 and 0.0063, respectively), demonstrating highly consistent and stable performance. The gap between training and validation F1 scores never exceeds 0.03, indicating strong generalisation to unseen data.

\begin{figure}[h!]
\centering
\includegraphics[width=.45\linewidth]{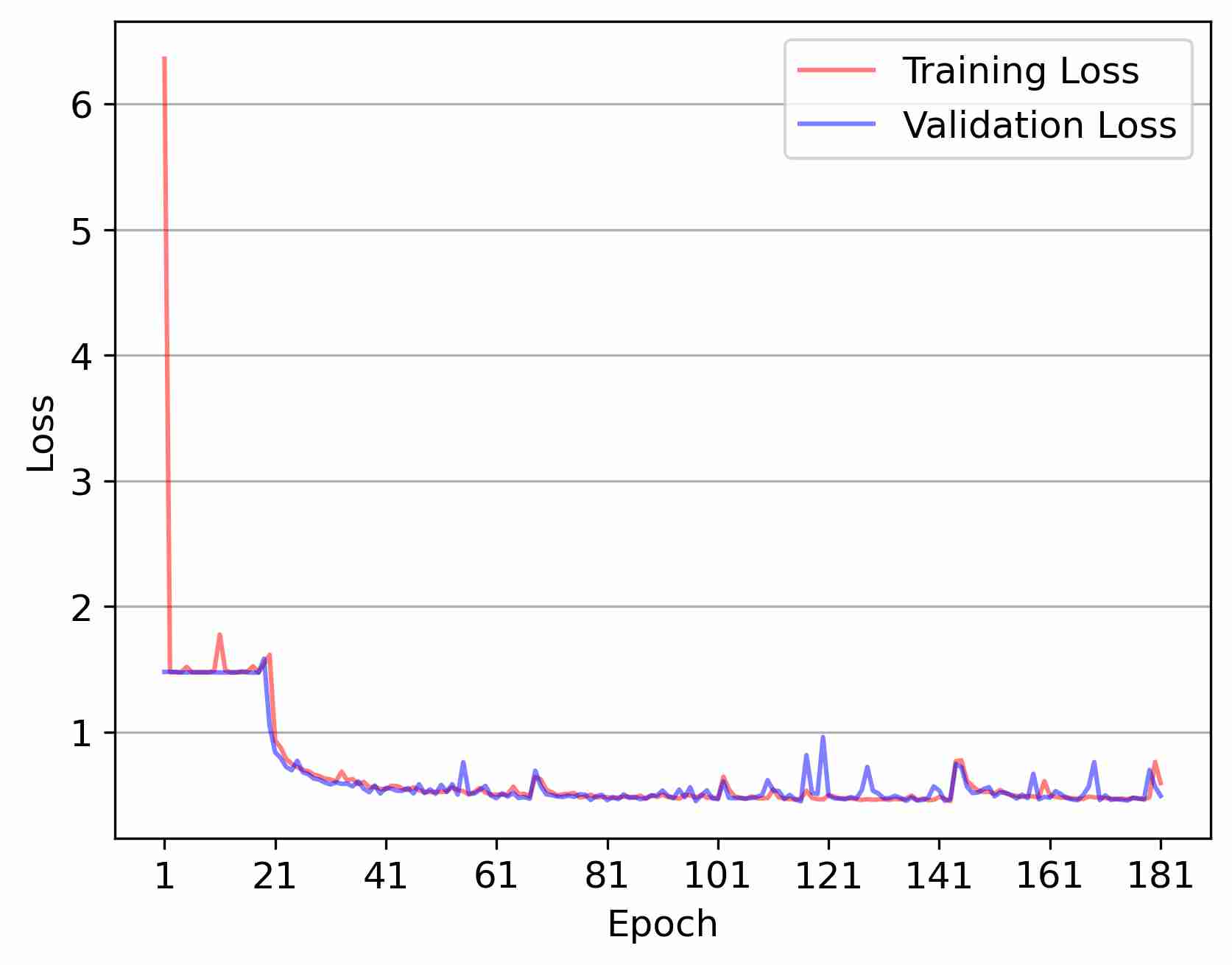}
\includegraphics[width=.45\linewidth]{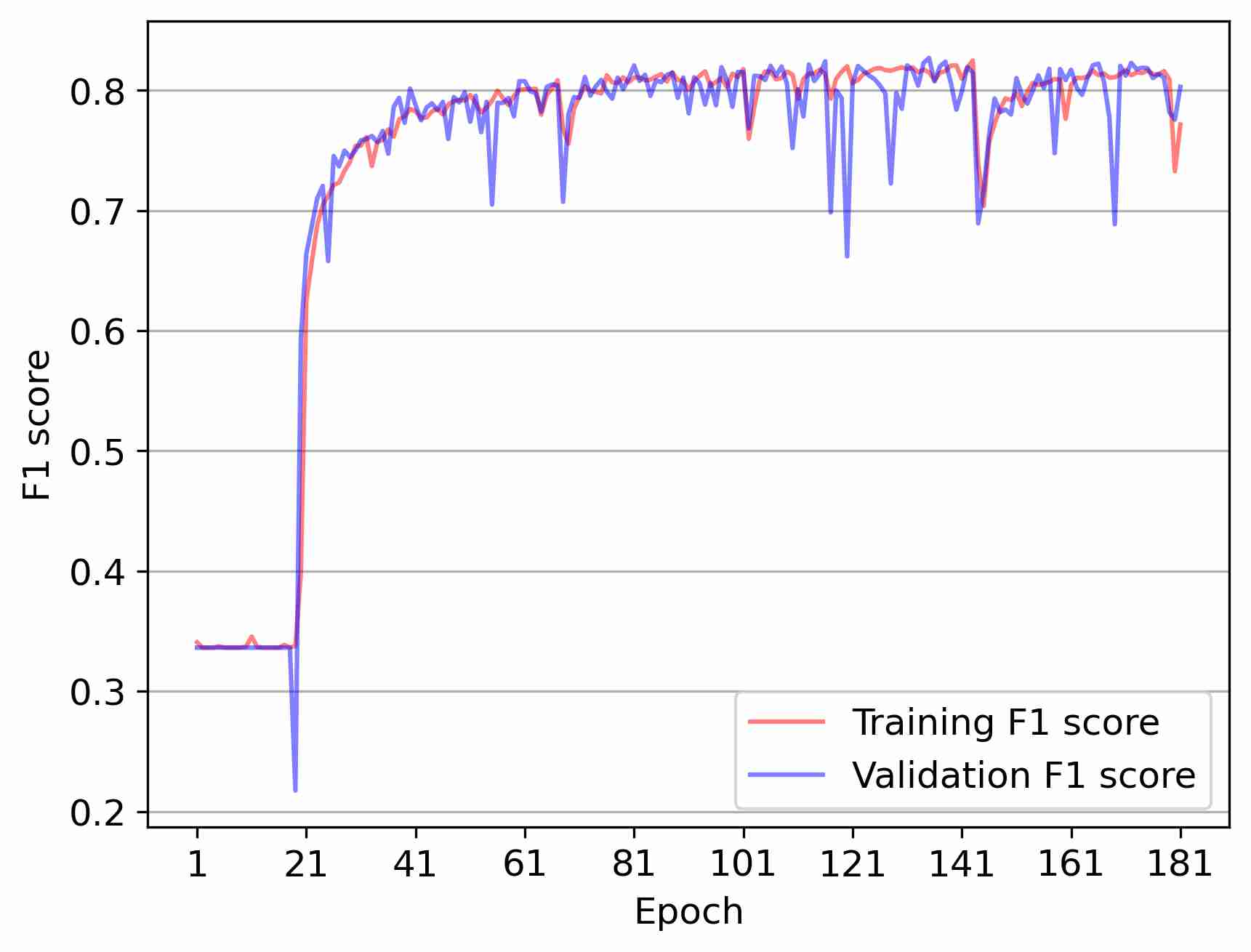}
\caption{Loss value and F1 score during training on the MicrobiaS1 dataset. The differences between training and validation loss values and F1 scores remain relatively small.}
\label{fig:curves_baseline_v4}
\end{figure}

The performance of MicrobiaNet on the MicrobiaS1 dataset is used as the baseline, as its F1 score is the closest to the average of validation sets and F1 score is the key metric. As shown in Figure \ref{fig:curves_baseline_v4}, the gap between training loss and validation loss remains consistently small throughout training on the MicrobiaS1 dataset. The gap between training and validation F1 scores is also small, despite the abrupt changes after 100 epochs. These observations suggest that the model is neither overfitting nor underfitting, which could be early terminated if the validation result stops improving. The training loss and validation loss on the MicrobiaS2-S5 datasets follow a similar trend as illustrated in Appendix~\ref{appendix-baseline}.

\begin{table}[h!]
       \centering
       \caption{Classification results on the MicrobiaS1 validation set.}
       \label{tab_cr_v4_valid}
       \begin{tabular}{p{0.2\textwidth}>{\centering}p{0.1\textwidth}>{\centering}p{0.1\textwidth}>{\centering\arraybackslash}p{0.15\textwidth}}
               \hline
               Class Name     & Precision & Recall & F1 score \\ \hline
               One-colony     & 0.94      & 0.98   & 0.96     \\
               Two-colonies   & 0.82      & 0.81   & 0.82     \\
               Three-colonies & 0.65      & 0.68   & 0.66     \\
               Four-colonies  & 0.48      & 0.38   & 0.43     \\
               Five-colonies  & 0.42      & 0.26   & 0.32     \\
               Six-colonies   & 0.64      & 0.61   & 0.63     \\
               Outlier        & 0.86      & 0.80   & 0.83     \\ \hline
       \end{tabular}
\end{table}

\begin{figure}[h!]
    \centering
    \includegraphics[width=0.45\linewidth]{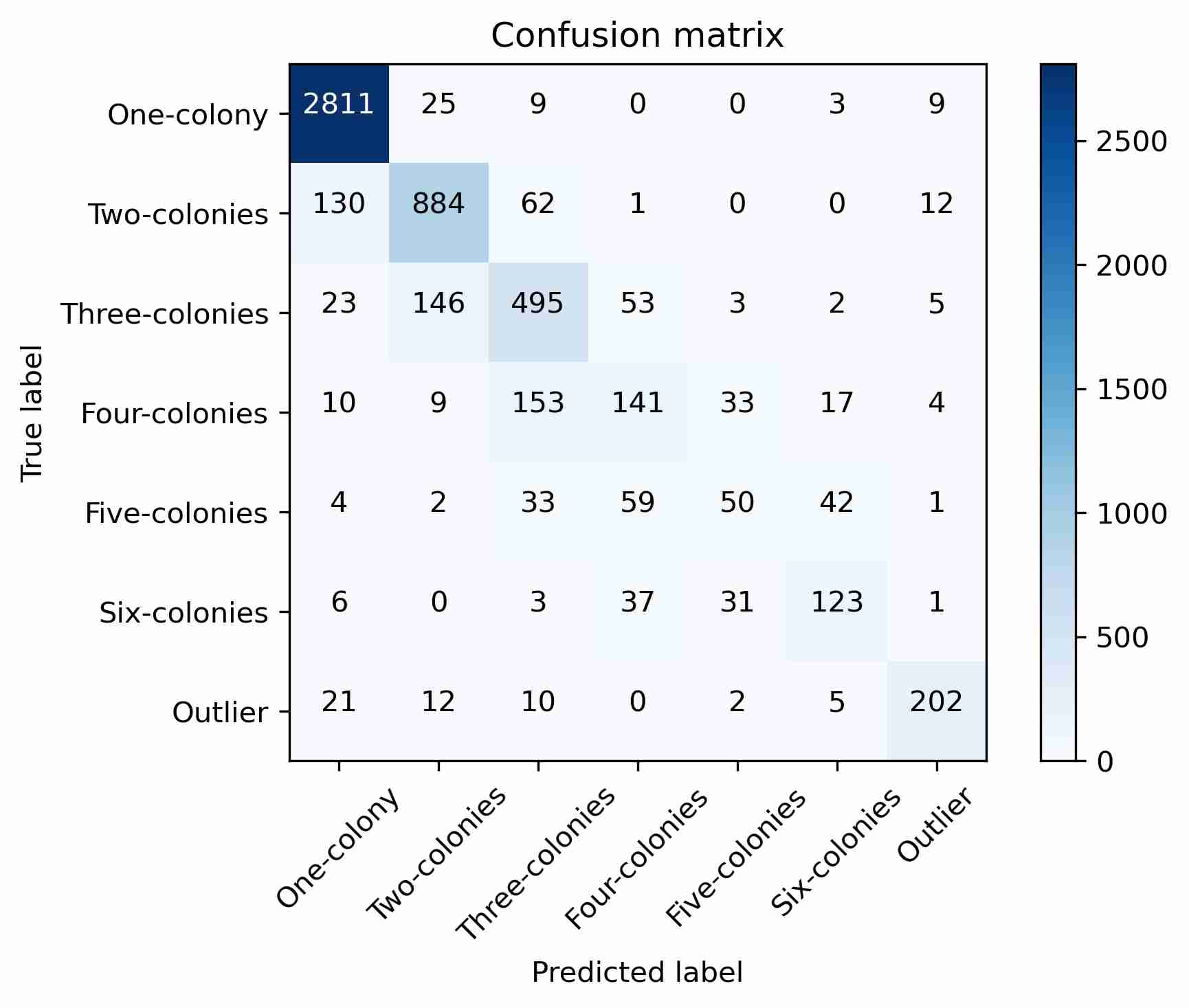}
    \caption{Confusion matrix for MicrobiaS1 validation results. Most misclassifications occur between neighbouring classes.}
    \label{fig:cm_v4_valid}
\end{figure}

Table \ref{tab_cr_v4_valid} presents the class-wise results for the MicrobiaS1 validation set, derived from the confusion matrix in Figure \ref{fig:cm_v4_valid}. Some minority classes, such as Five-colonies and Six-colonies, show F1 scores less than half that of the majority One-colony class, which may initially suggest bias. However, the confusion matrix shows that most misclassifications occur between neighbouring classes rather than being predicted as the majority class, suggesting the model struggles with similar classes rather than favouring the majority class. This pattern, also observed by MicrobiaNet’s authors, is consistent across the MicrobiaS2-S5 validation sets in Appendix~\ref{appendix-baseline}.

\begin{figure}[h!]
\centering
\includegraphics[width=0.9\linewidth]{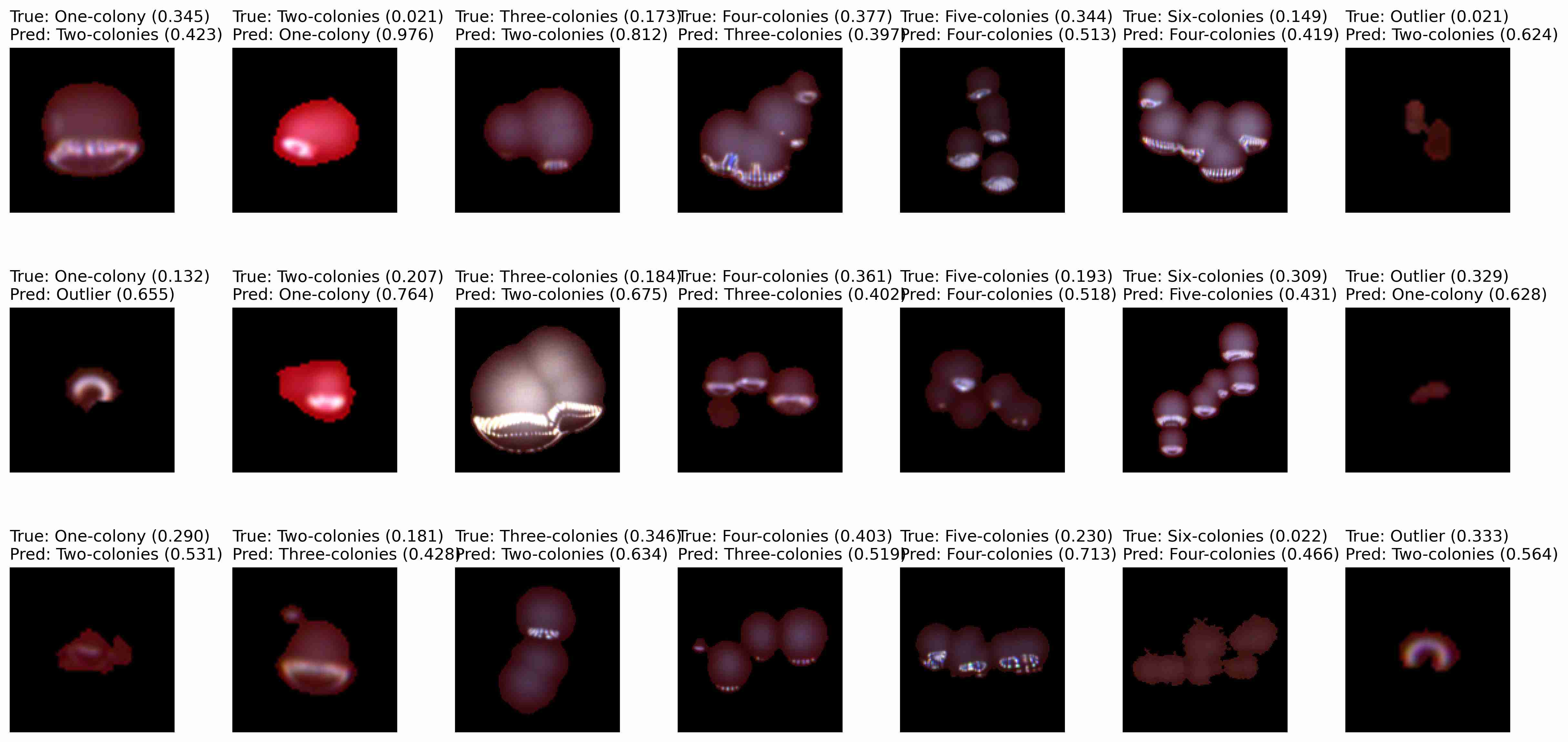}
\caption{Examples of incorrect predictions from the MicrobiaS1 validation set. Colonies with similar visual appearance are often misclassified. For example, Three-colonies images are frequently predicted as Two-colonies.}
\label{fig:incorrects_v4_valid}
\end{figure}

Examples of misclassified samples from the MicrobiaS1 validation set are shown in Figure \ref{fig:incorrects_v4_valid}. The probabilities of true and predicted labels, calculated by the softmax function, are displayed above each colony image. Three-colonies images are consistently predicted as Two-colonies, likely due to their visual similarity and the design of MicrobiaNet. A similar confusion occurs between Five-colonies and Six-colonies images, highlighting the model’s difficulty distinguishing visually similar neighbouring classes.

\subsubsection{Case study summary}
This case study has empirically demonstrated that MicrobiaNet achieves an average validation F1 score of 0.82 with a standard deviation of 0.0063 across five independent splits of Microbia Dataset. Additionally, it is uncovered that some minority classes have a significantly lower F1 score than that of the majority One-colony class. On the other hand, the minority Outlier class's F1 score is only 0.13 lower than that of the majority One-colony class in the baseline performance.

MicrobiaNet does not exhibit bias towards the majority class, as most of the incorrect predictions occur between neighbouring classes rather than being disproportionately assigned to the majority One-colony class. The experimental results also demonstrate that MicrobiaNet has difficulty distinguishing colonies of three or more individuals. This limitation may be a result of high visual similarity across classes in the dataset together with the design of MicrobiaNet. With the selection of baseline performance, other case studies will investigate MicrobiaNet's performance with XAI.

\subsection{Explainability of the colony cardinality classification network}
\label{sec_interpretability_of_baseline_model}
To date, no studies have investigated MicrobiaNet's explainability. Therefore, a series of experiments is conducted to evaluate its baseline performance. The goal is to leverage XAI to identify factors underlying MicrobiaNet’s validated limitation and guide future improvements. Unless otherwise specified, the trained MicrobiaNet in this section refers to the one obtained from the baseline performance, i.e. the MicrobiaS1 dataset. The XAI approaches used in this analysis include visualisation of network layer outputs, learned features, and class activation maps, as explained and justified in \S~\ref{subsec-xai}.

\subsubsection{Visualisation of network layer outputs}
\begin{figure}[h!]
\centering
\includegraphics[width=.45\linewidth]{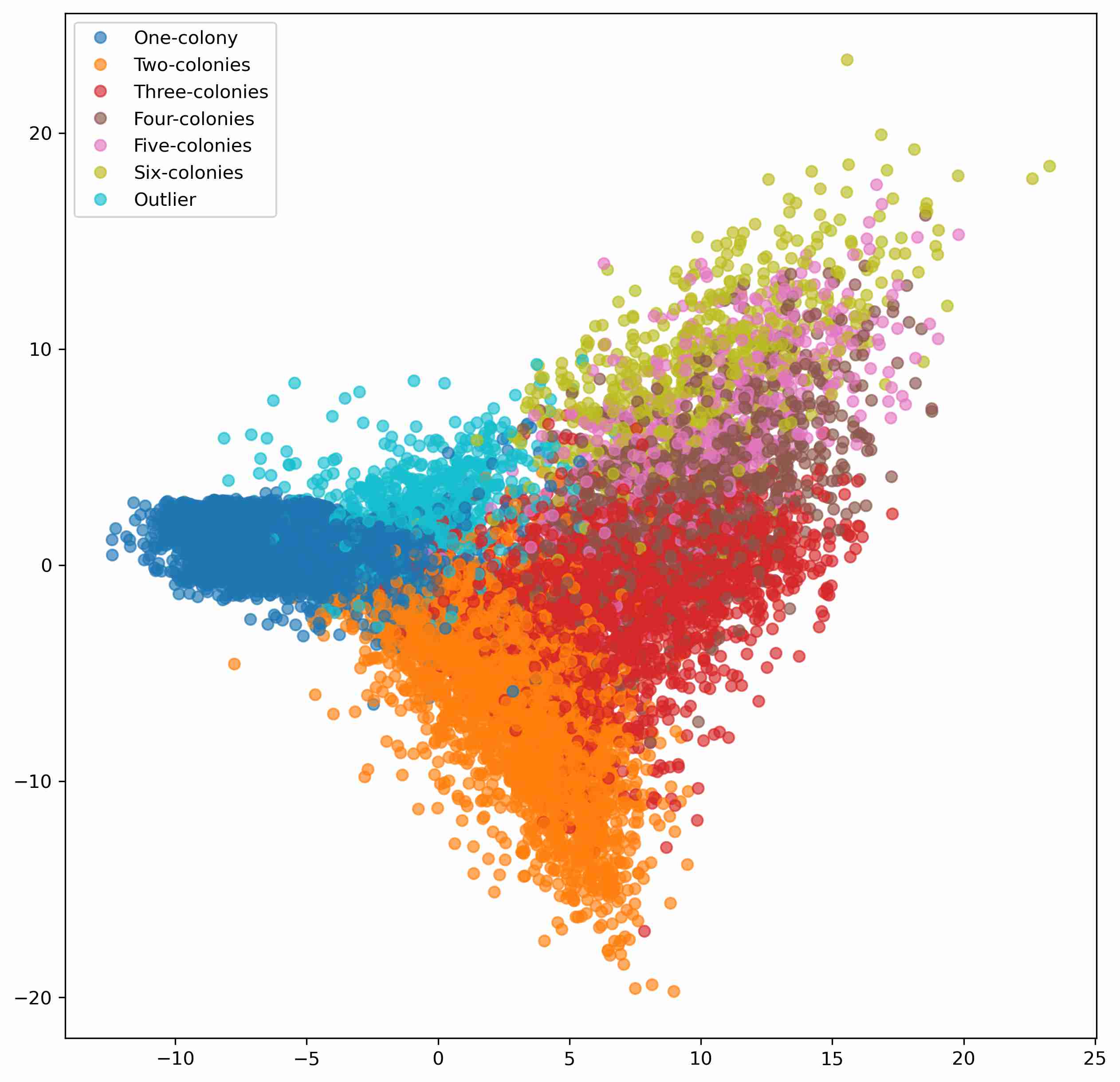}
\includegraphics[width=.45\linewidth]{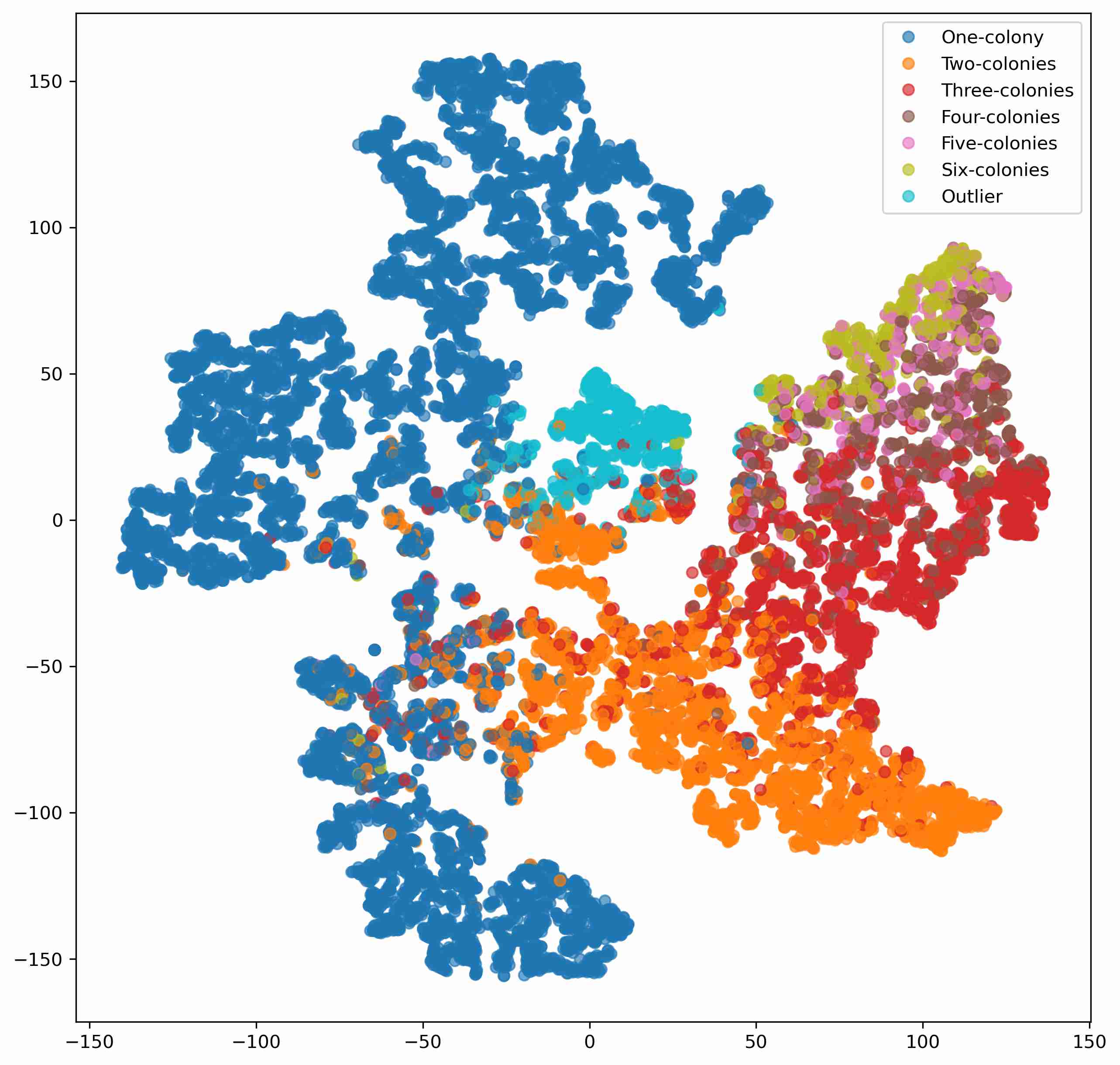}
\caption{PCA-reduced (left) and t-SNE-reduced (right) representations of the last network layer outputs from the baseline model on the MicrobiaS1 training set. Each colour corresponds to one of the seven classes in the dataset. Some classes form distinct clusters (blue, orange, and cyan regions), whereas others (red, brown, pink, and olive regions) are entangled and overlapped extensively.}
\label{fig:visualisation_of_network_layers_with_pca_training}
\end{figure}

\begin{figure}[h!]
\centering
\includegraphics[width=.45\linewidth]{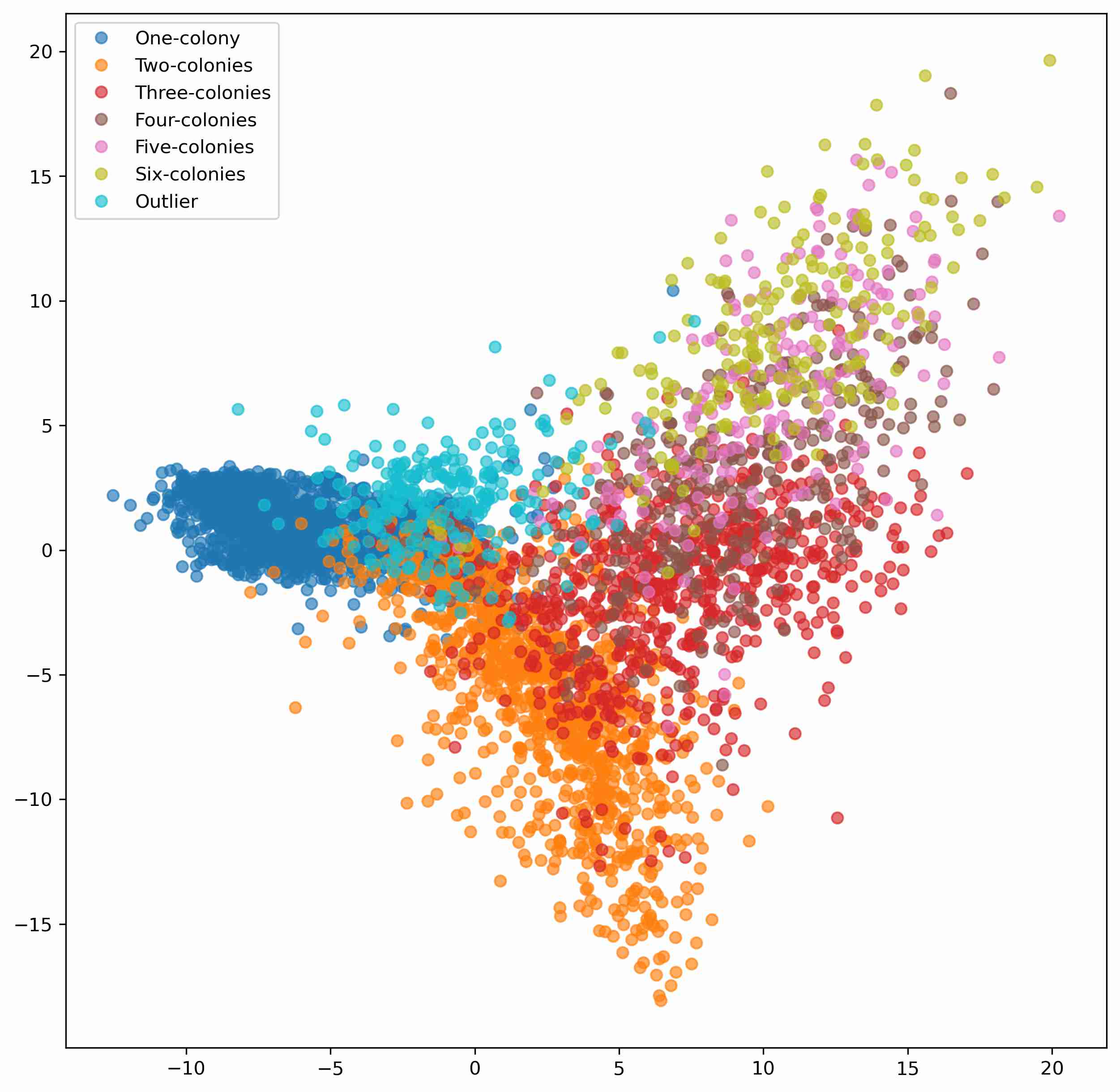}
\includegraphics[width=.45\linewidth]{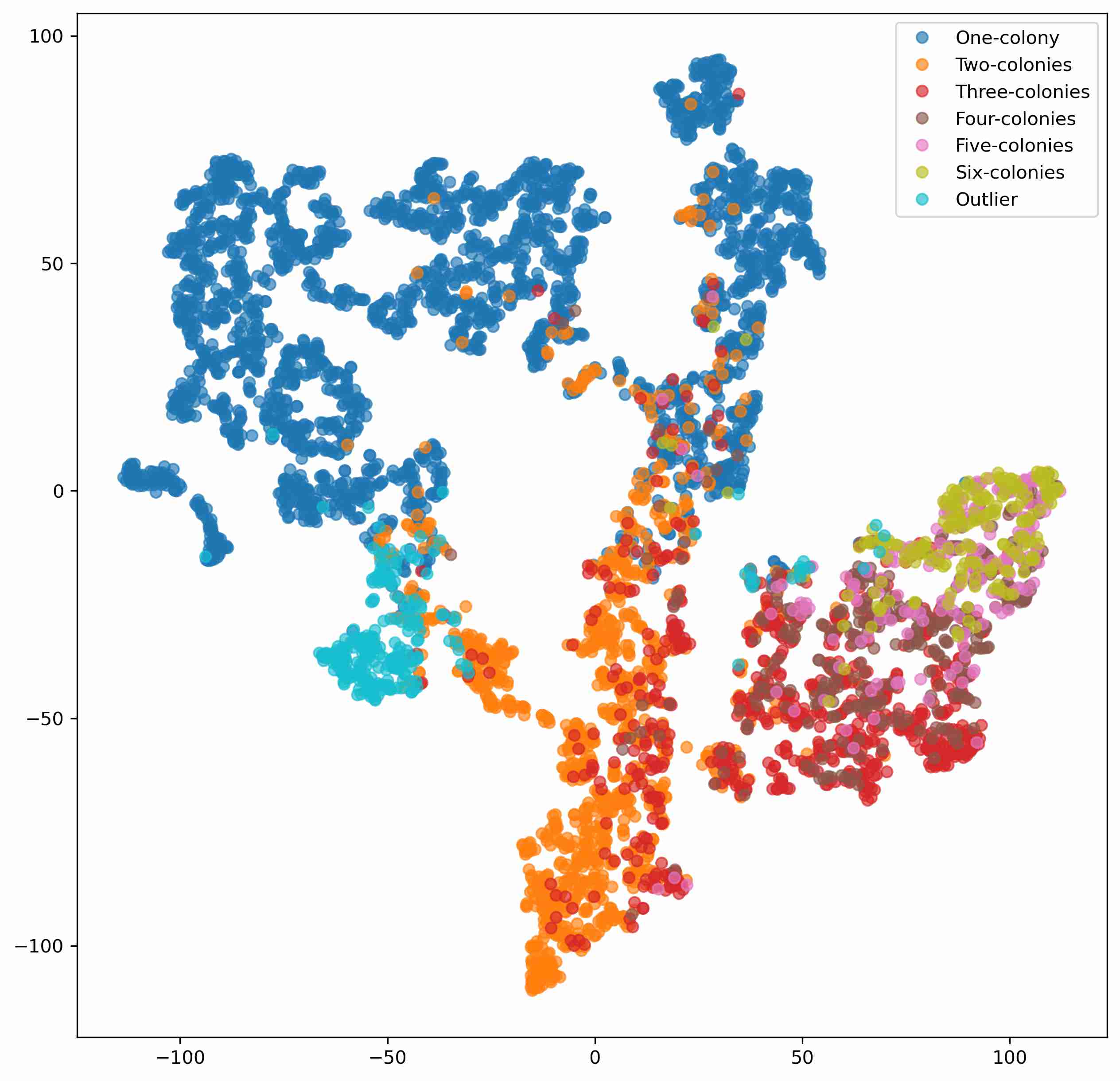}
\caption{Same as Figure \ref{fig:visualisation_of_network_layers_with_pca_training}, but for the validation set.}
\label{fig:visualisation_of_network_layers_with_pca_valid}
\end{figure}

Figures \ref{fig:visualisation_of_network_layers_with_pca_training} and \ref{fig:visualisation_of_network_layers_with_pca_valid} present PCA-reduced and t-SNE-reduced representations of the last network layer outputs from the baseline model on the MicrobiaS1 training and validation sets, respectively. Both visualisations clearly reflect the imbalanced class distribution. Specifically, the One-colony class forms the largest and most distinct cluster (blue region), which is consistent with its dominance in the dataset. In addition, Two-colonies and Outlier classes are distinct with their own clusters (orange and cyan regions), which explains their strong performance in Table \ref{tab_cr_v4_valid}.

Conversely, Three-colonies, Four-colonies, Five-colonies, and Six-colonies classes (red, brown, pink, and olive regions) are highly entangled and overlapped extensively, almost forming a single cluster. This lack of separability aligns with their comparatively poorer performance in Table \ref{tab_cr_v4_valid}. The observed overlapping feature space corroborates the visual similarity across these classes in the dataset, suggesting that such similarity is a primary factor contributing to MicrobiaNet’s reduced performance for these classes. Additional analyses on the penultimate layer representations using t-SNE with varying perplexity settings are presented in Appendix~\ref{appendix-explainability}, as their visualisations exhibit patterns consistent with those of the final layer.

\subsubsection{Visualisation of learned features}
\begin{figure}[h!]
\centering
\includegraphics[width=0.85\linewidth]{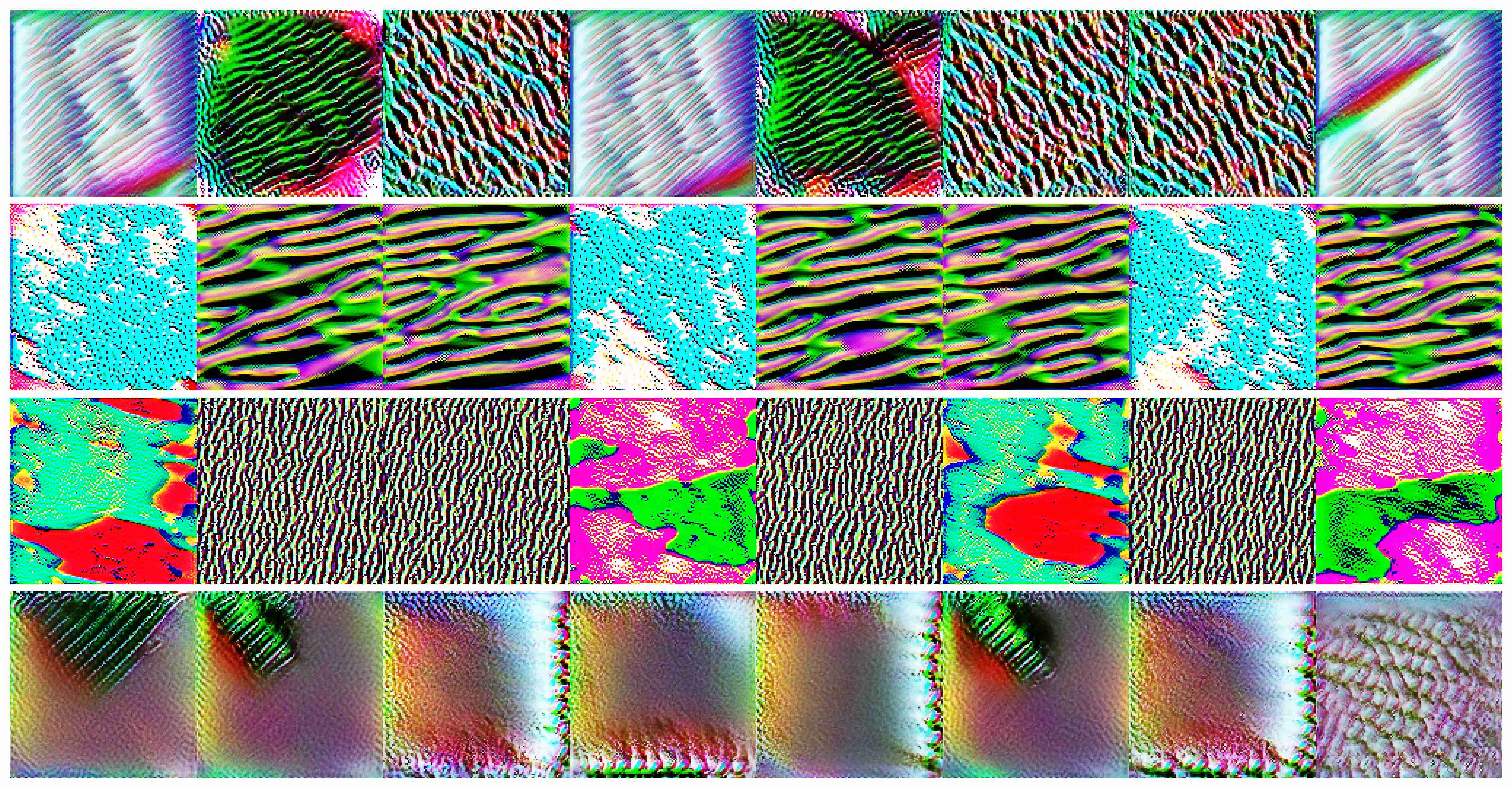}
\caption{Visualisation of feature maps that strongly activate the first convolutional kernel in each of the four convolutional layers. Eight feature maps (shown in the eight columns) are displayed for each of the four convolutional layers (shown in the four rows). The first convolutional kernels in the first three convolutional layers (rows 1-3) appear to capture simple texture-based representations. In contrast, the first convolutional kernel in the fourth convolutional layer (row 4) demonstrates sensitivity to red blob-like shapes that bear a vague resemblance to colonies.}
\label{fig:vis_v4_train_first_conv}
\end{figure}

\begin{figure}[h!]
\centering
\includegraphics[width=0.85\linewidth]{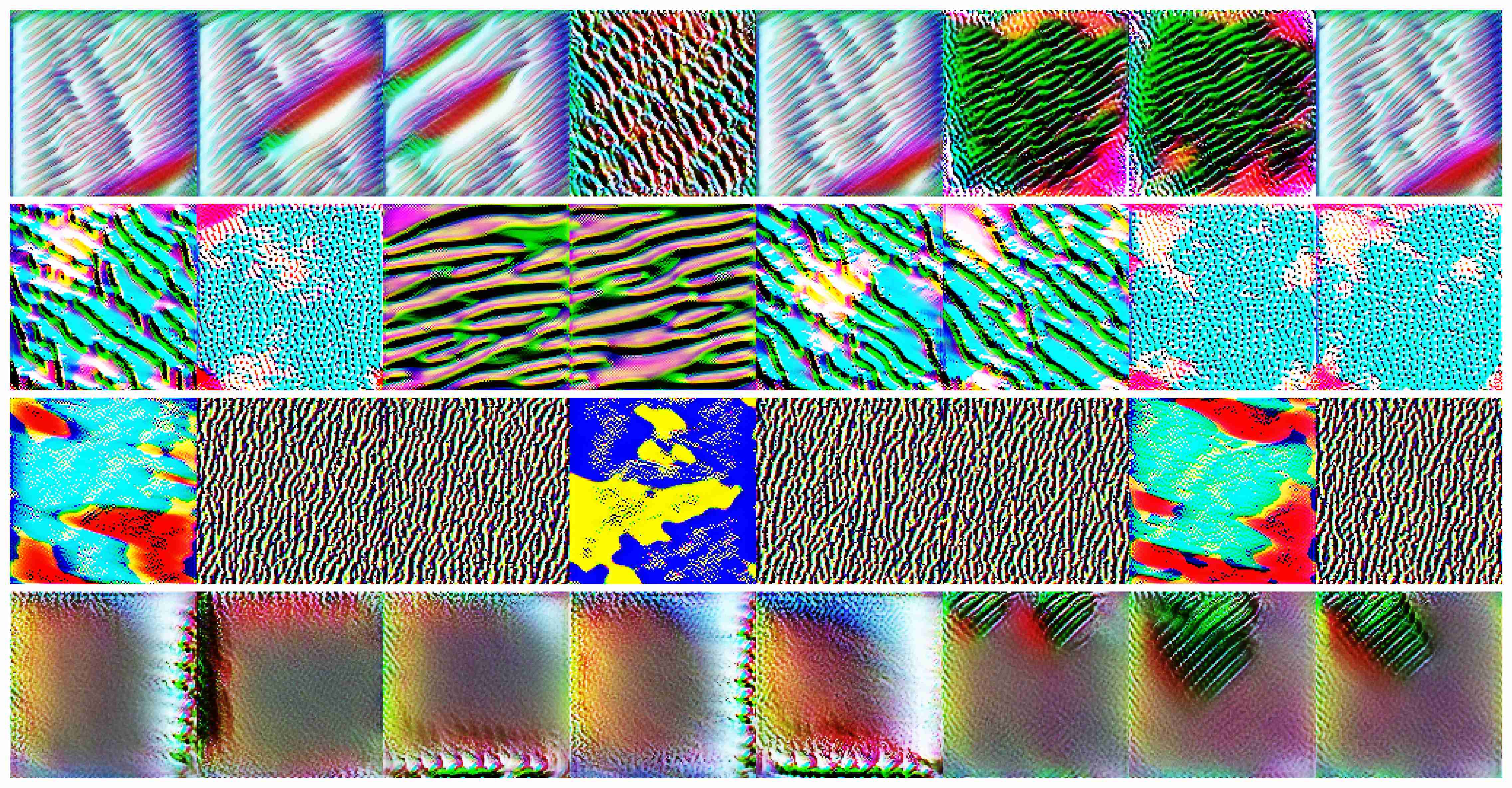}
\caption{Same as Figure \ref{fig:vis_v4_train_first_conv}, but for the second convolutional kernel.}
\label{fig:vis_v4_train_second_conv}
\end{figure}

Figures \ref{fig:vis_v4_train_first_conv} and \ref{fig:vis_v4_train_second_conv} illustrate eight features that activate the first and second kernels across the four convolutional layers of the trained MicrobiaNet, respectively. At a visual level, these feature maps do not clearly reveal colony-specific information, such as shape or size. Nonetheless, certain patterns are observable. For instance, kernels in the first convolutional layer, particularly those shown in the first row of Figures \ref{fig:vis_v4_train_first_conv} and \ref{fig:vis_v4_train_second_conv}, appear to learn simple texture-based representations. Additionally, the first kernel in the fourth convolutional layer (the fourth row in Figure \ref{fig:vis_v4_train_first_conv}) demonstrates sensitivity to red blob-like shapes that bear a vague resemblance to colonies.

Similar trends are observed in the features extracted by the second kernel across the four layers, as shown in Figure \ref{fig:vis_v4_train_second_conv}. Overall, these visualisations provide limited intuitive insight into colony characteristics. This finding is consistent with the claim made by~\citet{salahuddin_transparency_2022} that it is hard to interpret structures and patterns in medical images as they are not very obvious. Additional visualisations for the third, fourth, fifth, and sixth kernels in each layer are provided in Appendix~\ref{appendix-explainability} with the same finding.

\subsubsection{Visualisation of class activation maps}
\begin{figure}[h!]
\centering
\includegraphics[width=0.85\linewidth]{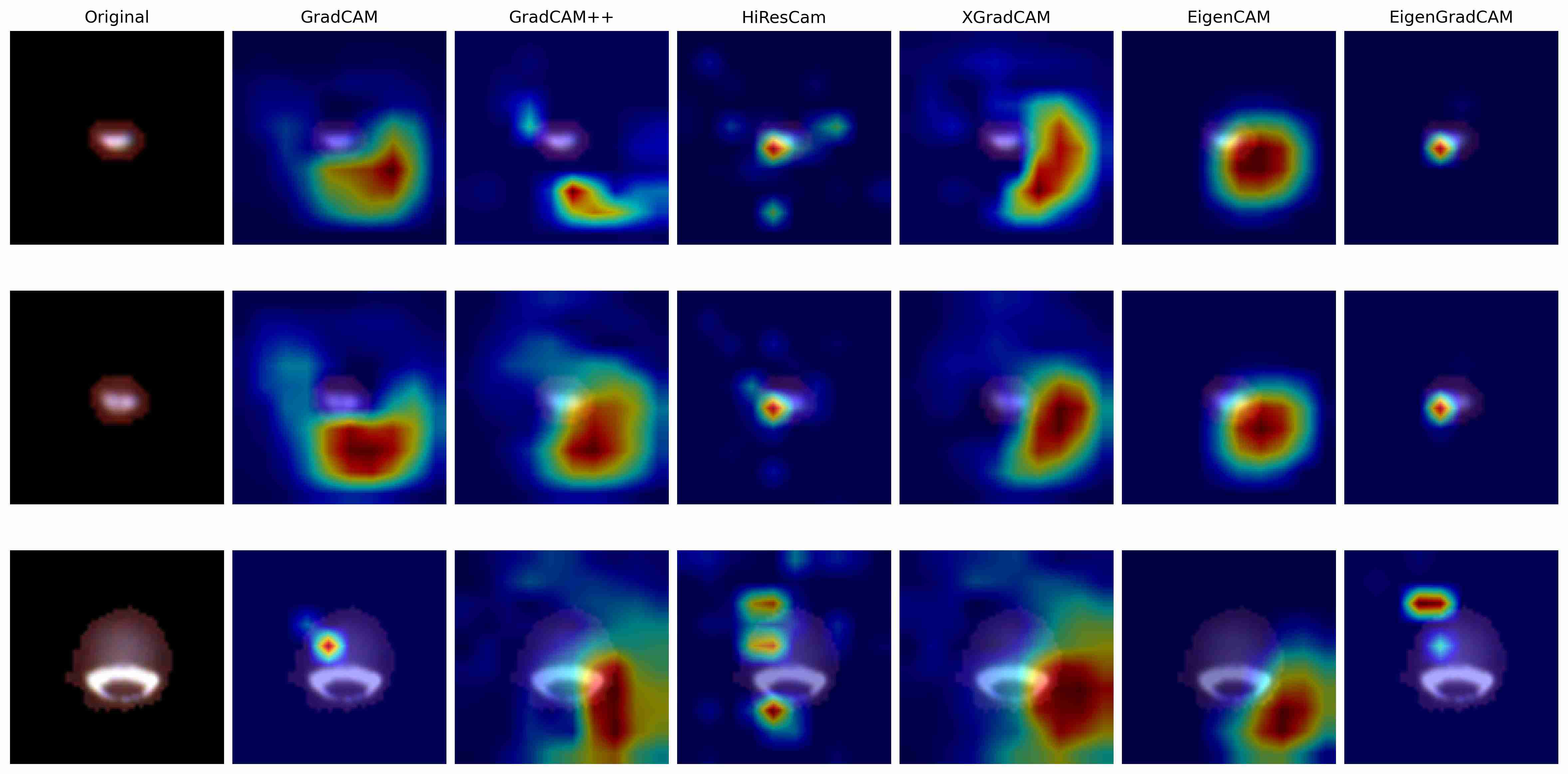}
\caption{Class activation map visualisation for One-colony images. Highlighted regions indicate the areas that contribute most strongly to the model’s predictions, with colour intensity representing the degree of contribution.}
\label{fig:cam_vis_v4_one_colony_conv}
\end{figure}

Figure \ref{fig:cam_vis_v4_one_colony_conv} presents visualised activation maps for the One-colony class from the MicrobiaS1 training set. Highlighted regions indicate the areas contributing most strongly to the model’s predictions, with colour intensity representing the degree of contribution. Among the evaluated explainability methods, Grad-CAM demonstrates comparatively weak performance, as its activation maps show limited overlap with the target colony region. In contrast, the other methods, such as Grad-CAM++, HiResCAM, XGrad-CAM, Eigen-CAM, and EigenGrad-CAM, produce activation maps with noticeably stronger alignment to the colony. 

Based on qualitative inspection, Eigen-CAM provides the most accurate localisation. This observation indicates that the projections onto the first principal component of the feature maps carry valuable spatial information for identifying the target colony. Nevertheless, none of the visualisation methods clearly reveal a direct causal link between the colony and its assigned class. Additional activation maps for the remaining classes are provided in Appendix~\ref{appendix-explainability}, as they show similar patterns.

\subsubsection{Case study summary}
This case study has investigated MicrobiaNet's explainability based on its baseline performance by visualising network layer outputs, features, and class activation maps. The main insight is that the One-colony, Two-colonies, and Outlier classes are distinct with their own clusters in the feature space, whereas all other classes are entangled and overlapped extensively. These overlapped classes almost form a single cluster, which is a direct result of their high visual similarities. 

Despite these observations, neither feature visualisation nor class activation map visualisation provides informative interpretation of predictions. This limitation may arise from the inherent constraints of these methods. Feature visualisation may fail to localise semantically meaningful structures when inter-class differences are subtle. Class activation map visualisation often produces diffuse or noisy activation maps when class-discriminative features exhibit low contrast relative to the background. Both approaches are also sensitive to model architecture and layer selection, which can further reduce their interpretability in fine-grained or texture-dependent tasks.

Nonetheless, some useful patterns can still be identified. For example, the first kernel in MicrobiaNet’s fourth convolutional layer appears to detect red blob-like structures that loosely resemble colony shapes. Similarly, the projections onto the first principal component of the feature maps have demonstrated utility for localising target colonies as evidenced by the Eigen-CAM visualisations.

\subsection{Analysis of the impact of class imbalance on MicrobiaNet}
\label{subsec_tackling_class_imb_down_sampling}
While the previous case study demonstrates that the high visual similarity across classes limits MicrobiaNet’s performance, class imbalance may also play a role. The Three-colonies, Four-colonies, Five-colonies, and Six-colonies classes contain substantially fewer samples, which could affect counting accuracy. To investigate this, MicrobiaNet is trained on a balanced dataset, referred to as the balanced MicrobiaNet model.

\subsubsection{Experimental setup}
Data downsampling is used to tackle class imbalance by randomly removing extra data points from all classes except the least minority class. It is selected in this experiment due to its simplicity. Data upsampling, which randomly duplicates data points from each class except the majority class, is not considered to avoid repetitive features in the feature space during the interpretation process. This means that images from the One-colony, Two-colonies, Three-colonies, Four-colonies, Six-colonies, and Outlier classes are downsampled to match the number of images in the Five-colonies class.

Data downsampling is only applied to the MicrobiaS1 training set with five different seeds, generating five balanced training set. Each set is used to train MicrobiaNet independently for robust evaluation. The trained balanced models are evaluated on the MicrobiaS1 validation set which is still imbalanced. The downsampled MicrobiaS1 set is referred to as MicrobiaS1B1 dataset, where B stands for balanced and the number indicates the downsampling seed.\footnote{The MicrobiaS1B1 validation set is identical to the MicrobiaS1 validation set.}

Training and evaluation follow the procedures described in \S \ref{subsubsec_training_and_eval}. Similar to the previous case studies in \S \ref{sec_colony_card_cls_baseline} and \S \ref{sec_interpretability_of_baseline_model}, one of these five trained balanced MicrobiaNet models will be selected to interpret by visualising network layer outputs. The visualisation of features and class activation maps are excluded here as \S \ref{sec_interpretability_of_baseline_model} concludes that they are not informative.

\subsubsection{Results}

\begin{table}[h!]
       \centering
       \caption{Overall evaluation results on the MicrobiaS1B1, MicrobiaS1B2, MicrobiaS1B3, MicrobiaS1B4, and MicrobiaS1B5 datasets.}
       \label{Microbia_model_results_from_balanced_sets}
       \begin{tabular}{p{0.02\textwidth}>{\centering}p{0.18\textwidth}>{\centering}p{0.05\textwidth}>{\centering}p{0.05\textwidth}>{\centering}p{0.05\textwidth}>{\centering}p{0.05\textwidth}>{\centering}p{0.05\textwidth}>{\centering}p{0.06\textwidth}>{\centering\arraybackslash}p{0.06\textwidth}}
               \hline
                                                                     & \multirow{2}{*}{Metric} & \multicolumn{5}{c}{S1 Balanced with Different Seeds} & \multirow{2}{*}{Mean} & \multirow{2}{*}{Std}                                                 \\ \cmidrule{3-7}
                                                                     &                         & B1                                                           & B2                    & B3                        & B4        & B5        &           &           \\ \hline
               \multirow{4}{*}{\rotatebox[origin=c]{90}{Training}}   & Precision               & 0.87                                                         & 0.93                  & 0.68                      & 0.76      & 0.88      & 0.82      & 0.0915      \\
                                                                     & Recall                  & 0.86                                                         & 0.93                  & 0.68                      & 0.76      & 0.87      & 0.82      & 0.0910      \\
                                                                     & F1 score                & \bf{0.86}                                                    & \bf{0.93}             & \bf{0.67}                 & \bf{0.75} & \bf{0.87} & \bf{0.82} & \bf{0.0936} \\
                                                                     & Accuracy (\%)           & 86.41                                                        & 92.99                 & 67.58                     & 75.98     & 87.39     & 82.07     & 9.1000      \\ \cmidrule{1-9}
               \multirow{4}{*}{\rotatebox[origin=c]{90}{Validation}} & Precision               & 0.77                                                         & 0.77                  & 0.75                      & 0.76      & 0.77      & 0.77      & 0.0071      \\
                                                                     & Recall                  & 0.76                                                         & 0.76                  & 0.75                      & 0.76      & 0.76      & 0.76      & 0.0045      \\
                                                                     & F1 score                & \bf{0.76}                                                    & \bf{0.77}             & \bf{0.75}                 & \bf{0.76} & \bf{0.76} & \bf{0.76} & \bf{0.0050} \\
                                                                     & Accuracy (\%)           & 76.27                                                        & 76.21                 & 75.12                     & 76.37     & 75.95     & 75.99     & 0.0045      \\ \hline
       \end{tabular}
\end{table}

The overall evaluation results from MicrobiaNet on the MicrobiaS1B1-S1B5 datasets are listed in Table \ref{Microbia_model_results_from_balanced_sets}. On average, the balanced MicrobiaNet model achieves a training F1 score of 0.82 and a validation F1 score of 0.76. The corresponding standard deviations across the five datasets are 0.0936 and 0.0050, respectively. Compared with the results when the model is trained on imbalanced data in Table \ref{Microbia_model_results_with_diff_data_split}, downsampling contributes to only a marginal decrease of 0.06 in the validation F1 score despite the significant reduction of training data. This observation suggests that class imbalance plays a limited role in the baseline model performance.

\begin{table}[h!]
       \centering
       \caption{Classification results on the MicrobiaS1B1 training set.}
       \label{tab_cr_v440_train}
       \begin{tabular}{p{0.2\textwidth}>{\centering}p{0.1\textwidth}>{\centering}p{0.1\textwidth}>{\centering\arraybackslash}p{0.15\textwidth}}
               \hline
               Class Name     & Precision & Recall & F1 score \\ \hline
               One-colony     & 0.85      & 0.98   & 0.91     \\
               Two-colonies   & 0.86      & 0.94   & 0.90     \\
               Three-colonies & 0.80      & 0.80   & 0.80     \\
               Four-colonies  & 0.88      & 0.71   & 0.78     \\
               Five-colonies  & 0.78      & 0.83   & 0.80     \\
               Six-colonies   & 0.91      & 0.83   & 0.87     \\
               Outlier        & 0.99      & 0.97   & 0.98     \\ \hline
       \end{tabular}
\end{table}

\begin{figure}[h!]
\centering
\includegraphics[width=0.45\linewidth]{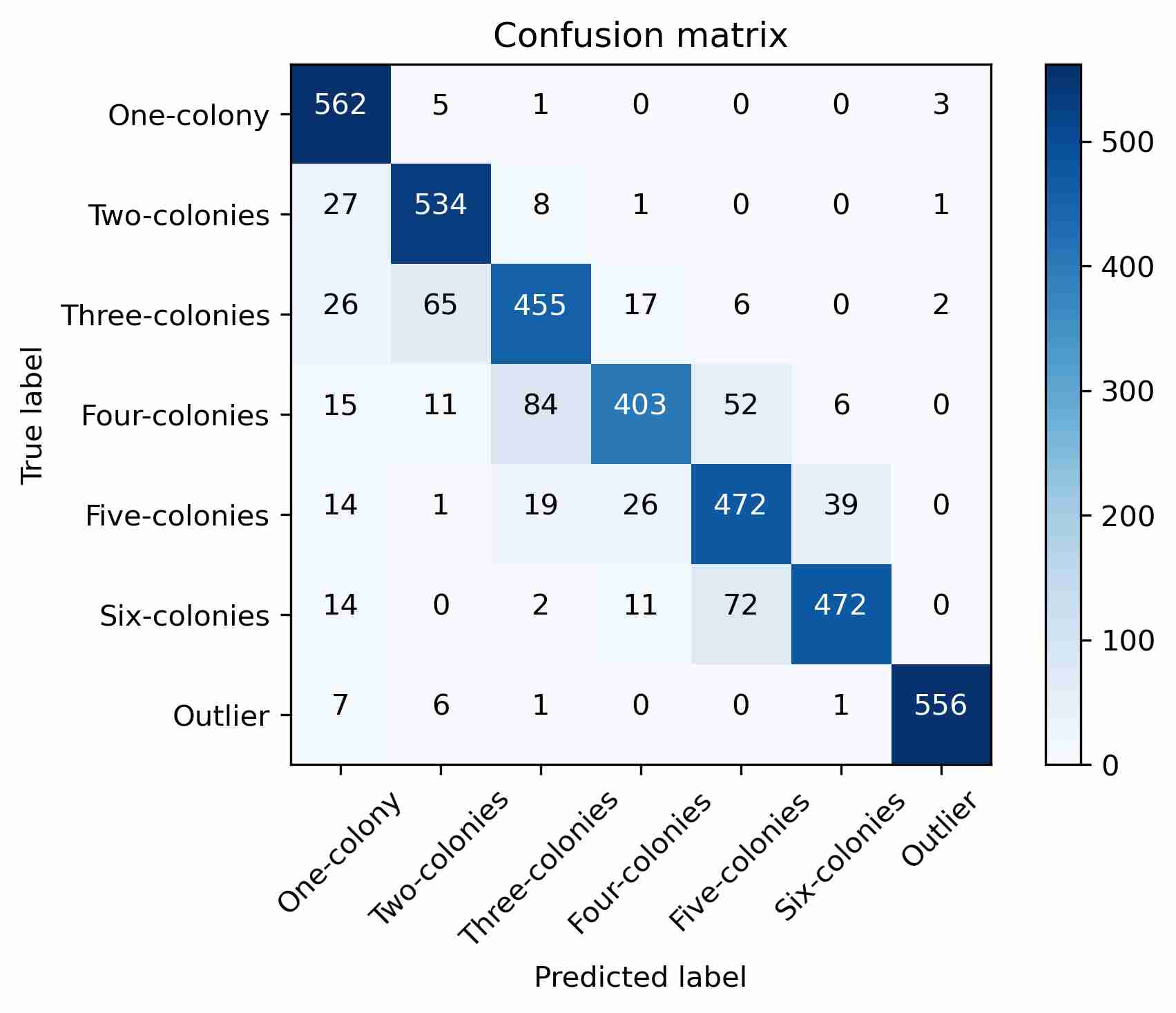}
\caption{Confusion matrix for MicrobiaS1B1 training results. Most misclassifications occur between neighbouring classes.}
\label{fig:cm_v440_train}
\end{figure}

\begin{table}[h!]
       \centering
       \caption{Classification results on the MicrobiaS1(B1) validation set.}
       \label{tab_cr_v440_valid}
       \begin{tabular}{p{0.2\textwidth}>{\centering}p{0.1\textwidth}>{\centering}p{0.1\textwidth}>{\centering\arraybackslash}p{0.15\textwidth}}
               \hline
               Class Name     & Precision & Recall & F1 score \\ \hline
               One-colony     & 0.94      & 0.94   & 0.94     \\
               Two-colonies   & 0.76      & 0.73   & 0.74     \\
               Three-colonies & 0.58      & 0.48   & 0.52     \\
               Four-colonies  & 0.33      & 0.25   & 0.28     \\
               Five-colonies  & 0.24      & 0.48   & 0.32     \\
               Six-colonies   & 0.54      & 0.50   & 0.52     \\
               Outlier        & 0.66      & 0.88   & 0.76     \\ \hline
       \end{tabular}
\end{table}

\begin{figure}[h!]
\centering
\includegraphics[width=0.45\linewidth]{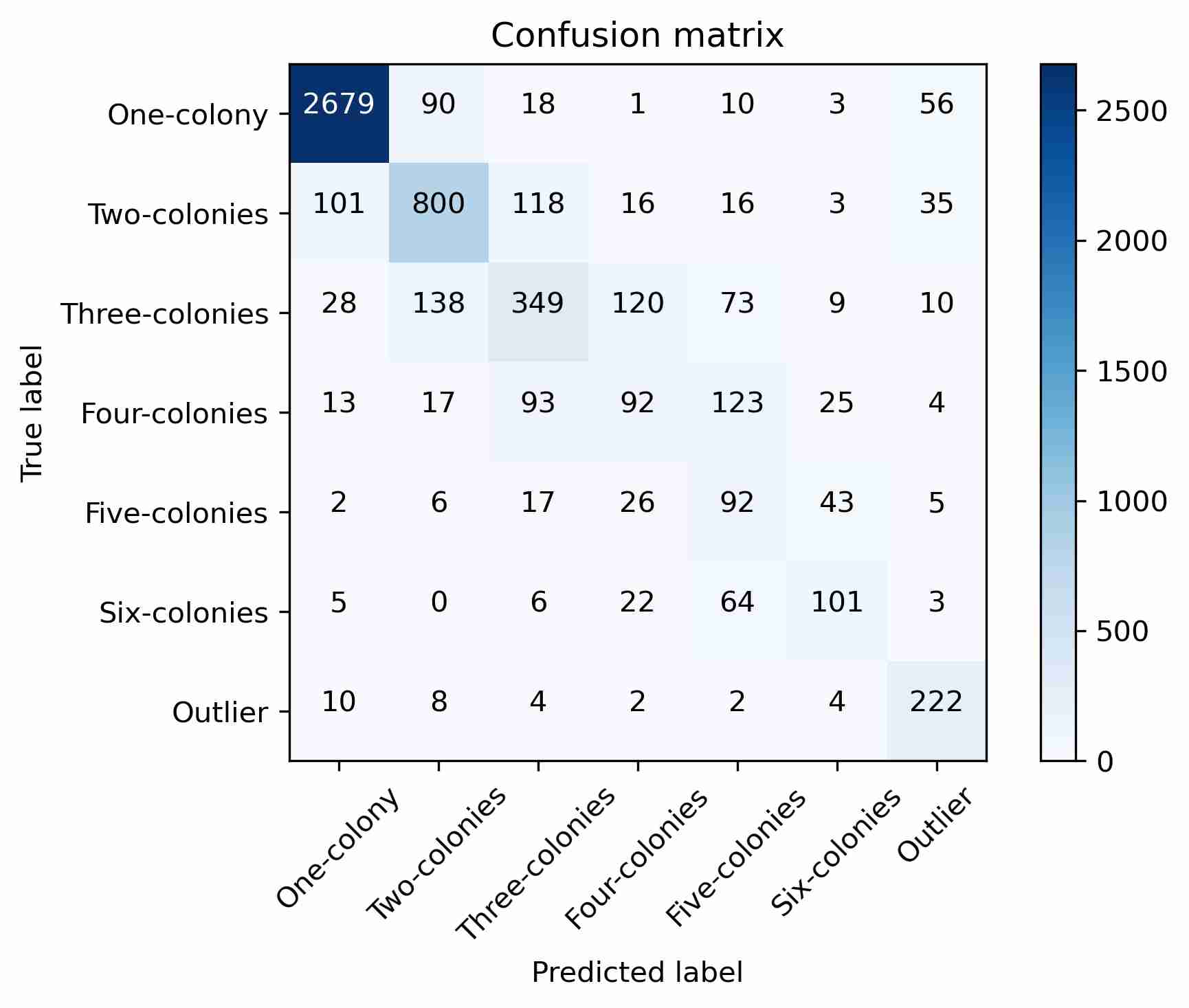}
\caption{Confusion matrix for MicrobiaS1(B1) validation results. Most misclassifications occur between neighbouring classes.}
\label{fig:cm_v440_valid}
\end{figure}

The balanced MicrobiaNet model trained on the MicrobiaS1B1 training set is selected for further analysis with XAI, as its training and validation F1 scores closely match the overall mean. Detailed performance metrics are reported in Tables \ref{tab_cr_v440_train} and \ref{tab_cr_v440_valid}, which are computed from the confusion matrices in Figures \ref{fig:cm_v440_train} and \ref{fig:cm_v440_valid}. Consistent with earlier findings in the baseline performance, One-colony, Two-colonies, and Outlier classes remain the top-$3$ performant classes. The training F1 scores across all classes have a standard deviation of 0.0668, whereas the validation F1 scores have a standard deviation of 0.2232. This further supports the conclusion that class imbalance has only a limited effect on model performance.

Most of the wrong predictions occur between a class and its neighbouring classes. For example, in the Four-colonies class in Figure \ref{fig:cm_v440_valid}, 92 instances are correctly classified, while 93 are predicted as Three-colonies and 123 as Five-colonies. This indicates that the high visual similarity between neighbouring classes is the primary challenge for performance improvement.

\begin{figure}[h!]
\centering
\includegraphics[width=.45\linewidth]{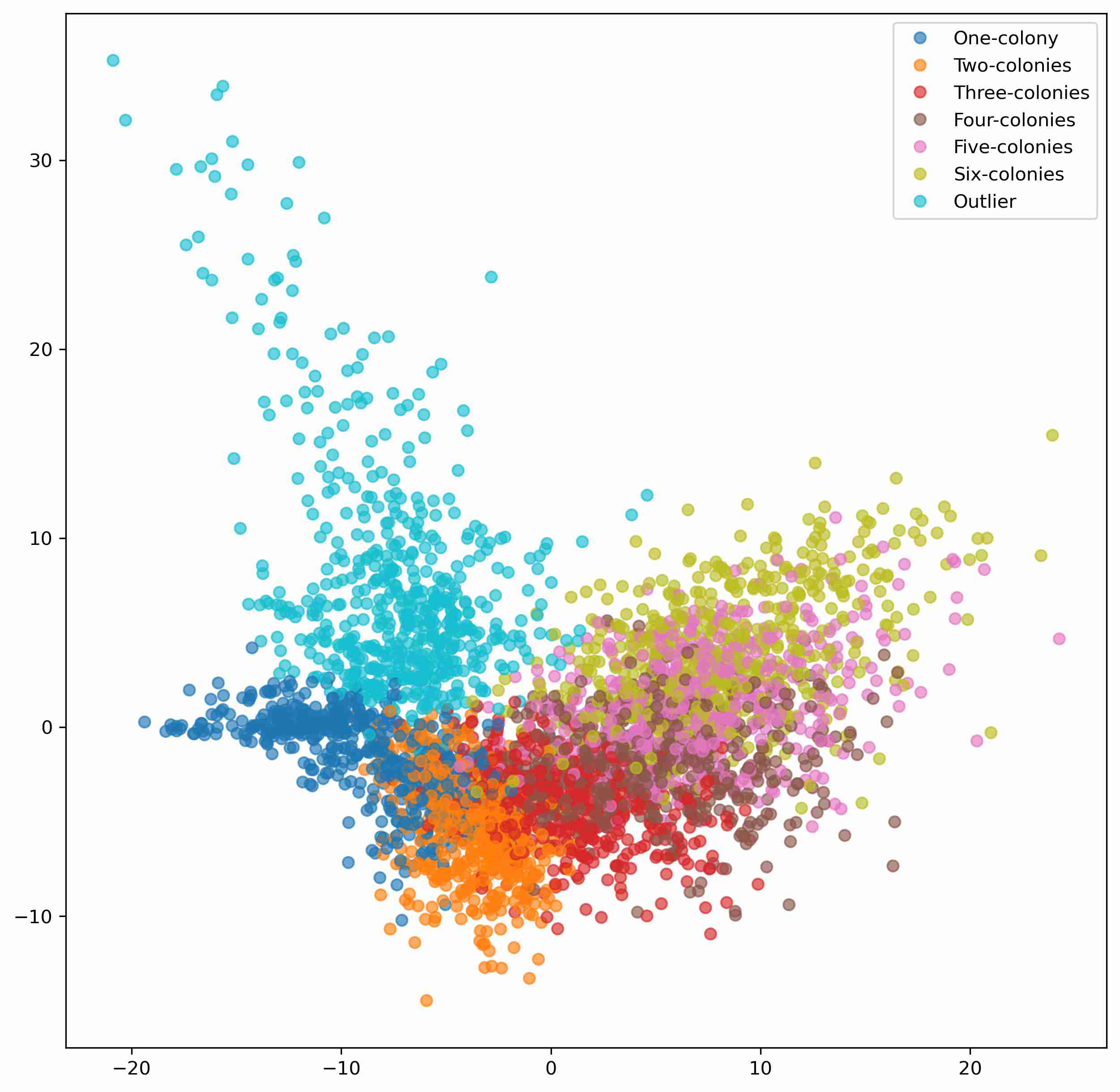}
\includegraphics[width=.45\linewidth]{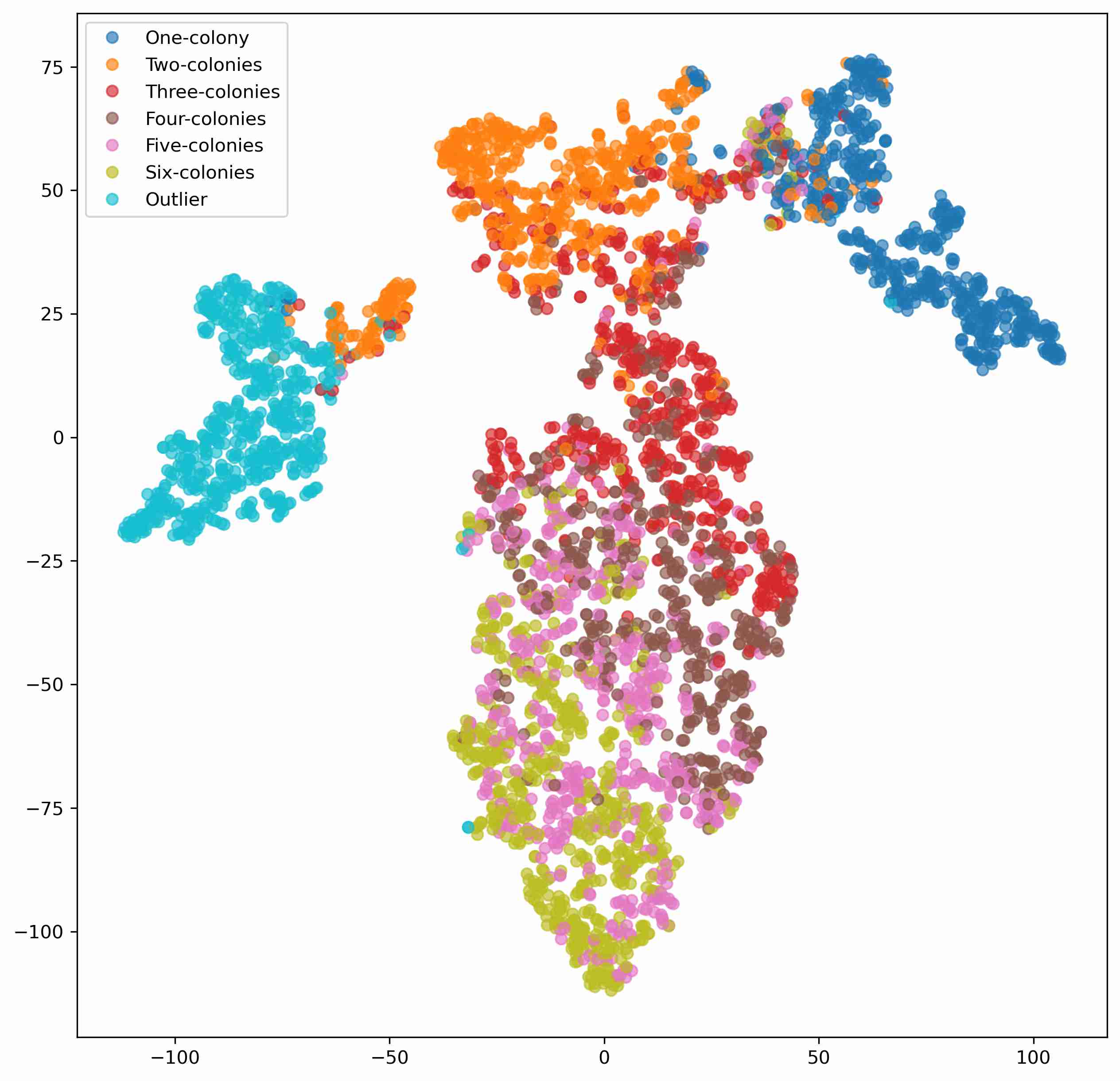}
\caption{PCA-reduced (left) and t-SNE-reduced (right) representations of the last network layer outputs from the balanced MicrobiaNet model on the MicrobiaS1B1 training set. Each colour corresponds to one of the seven classes in the dataset. Some classes form distinct clusters (blue, orange, and cyan regions), whereas others (red, brown, pink, and olive regions) are entangled and overlapped extensively.}
\label{fig:visualisation_of_network_layers_with_pca_training_v440}
\end{figure}

\begin{figure}[h!]
\centering
\includegraphics[width=.45\linewidth]{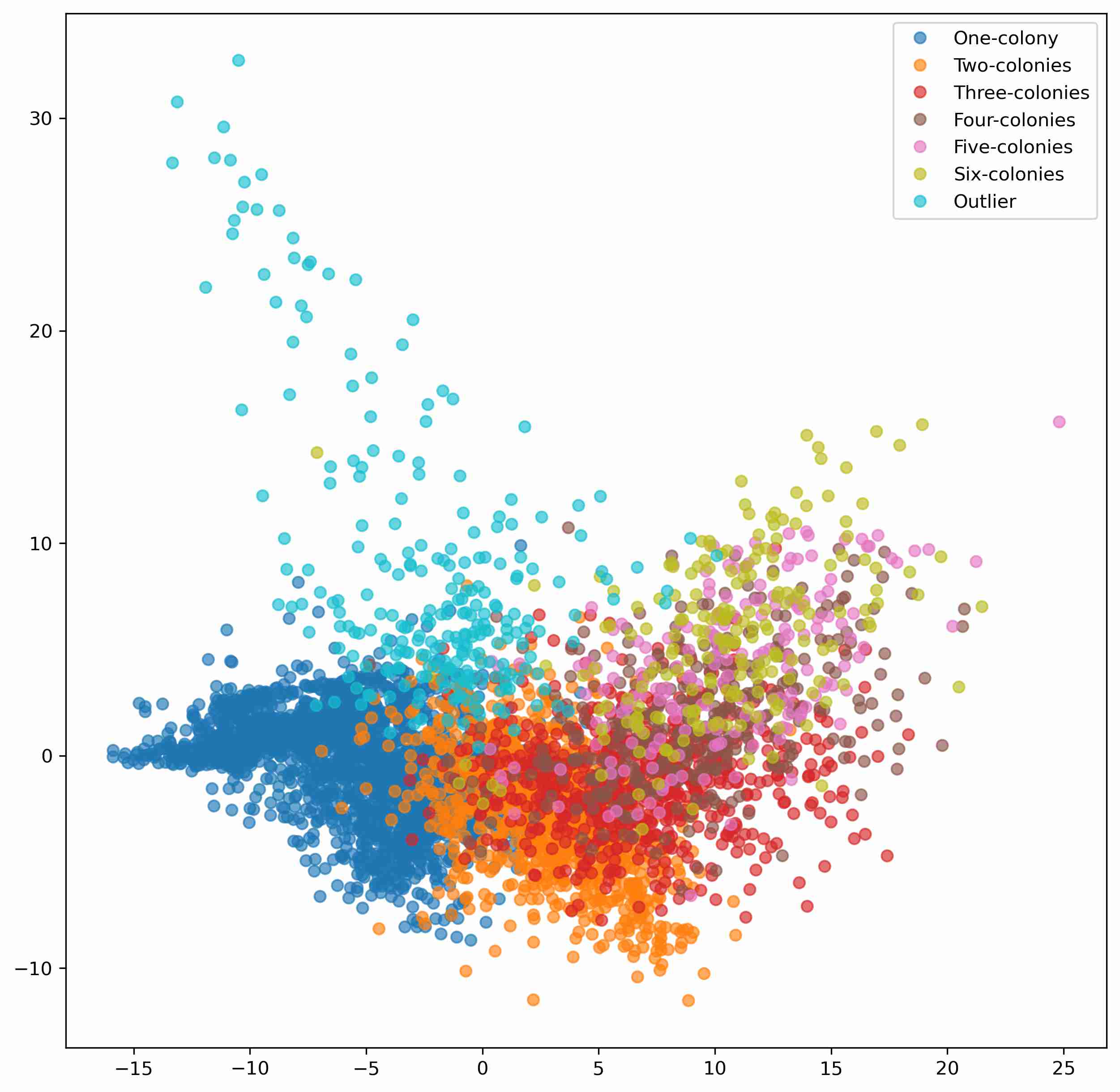}
\includegraphics[width=.45\linewidth]{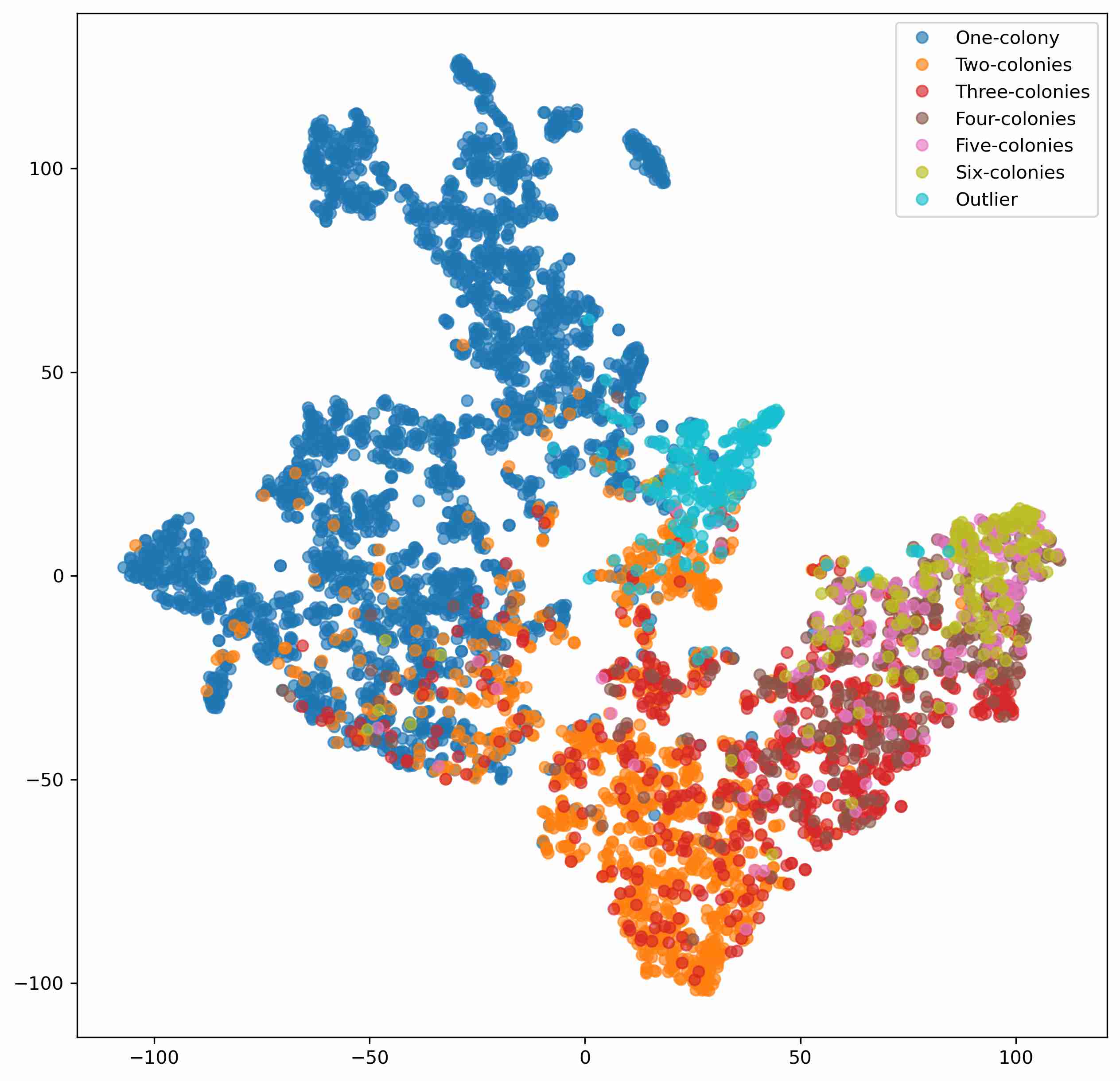}
\caption{Same as Figure \ref{fig:visualisation_of_network_layers_with_pca_training_v440}, but for the MicrobiaS1(B1) validation set.}
\label{fig:visualisation_of_network_layers_with_pca_valid_v440}
\end{figure}

Figure \ref{fig:visualisation_of_network_layers_with_pca_training_v440} demonstrates the network layer outputs from the balanced MicrobiaS1B1 training set. One-colony, Two-colonies, and Outlier are distinct with their own clusters regardless of the dimensionality reduction method. In contrast, all other classes are entangled and overlapped extensively, almost forming a single cluster. The same pattern is observed from the network layer outputs obtained from the MicrobiaS1(B1) validation sets in Figure \ref{fig:visualisation_of_network_layers_with_pca_valid_v440}. As can be seen, these colonies occupy an area to the lower-right of the figure, with the Two-colonies (orange) region to their lower-left and the One-colony (blue) and Outlier (cyan) regions further away. This new grouping consolidates the colonies with the highest visual similarity, as shown in the figure.

Because the model is trained on a balanced dataset, these patterns further confirm that the class imbalance is not a major factor limiting MicrobiaNet's performance. Instead, the model’s primary difficulty arises from the high visual similarity between neighbouring classes. Additional visualisations supporting this conclusion are provided in Appendix~\ref{appendix-impact-of-class-imbalance}.

\subsection{An attempt to address the high visual similarity}
This experiment aims to address the high visual similarity, identified as a key limitation of MicrobiaNet, by concatenating classes that exhibit strong visual resemblance. The four lowest-performing classes are combined into a single class to reduce the high inter-class similarity, as they exhibit substantial overlap in feature space. As a result, the Three-colonies, Four-colonies, Five-colonies, and Six-colonies classes are consolidated into a new class named More-colonies.

\subsubsection{Experimental setup}
Class concatenation is performed on the imbalanced MicrobiaS1 training and validation sets, resulting in a four-class dataset referred to as MicrobiaS1C1 dataset, where C stands for the concatenation. MicrobiaNet shown in Figure \ref{fig:microbia_network_architecture} is used in this experiment with the final layer modified to produce four scores instead of seven. The training and evaluation follow the same procedures explained in \S \ref{subsubsec_training_and_eval}. Additionally, the model interpretation is performed by visualising network layer outputs, as outlined in \S \ref{subsubsec-network-layer-out-vis}.

\subsubsection{Results}
\begin{table}[h!]
       \centering
       \caption{Overall evaluation results on the MicrobiaS1C1 dataset.}
       \label{Microbia_model_results_v4000}
       \begin{tabular}{p{0.18\textwidth}>{\centering}p{0.1\textwidth}>{\centering}p{0.1\textwidth}>{\centering}p{0.15\textwidth}>{\centering\arraybackslash}p{0.18\textwidth}}
               \hline
               Data type  & Precision & Recall & F1 score & Accuracy (\%) \\ \hline
               Training   & 0.93      & 0.93   & 0.93     & 92.54         \\
               Validation & 0.92      & 0.92   & 0.92     & 91.85         \\ \hline
       \end{tabular}
\end{table}

\begin{table}[h!]
       \centering
       \caption{Classification results on the MicrobiaS1C1 training set and validation set.}
       \label{tab_cr_v4000_train}
       \begin{tabular}{p{0.02\textwidth}>{\centering}p{0.2\textwidth}>{\centering}p{0.1\textwidth}>{\centering}p{0.1\textwidth}>{\centering\arraybackslash}p{0.15\textwidth}}
               \hline
              & Class Name     & Precision & Recall & F1 score \\ \hline
             \multirow{4}{*}{\rotatebox[origin=c]{90}{Training}} & One-colony     & 0.95      & 0.98   & 0.97     \\
              & Two-colonies   & 0.83      & 0.89   & 0.86     \\
              & More-colonies  & 0.96      & 0.86   & 0.91     \\
              & Outlier        & 0.92      & 0.84   & 0.88     \\ \hline
             \multirow{4}{*}{\rotatebox[origin=c]{90}{Validation}} & One-colony     & 0.94      & 0.98   & 0.96     \\
              & Two-colonies   & 0.82      & 0.84   & 0.83     \\
              & More-colonies  & 0.94      & 0.87   & 0.90     \\
              & Outlier        & 0.91      & 0.84   & 0.87     \\ \hline
       \end{tabular}
\end{table}

\begin{figure}[h!]
\centering
\includegraphics[width=0.45\linewidth]{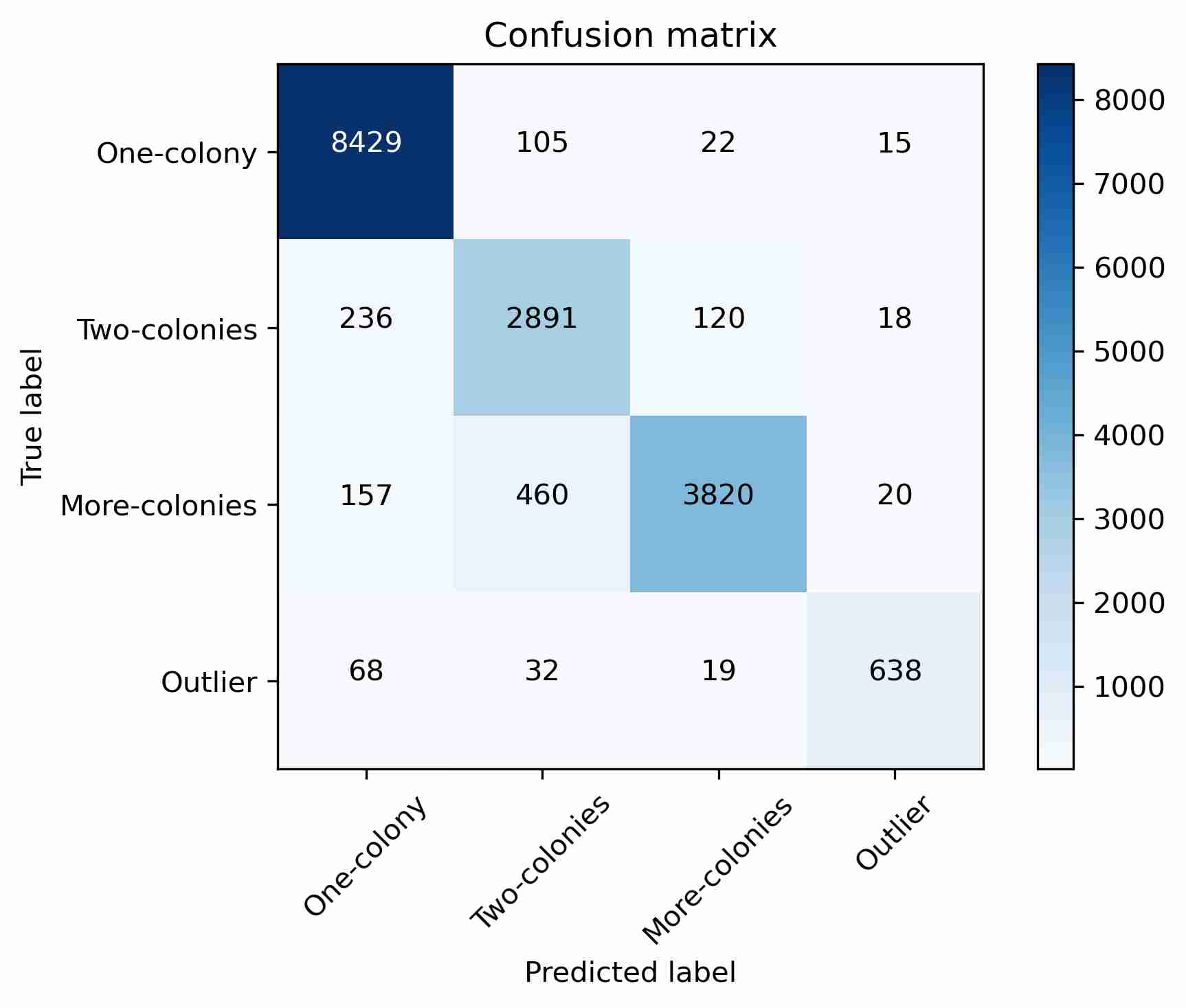}
\includegraphics[width=0.45\linewidth]{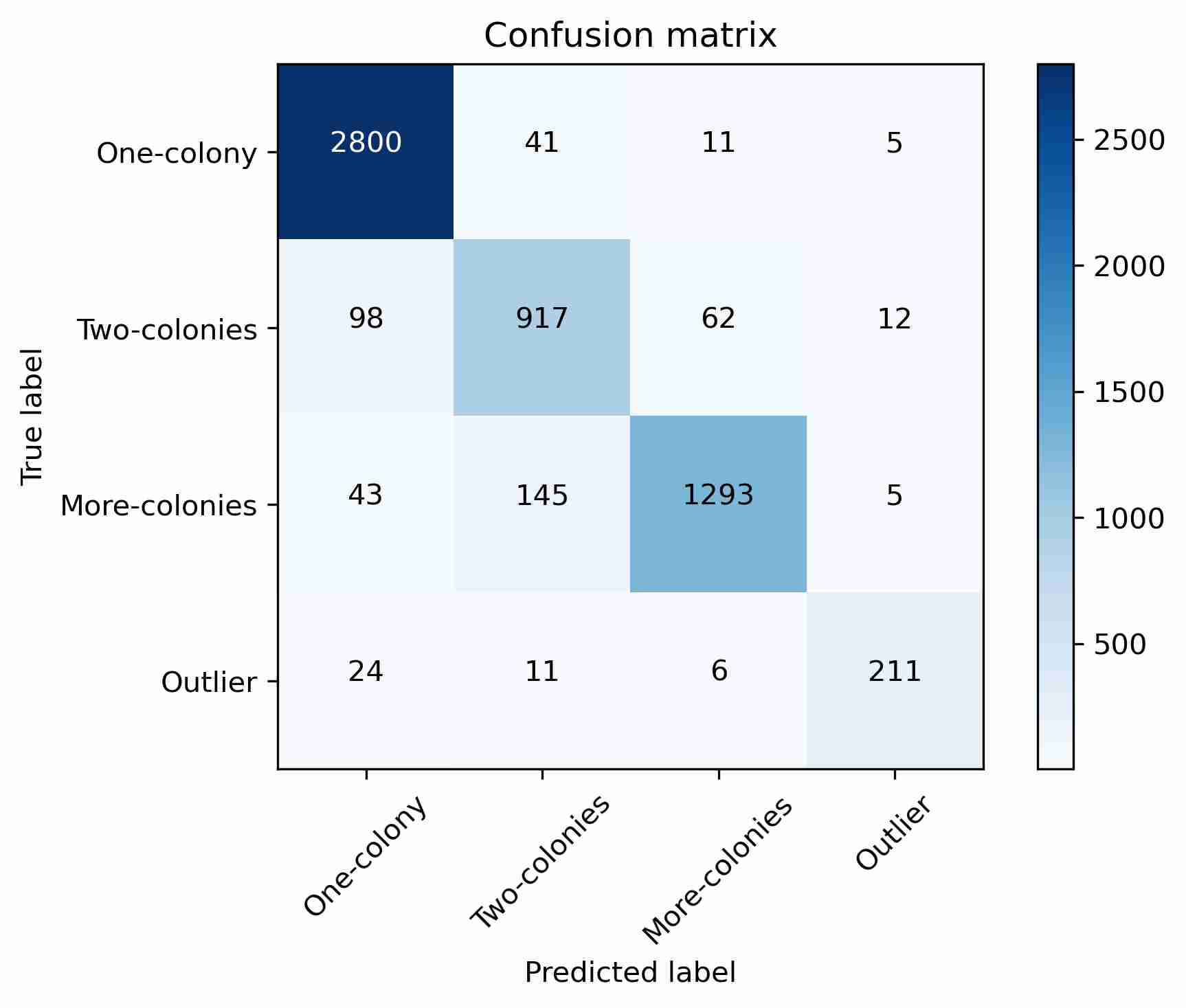}
\caption{Confusion matrices for MicrobiaS1C1 training results (left) and validation results (right). Most misclassifications occur between neighbouring classes.}
\label{fig:cm_v4000_train}
\end{figure}

The overall evaluation results on the MicrobiaS1C1 dataset are listed in Table \ref{Microbia_model_results_v4000}. The model achieves 0.93 and 0.92 in training F1 score and validation F1 score. Detailed classification metrics for both sets are presented in Table \ref{tab_cr_v4000_train}, which is computed from their corresponding confusion matrices in Figure \ref{fig:cm_v4000_train}. These results indicate the model is not biased towards the majority class because the standard deviation of training F1 scores and validation F1 scores are 0.04 and 0.05. Additionally, the wrong predictions are not dominated by the majority One-colony class as illustrated in Figure \ref{fig:cm_v4000_train}.

\begin{table}[h!]
       \centering
       \caption{MicrobiaS1 validation results after converting the baseline seven-class predictions (Table
       \ref{tab_cr_v4_valid}) into four classes.}
       \label{tab_cr_v4_lumped_valid}
       \begin{tabular}{p{0.2\textwidth}>{\centering}p{0.1\textwidth}>{\centering}p{0.1\textwidth}>{\centering\arraybackslash}p{0.15\textwidth}}
               \hline
               Class Name     & Precision & Recall & F1 score \\ \hline
               One-colony     & 0.94      & 0.98  & 0.96     \\
               Two-colonies   & 0.82      & 0.81  & 0.82     \\
               More-colonies  & 0.93      & 0.86  & 0.89     \\
               Outlier        & 0.86      & 0.80  & 0.83     \\ \hline
       \end{tabular}
\end{table}

\begin{figure}[h!]
       \centering
       \includegraphics[width=0.45\linewidth]{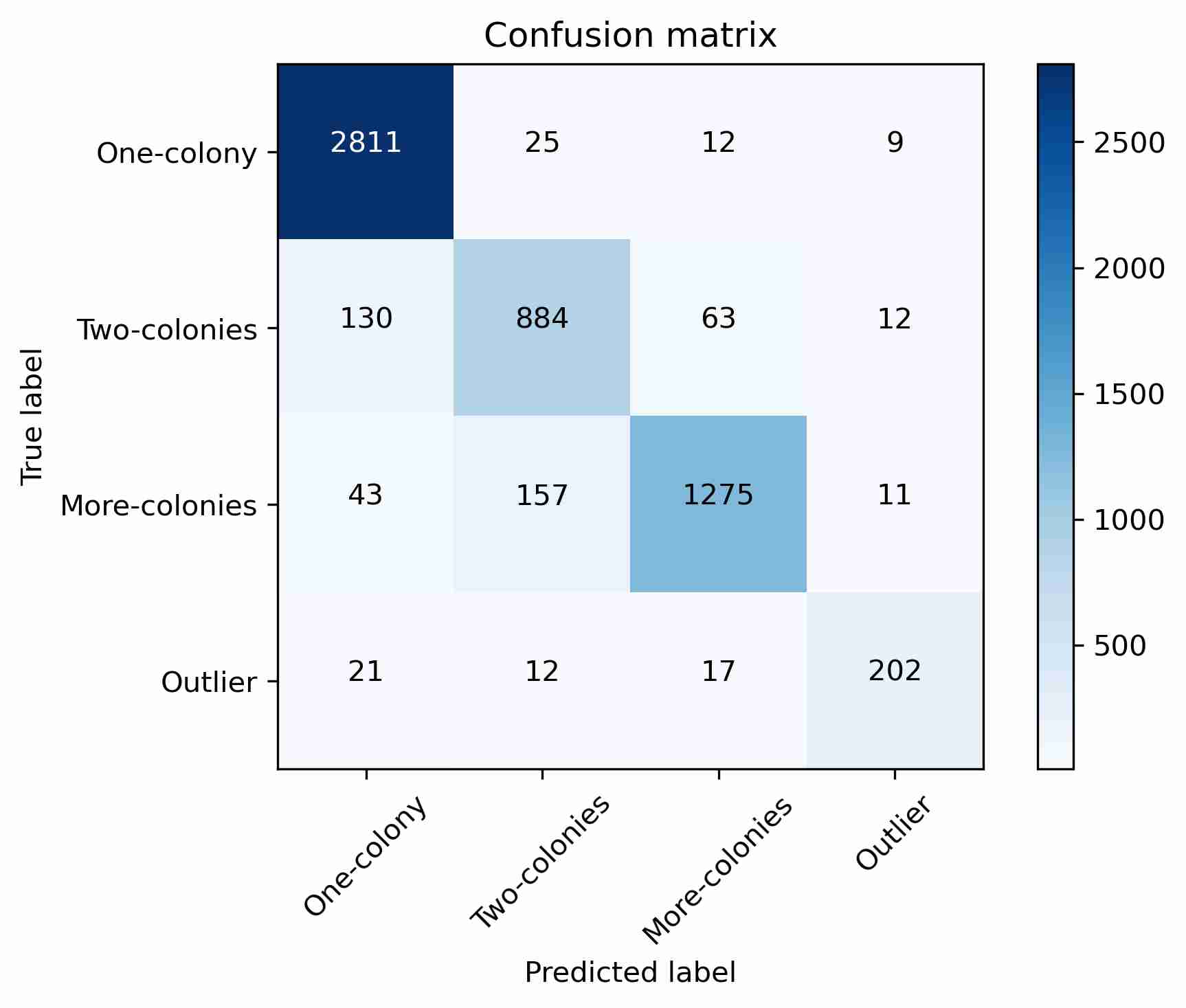}
       \caption{Four-class confusion matrix obtained by converting the baseline seven-class MicrobiaS1 validation results (Figure \ref{fig:cm_v4_valid}). Most misclassifications occur between neighbouring classes.}
       \label{fig:cm_v4_lumped_valid}
\end{figure}

Table \ref{tab_cr_v4_lumped_valid} and Figure \ref{fig:cm_v4_lumped_valid} show the detailed classification results after converting the baseline validation results (Table \ref{tab_cr_v4_valid} and Figure \ref{fig:cm_v4_valid}) to four classes. Its overall validation precision, recall, F1 score, and accuracy are 0.91, 0.91, 0.91, and 82.79\%. The MicrobiaS1C1 validation F1 score is 0.01 higher than that of the baseline validation. This observation suggests that the baseline model was already treating Three-colonies, Four-colonies, Five-colonies, and Six-colonies classes as visually indistinguishable, and the merging of these classes aligns the labels with the learned representation. This is consistent with the feature-space plots in Figures \ref{fig:visualisation_of_network_layers_with_pca_training} and \ref{fig:visualisation_of_network_layers_with_pca_valid}, which show that samples from these classes form overlapping clusters. Therefore, merging the classes slightly improves the metrics without fundamentally changing the underlying structure. This marginal improvement further corroborates that the dominant limiting factor of MicrobiaNet is the high visual similarity across classes, rather than class imbalance.

\begin{figure}[h!]
\centering
\includegraphics[width=.45\linewidth]{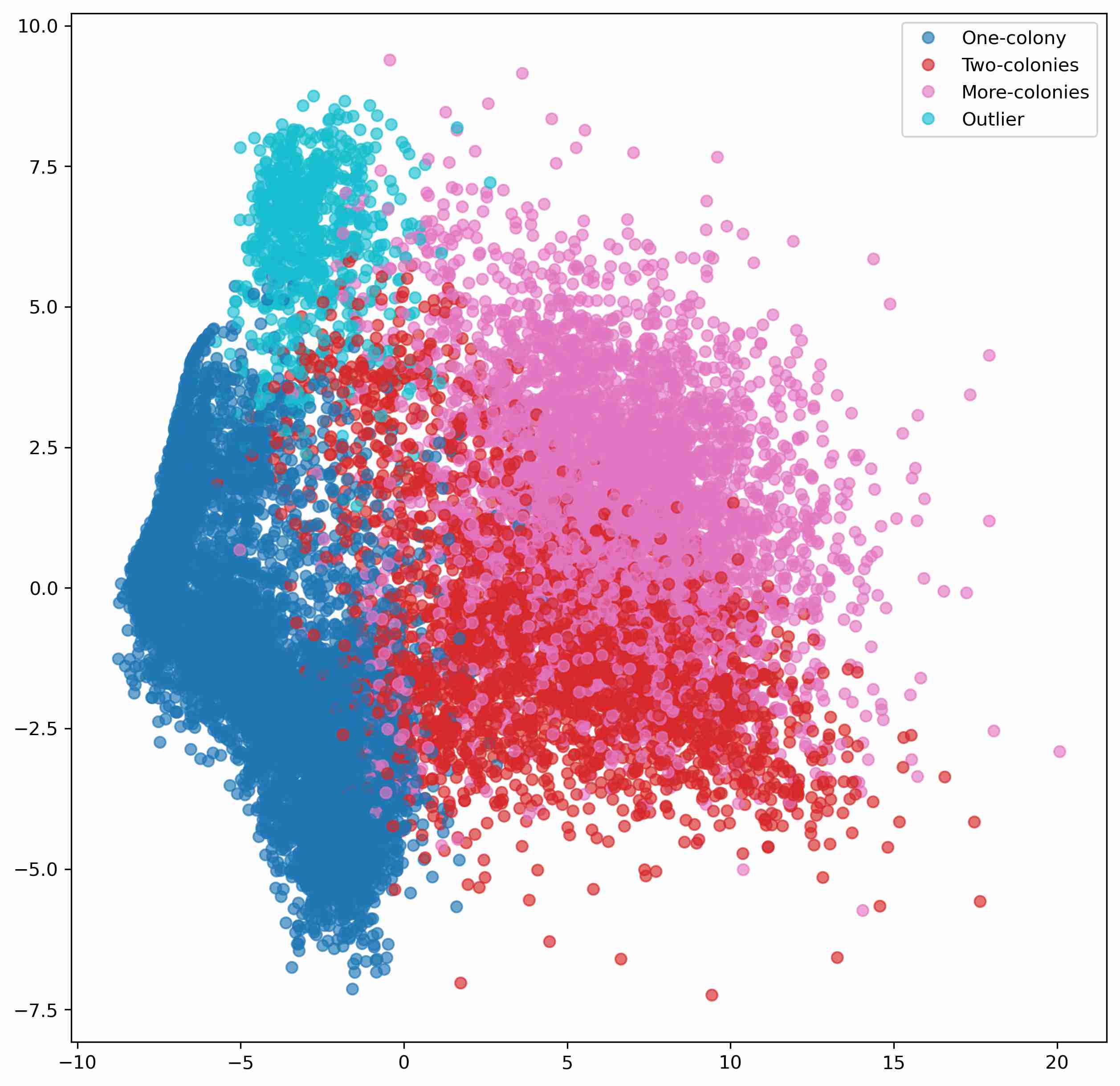}
\includegraphics[width=.45\linewidth]{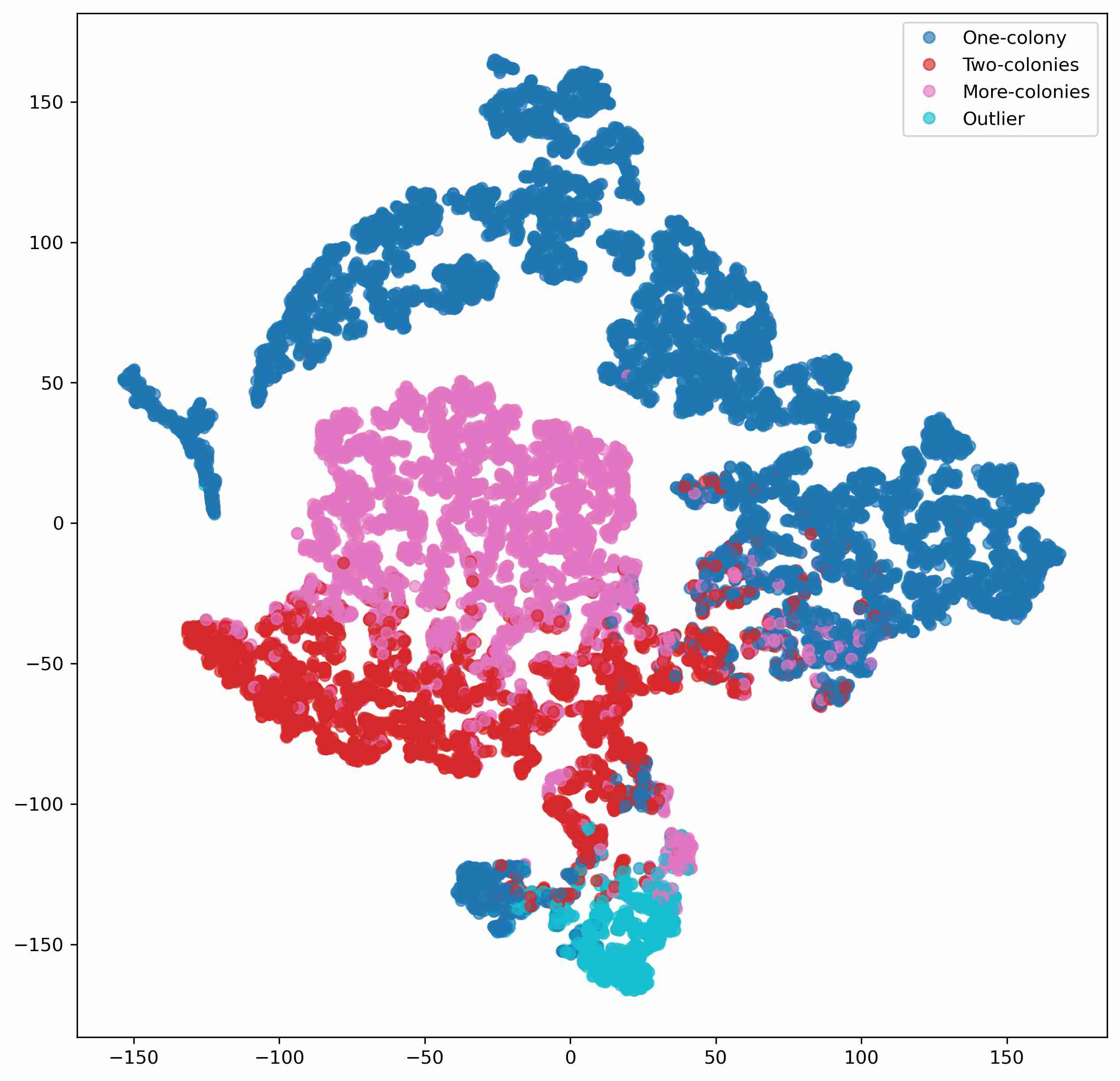}
\caption{PCA-reduced (left) and t-SNE-reduced (right) representations of the last network layer outputs from the model on the MicrobiaS1C1 training set. Each colour corresponds to one of the four classes in the dataset. Compared to the feature space of seven classes, they now form clusters that are more discriminative.}
\label{fig:visualisation_of_network_layers_with_pca_training_v4000}
\end{figure}

\begin{figure}[h!]
\centering
\includegraphics[width=.45\linewidth]{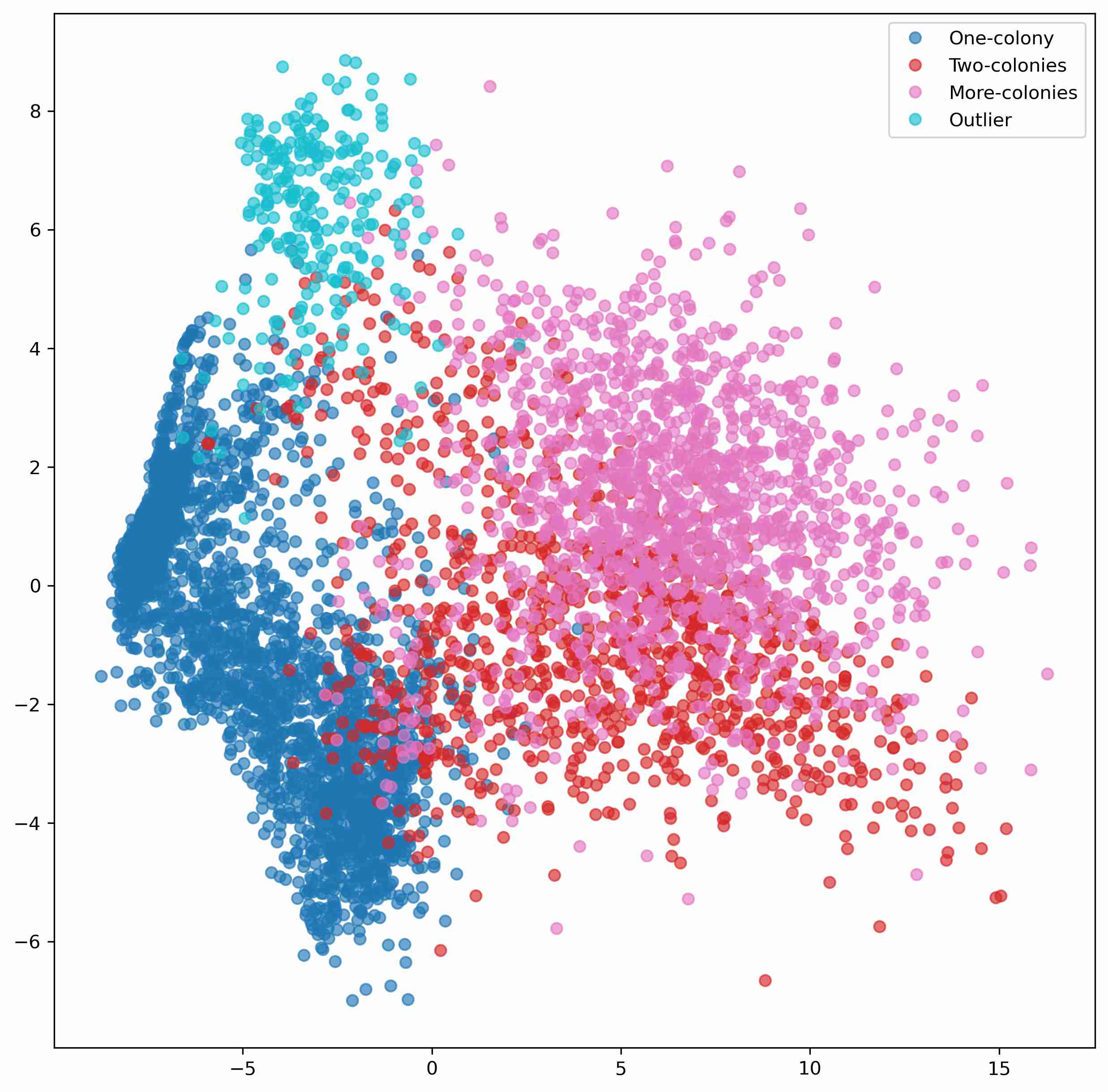}
\includegraphics[width=.45\linewidth]{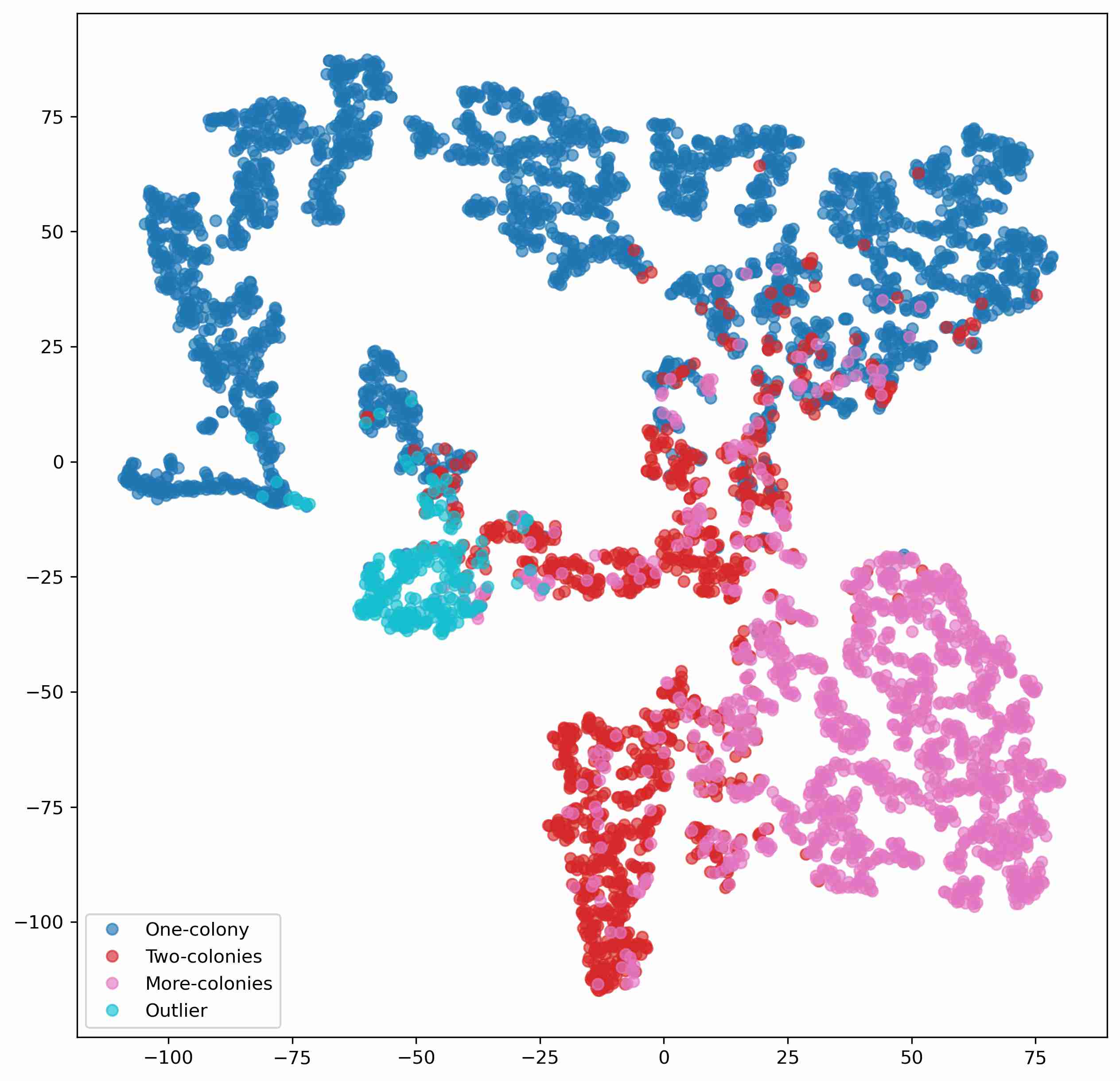}
\caption{Same as Figure \ref{fig:visualisation_of_network_layers_with_pca_training_v4000}, but for the validation set.}
\label{fig:visualisation_of_network_layers_with_pca_valid_v4000}
\end{figure}

Merging classes results in more discriminative clusters in the feature space. In the training set, comparisons of PCA and t-SNE plots for the original and merged classes (Figures \ref{fig:visualisation_of_network_layers_with_pca_training} and \ref{fig:visualisation_of_network_layers_with_pca_training_v4000}) show that the merged More-colonies class effectively replaces the four previously entangled classes. As a result, these classes form a single distinct cluster. Similar improvements in class separability are also observed in the validation set (Figures \ref{fig:visualisation_of_network_layers_with_pca_valid} and \ref{fig:visualisation_of_network_layers_with_pca_valid_v4000}). This finding underscores the potential benefits of directly addressing the high visual similarity across classes. Future work could explore algorithmic strategies specifically designed to better address this factor, potentially enabling continued improvements in colony cardinality classification.

Several alternative strategies were investigated to enhance colony cardinality classification performance. These included increasing data diversity through augmentation of the existing training set and generating synthetic samples using Generative Adversarial Networks (GANs)~\citep{goodfellow_generative_2020}. Additional efforts focussed on improving model training by incorporating alternative loss functions, such as focal loss~\citep{lin_focal_2020} and Class-Balanced Loss~\citep{cui_classbalanced_2019}. The architecture of MicrobiaNet was also fine-tuned by varying the number of layers and the model size. However, none of these approaches produced measurable improvements in counting performance. Consequently, they are not included in the main results of this paper. Future research may consider revisiting these strategies when exploring new methods for performance improvement.

\section{Conclusions}
\label{sec-conclusions}
This work has addressed an important research gap in colony counting: the explainability of the best-performing cardinality classification method for colony counting, MicrobiaNet, has not been studied. We establish a baseline performance assessment which shows that MicrobiaNet has difficulty distinguishing colonies of three or more individuals, as most errors occur between neighbouring classes. The baseline model is investigated with XAI approaches that focus on post-model explainability: network layer output, feature, and class activation map visualisations.

Our experimental results demonstrate that XAI can diagnose how data properties constrain cardinality classification performance, highlighting its potential as a general diagnostic tool beyond post hoc interpretability. MicrobiaNet’s primary limitation arises from high visual similarity across classes, as network layer output visualisations reveal that features for colonies of three or more individuals are highly entangled in feature space. However, feature and class activation map visualisations offer limited insight into MicrobiaNet’s decision-making, likely due to their reliance on strong inter-class differences, distinctive class-discriminative features, and specific model architectures. Further research is required to evaluate the applicability of these approaches to medical images.

In addition, our experimental results show that class imbalance has a minimal impact on MicrobiaNet, revising the assertion made by its authors that mitigating class imbalance would improve performance. Data downsampling reduces the validation F1 score by only 0.06, while merging visually similar classes and adapting the architecture for four-class classification improves the validation F1 score by 0.01. Network layer output visualisations justify merging these classes since they almost form a single cluster in feature space. The marginal performance gain reinforces that high inter-class visual similarity is the primary limiting factor. Because CNNs rely on discriminative features for image classification, their performance degrades when classes lack visual separability. Collectively, the findings suggest that MicrobiaNet is constrained by a combination of high inter-class similarity, rather than class imbalance, and its intrinsic characteristics. Further investigation is needed to confirm the specific contribution of its intrinsic characteristics derived from CNNs.

Our analysis has potential wider impacts on applications of CNN classifiers, particularly in scenarios complicated by imbalanced class distributions. The challenges and insights highlighted in this work are relevant to a wide range of imaging-based classification tasks. These include applications in medical imaging (e.g. histopathology slide classification), environmental monitoring (e.g. marine plankton classification), and other biological image analyses (e.g. microscopy-based cell phenotyping). In such settings, visual distinctions are similarly subtle.

\section{Limitations and future work}
\label{sec-limitations-future-work}
Although this study demonstrates promising performance on Microbia Dataset, generalisation to real laboratory settings may be limited by dataset-specific characteristics. Microbia Dataset contains images captured under controlled conditions, whereas real laboratory settings often introduce variability in illumination, shadows, camera angle, plate type, and imaging equipment. For example, uneven lighting or differences in agar appearance may obscure colony boundaries and degrade performance. Another criticism may be the reliance on a single dataset. We acknowledge this limitation as Microbia Dataset is the only large labelled dataset to support colony cardinality classification.

Future work could train and evaluate MicrobiaNet on additional datasets with greater diversity in imaging conditions and colony morphologies. For example, AGAR~\citep{majchrowska_agar_2021} includes five microbial species as single and mixed cultures on agar plates from different liquid dilutions, despite not having colony cardinality labels. Additionally, future work could focus on developing models that explicitly account for the high visual similarity across classes. In particular, density-estimation-based counting methods may provide a promising direction~\citep{Lempitsky2010}, such as FamNet~\citep{ranjan_learning_2021}. These methods avoid the difficulty of detecting and counting clustered objects by directly predicting a density value that reflects the number of objects in a region. Specifically, the model learns to map an input image to a density map whose integrated (summed) density corresponds to the object count.

\section*{CRediT authorship contribution statement}
\textbf{Minghua Zheng:} Conceptualization, Methodology, Software, Formal analysis, Investigation, Writing - Original Draft, Writing - Review \& Editing, Visualization. \textbf{Na Helian:} Resources, Writing - Review \& Editing, Supervision, Project administration, Funding acquisition. \textbf{Peter C. R. Lane:} Resources, Writing - Review \& Editing, Supervision, Funding acquisition. \textbf{Yi Sun:} Resources, Writing - Review \& Editing, Supervision, Funding acquisition. \textbf{Allen Donald:} Resources, Supervision, Project administration, Funding acquisition.

\section*{Declaration of competing interest}
We have nothing to declare.

\section*{Data availability}
The data that support the findings of this study were originally made available by~\citet{Ferrari2017} at www.microbia.org/index.php/resources. Please contact them to request the data, as the website is currently inaccessible.

\section*{Acknowledgments}
This work was conducted while Minghua Zheng was affiliated with the Department of Computer Science at the University of Hertfordshire. It was partially funded by Hertfordshire Knowledge Exchange Partnership (HKEP), a joint funding provided by European Regional Development Fund (ERDF), University of Hertfordshire, and Synoptics Ltd.


\appendix
\section{Colony cardinality classification baseline performance}
\label{appendix-baseline}
This section presents more results on other Microbia datasets.
\begin{figure}[h!]
\centering
\includegraphics[width=.45\linewidth]{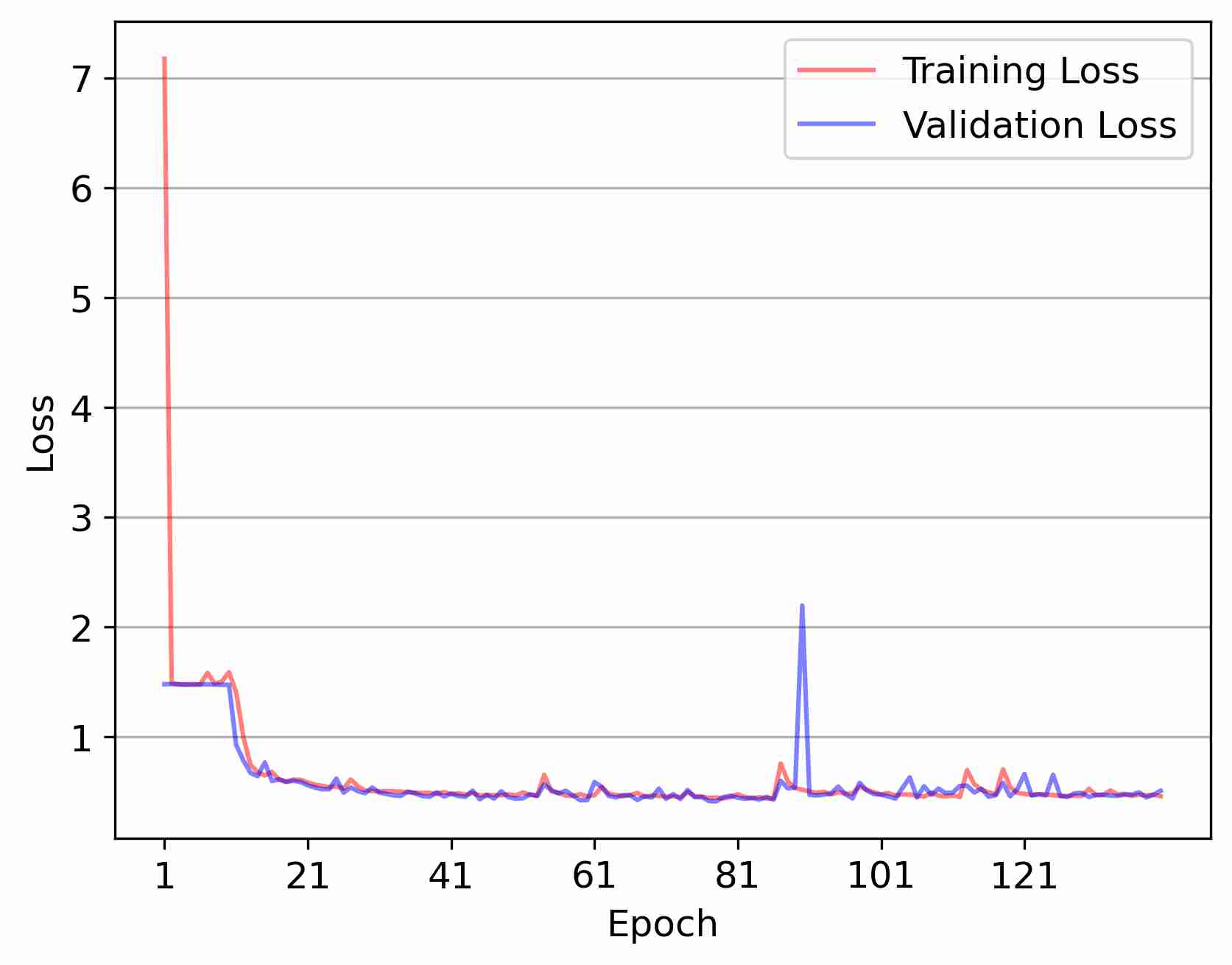}
\includegraphics[width=.45\linewidth]{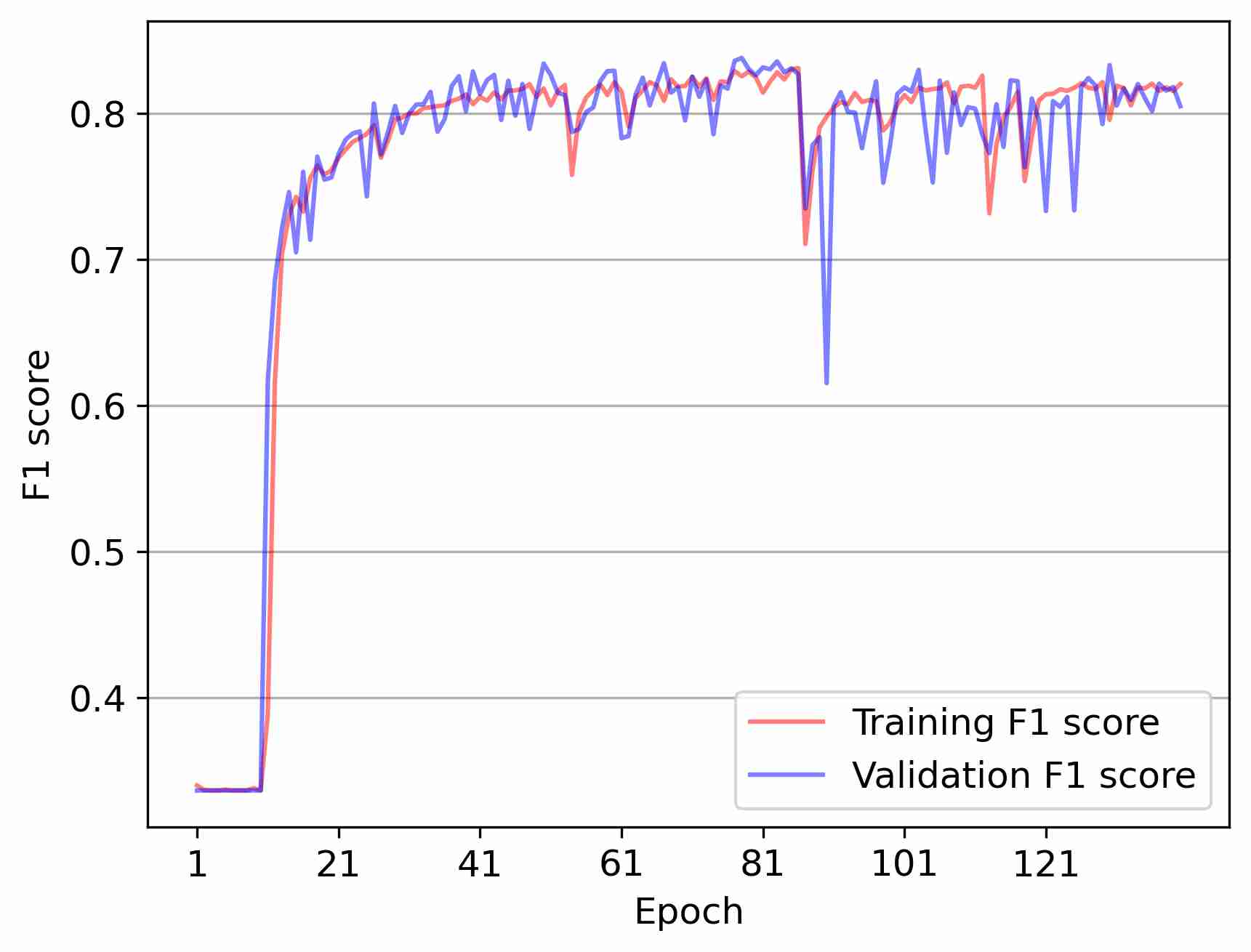}
\caption{Loss value and F1 score during training on the MicrobiaS2 dataset. The differences between training and validation loss values and F1 scores remain relatively small.}
\label{fig:curves_baseline_v41}
\end{figure}

\begin{table}[h!]
        \centering
        \caption{Classification results on the MicrobiaS2 validation set.}
        \label{tab_cr_v41_valid}
        \begin{tabular}{p{0.2\textwidth}>{\centering}p{0.1\textwidth}>{\centering}p{0.1\textwidth}>{\centering\arraybackslash}p{0.15\textwidth}}
                \hline
                Class Name     & Precision & Recall & F1 score \\ \hline
                One-colony     & 0.95      & 0.98   & 0.96     \\
                Two-colonies   & 0.84      & 0.81   & 0.83     \\
                Three-colonies & 0.66      & 0.68   & 0.67     \\
                Four-colonies  & 0.47      & 0.53   & 0.50     \\
                Five-colonies  & 0.50      & 0.14   & 0.22     \\
                Six-colonies   & 0.60      & 0.64   & 0.62     \\
                Outlier        & 0.83      & 0.90   & 0.86     \\ \hline
        \end{tabular}
\end{table}

\begin{figure}[h!]
\centering
\includegraphics[width=0.45\linewidth]{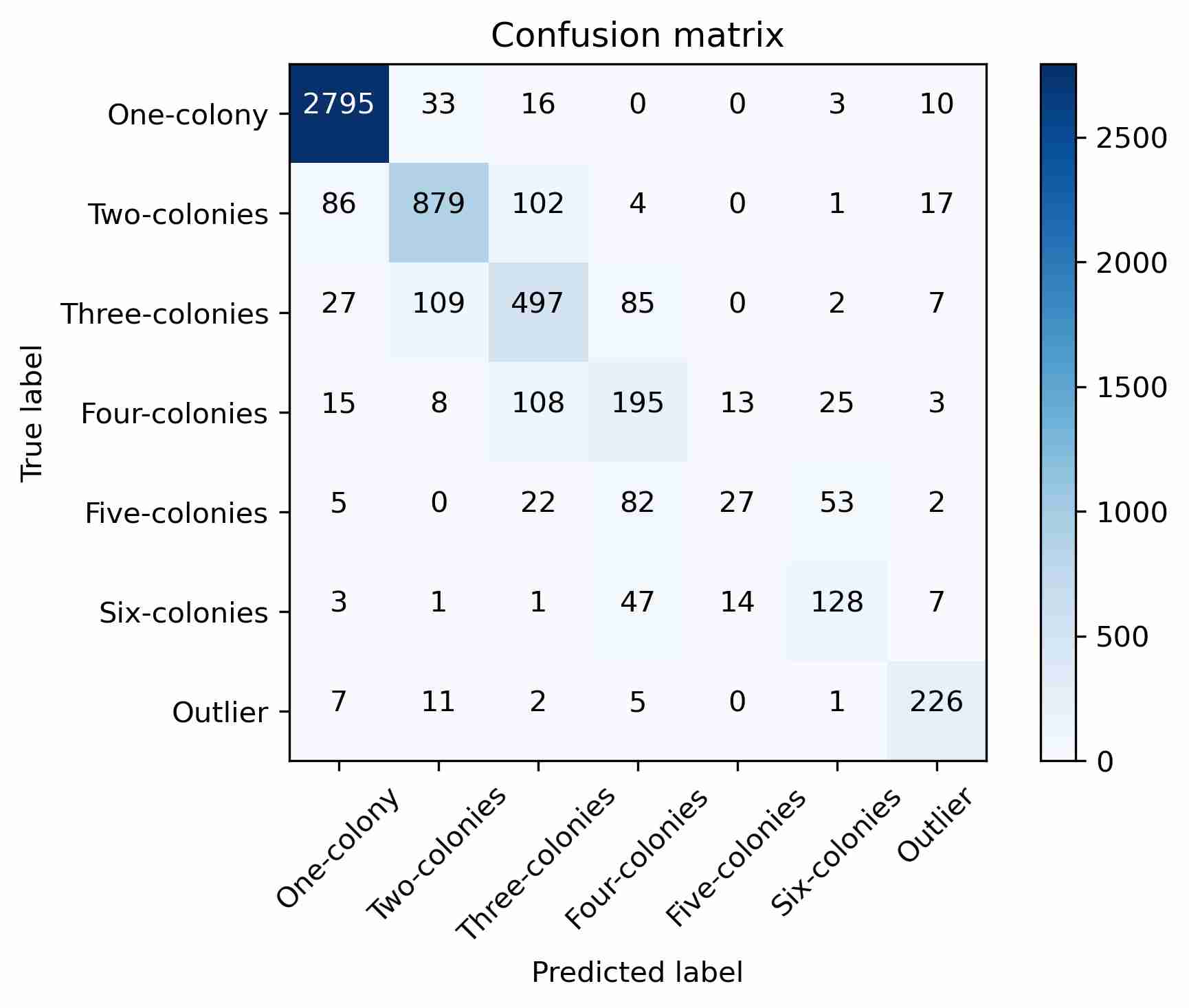}
\caption{Confusion matrix for MicrobiaS2 validation results. Most misclassifications occur between neighbouring classes.}
\label{fig:cm_v41_valid}
\end{figure}

\begin{figure}[h!]
\centering
\includegraphics[width=.45\linewidth]{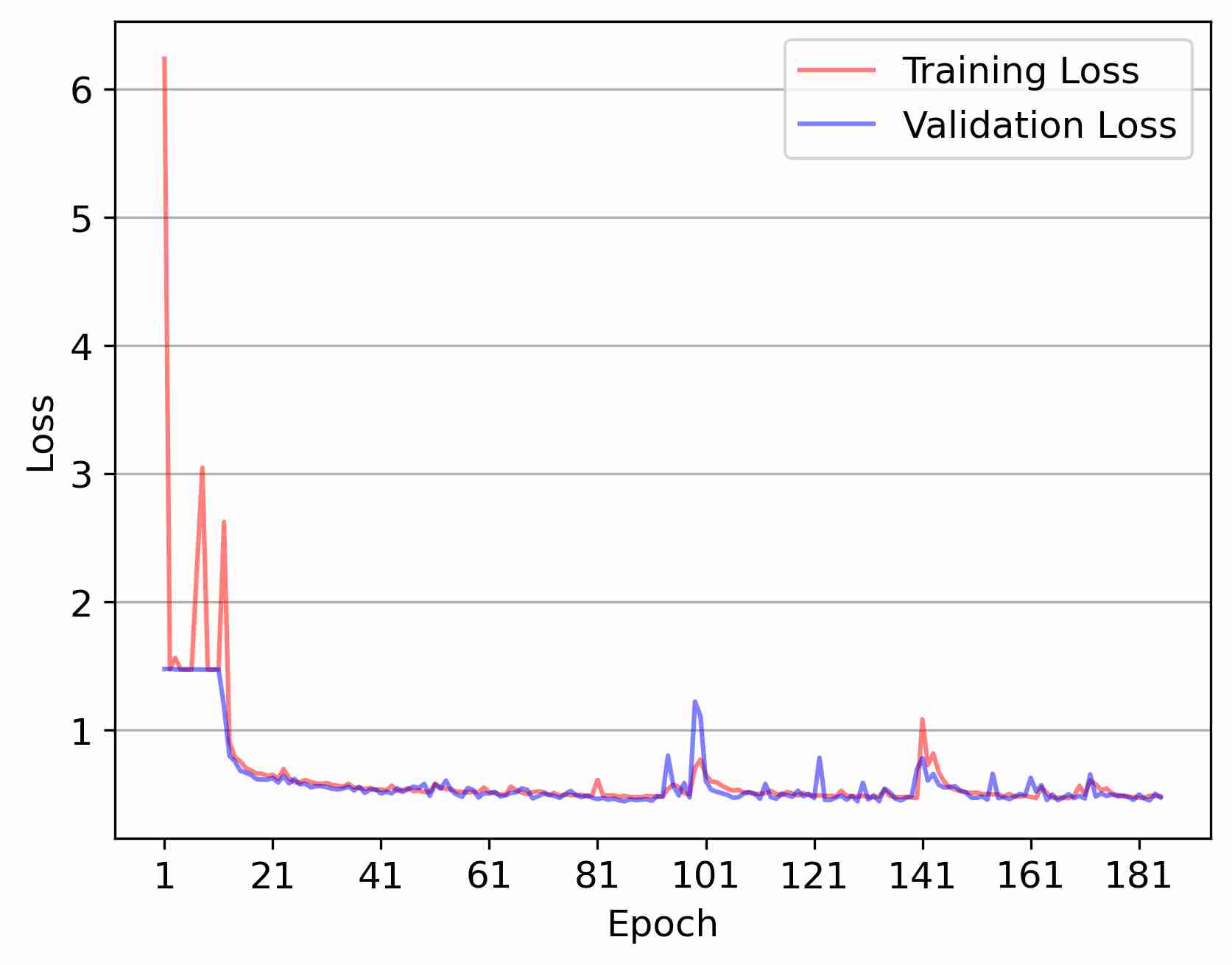}
\includegraphics[width=.45\linewidth]{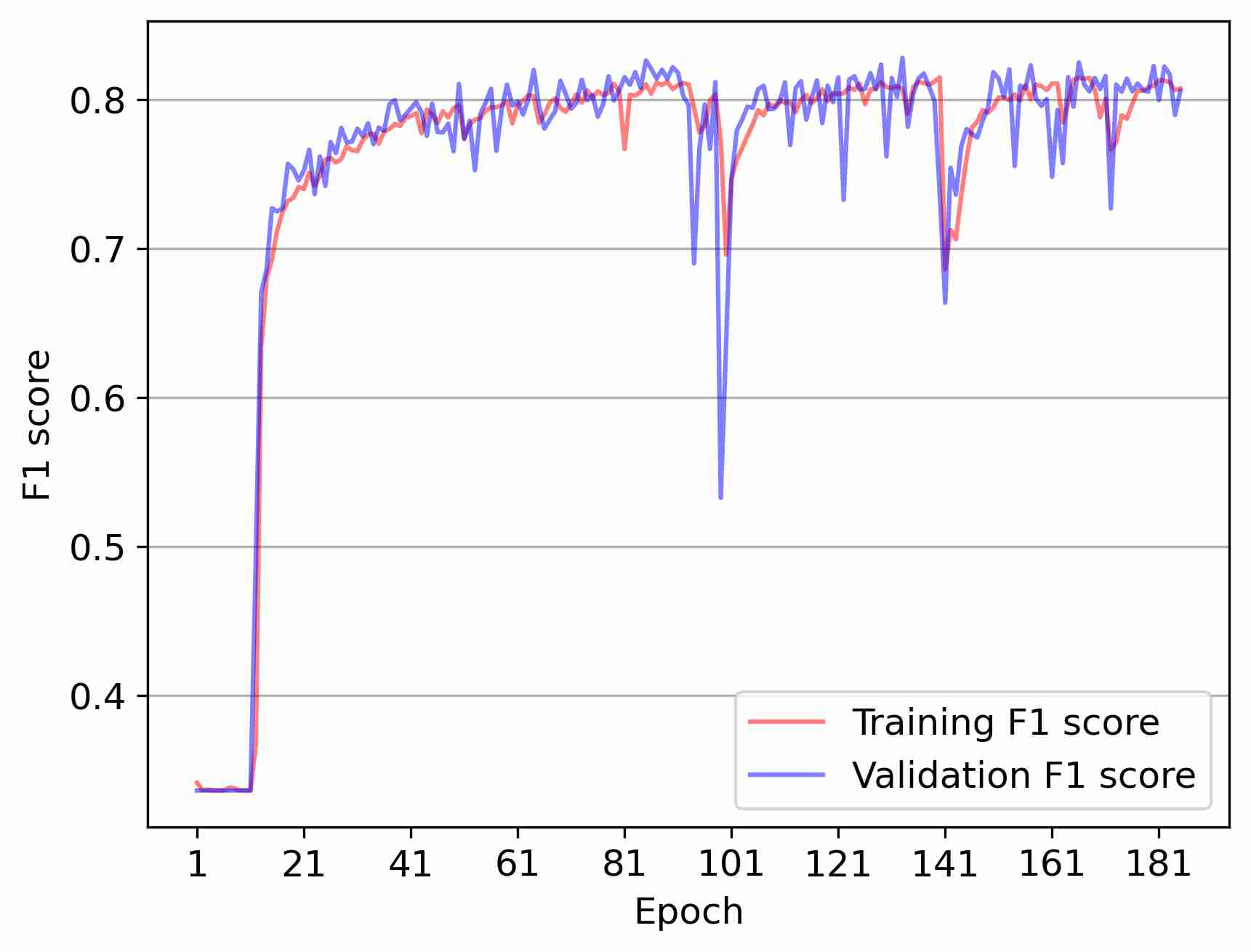}
\caption{Loss value and F1 score during training on the MicrobiaS3 dataset. The differences between training and validation loss values and F1 scores remain relatively small.}
\label{fig:curves_baseline_v42}
\end{figure}

\begin{table}[h!]
        \centering
        \caption{Classification results on the MicrobiaS3 validation set.}
        \label{tab_cr_v42_valid}
        \begin{tabular}{p{0.2\textwidth}>{\centering}p{0.1\textwidth}>{\centering}p{0.1\textwidth}>{\centering\arraybackslash}p{0.15\textwidth}}
                \hline
                Class Name     & Precision & Recall & F1 score \\ \hline
                One-colony     & 0.94      & 0.98   & 0.96     \\
                Two-colonies   & 0.82      & 0.84   & 0.83     \\
                Three-colonies & 0.70      & 0.72   & 0.71     \\
                Four-colonies  & 0.50      & 0.46   & 0.48     \\
                Five-colonies  & 0.33      & 0.31   & 0.32     \\
                Six-colonies   & 0.80      & 0.32   & 0.46     \\
                Outlier        & 0.85      & 0.80   & 0.82     \\ \hline
        \end{tabular}
\end{table}

\begin{figure}[h!]
    \centering
    \includegraphics[width=0.45\linewidth]{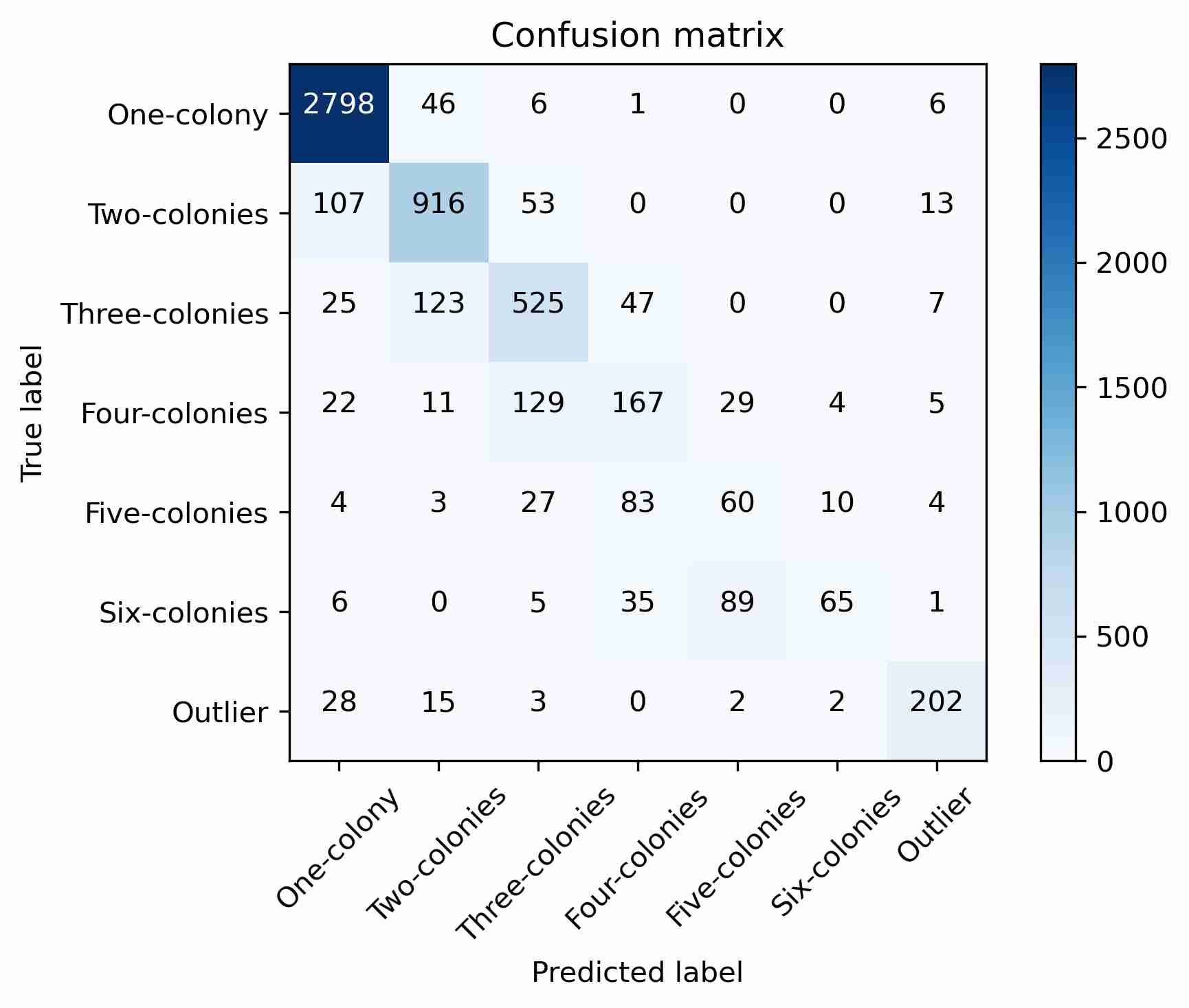}
    \caption{Confusion matrix for MicrobiaS3 validation results. Most misclassifications occur between neighbouring classes.}
    \label{fig:cm_v42_valid}
\end{figure}

\begin{figure}[h!]
\centering
\includegraphics[width=.45\linewidth]{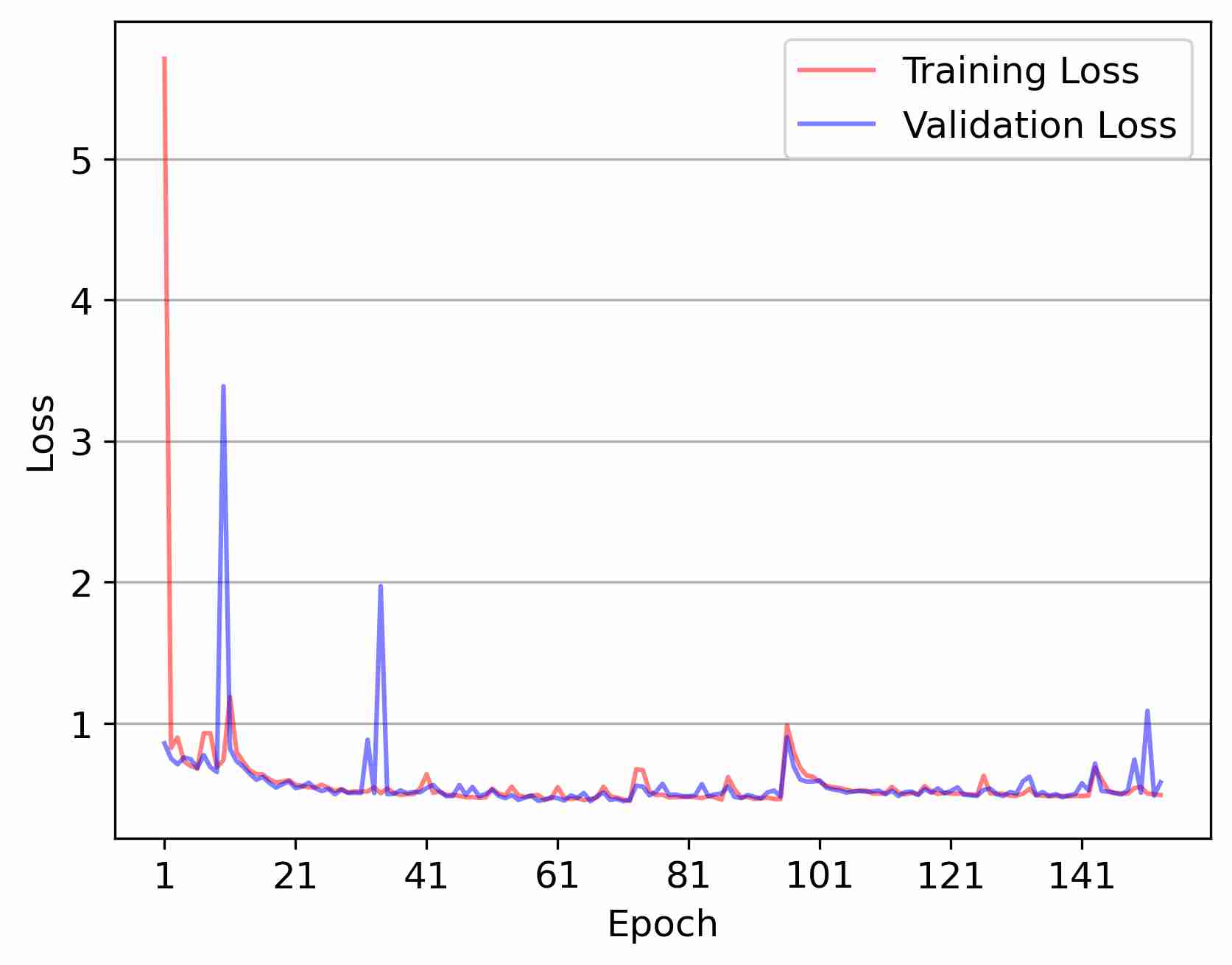}
\includegraphics[width=.45\linewidth]{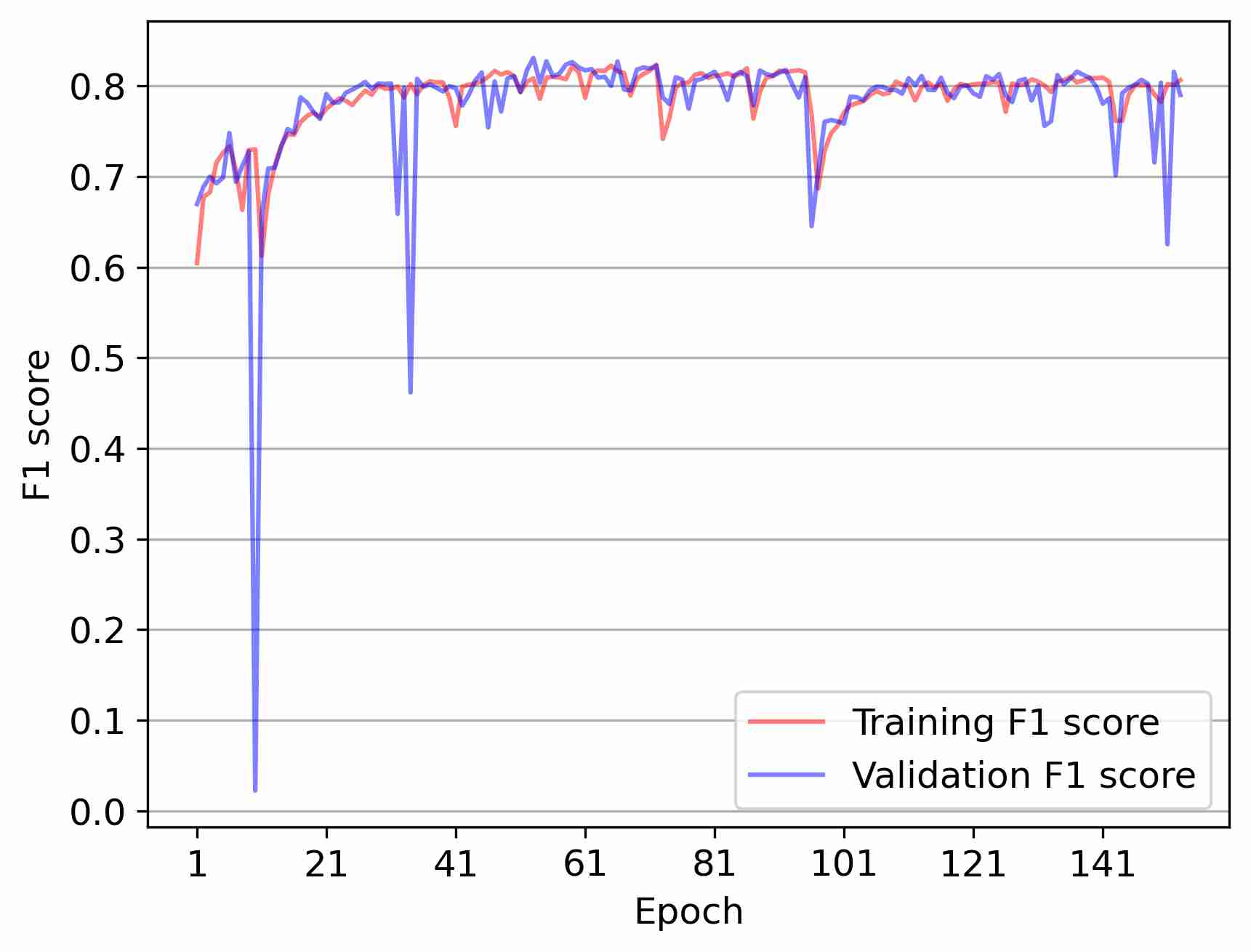}
\caption{Loss value and F1 score during training on the MicrobiaS4 dataset. The differences between training and validation loss values and F1 scores remain relatively small.}
\label{fig:curves_baseline_v43}
\end{figure}

\begin{table}[h!]
        \centering
        \caption{Classification results on the MicrobiaS4 validation set.}
        \label{tab_cr_v43_valid}
        \begin{tabular}{p{0.2\textwidth}>{\centering}p{0.1\textwidth}>{\centering}p{0.1\textwidth}>{\centering\arraybackslash}p{0.15\textwidth}}
                \hline
                Class Name     & Precision & Recall & F1 score \\ \hline
                One-colony     & 0.95      & 0.97   & 0.96     \\
                Two-colonies   & 0.83      & 0.83   & 0.83     \\
                Three-colonies & 0.71      & 0.65   & 0.68     \\
                Four-colonies  & 0.47      & 0.54   & 0.50     \\
                Five-colonies  & 0.37      & 0.24   & 0.29     \\
                Six-colonies   & 0.63      & 0.67   & 0.65     \\
                Outlier        & 0.85      & 0.80   & 0.82     \\ \hline
        \end{tabular}
\end{table}

\begin{figure}[h!]
    \centering
    \includegraphics[width=0.45\linewidth]{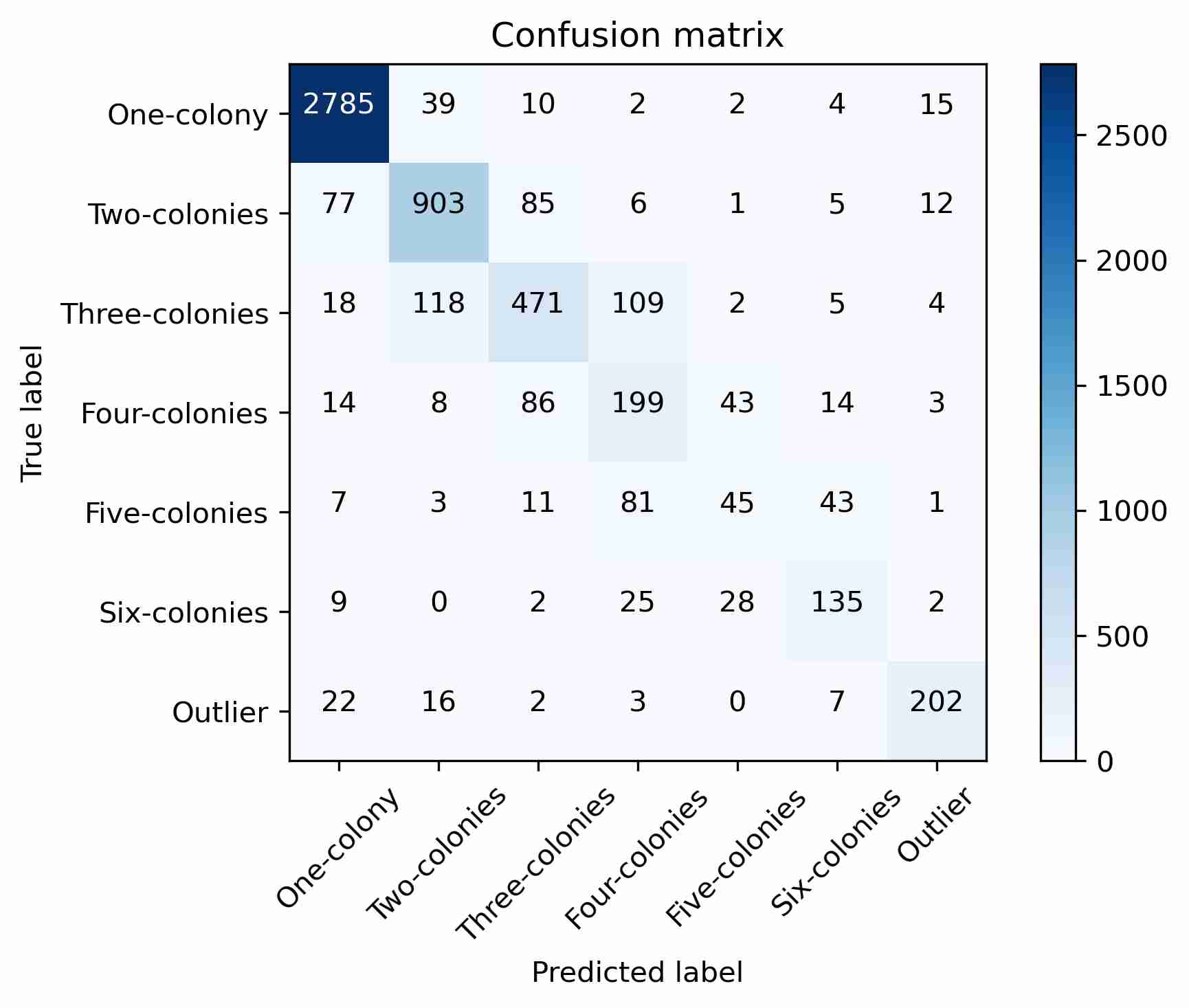}
    \caption{Confusion matrix for MicrobiaS4 validation results. Most misclassifications occur between neighbouring classes.}
    \label{fig:cm_v43_valid}
\end{figure}

\begin{figure}[h!]
\centering
\includegraphics[width=.45\linewidth]{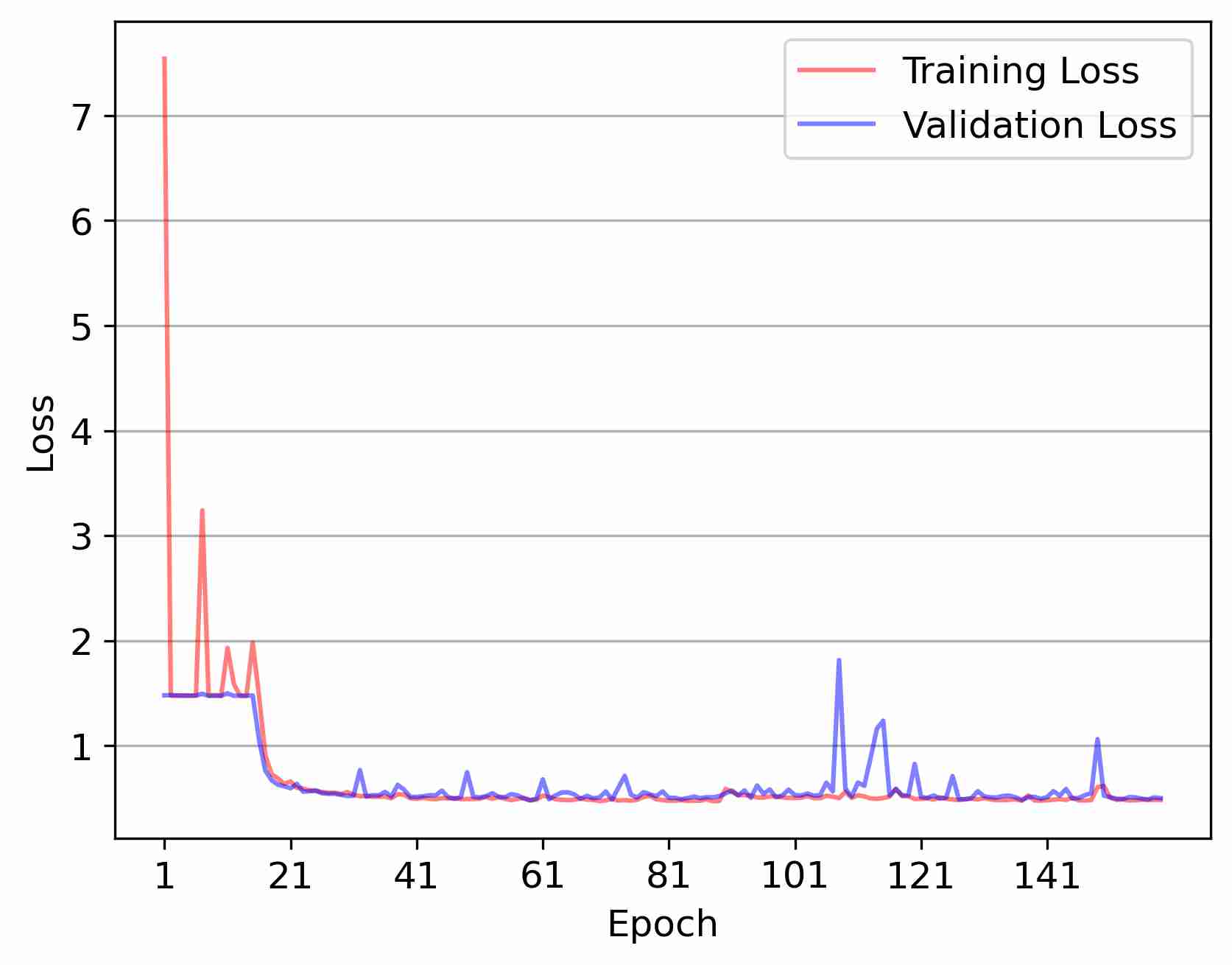}
\includegraphics[width=.45\linewidth]{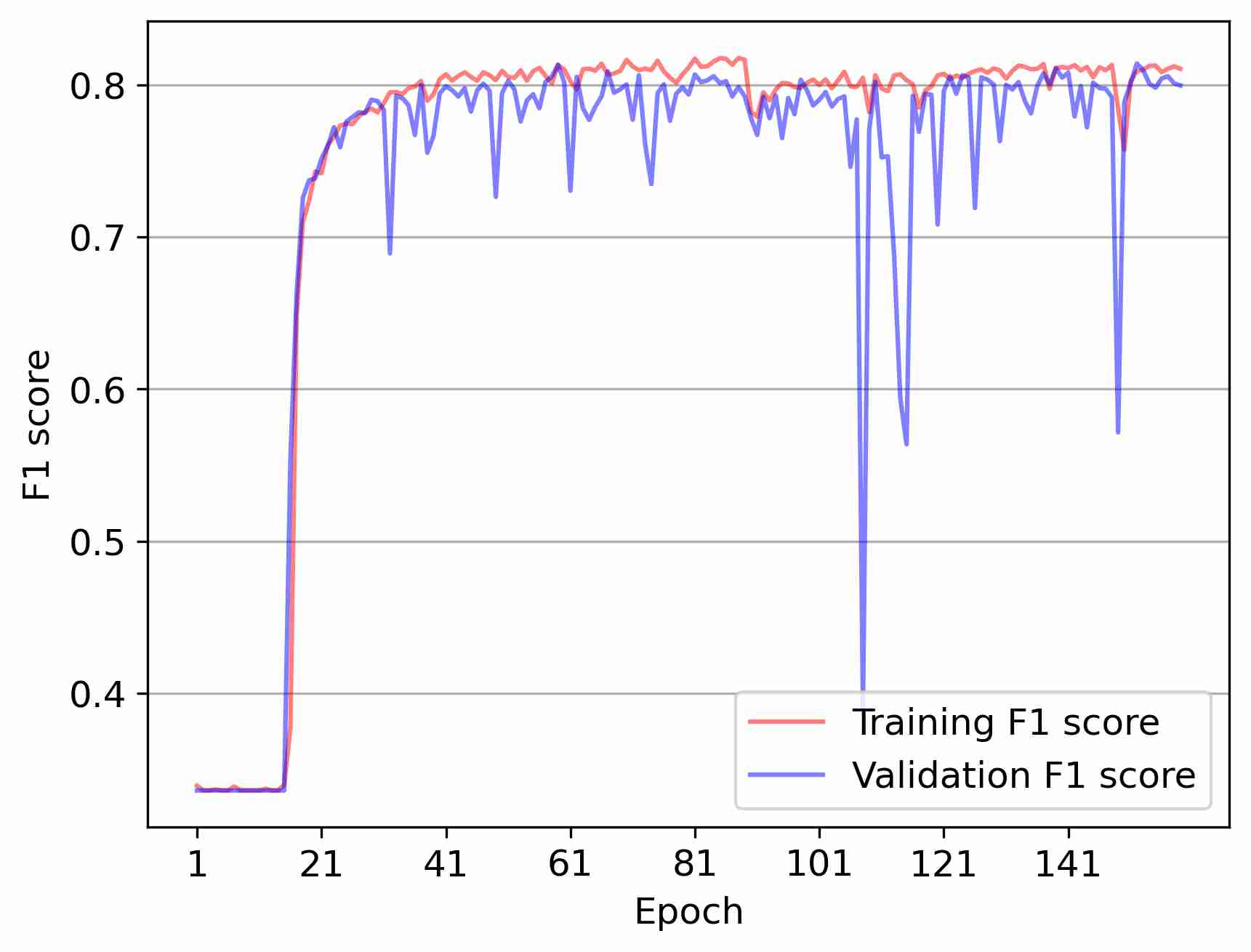}
\caption{Loss value and F1 score during training on the MicrobiaS5 dataset. The differences between training and validation loss values and F1 scores remain relatively small.}
\label{fig:curves_baseline_v44}
\end{figure}

\begin{table}[h!]
        \centering
        \caption{Classification results on the MicrobiaS5 validation set.}
        \label{tab_cr_v44_valid}
        \begin{tabular}{p{0.2\textwidth}>{\centering}p{0.1\textwidth}>{\centering}p{0.1\textwidth}>{\centering\arraybackslash}p{0.15\textwidth}}
                \hline
                Class Name     & Precision & Recall & F1 score \\ \hline
                One-colony     & 0.94      & 0.97   & 0.96     \\
                Two-colonies   & 0.80      & 0.83   & 0.81     \\
                Three-colonies & 0.60      & 0.69   & 0.64     \\
                Four-colonies  & 0.48      & 0.36   & 0.41     \\
                Five-colonies  & 0.36      & 0.23   & 0.28     \\
                Six-colonies   & 0.68      & 0.52   & 0.59     \\
                Outlier        & 0.87      & 0.82   & 0.84     \\ \hline
        \end{tabular}
\end{table}

\begin{figure}[h!]
    \centering
    \includegraphics[width=0.45\linewidth]{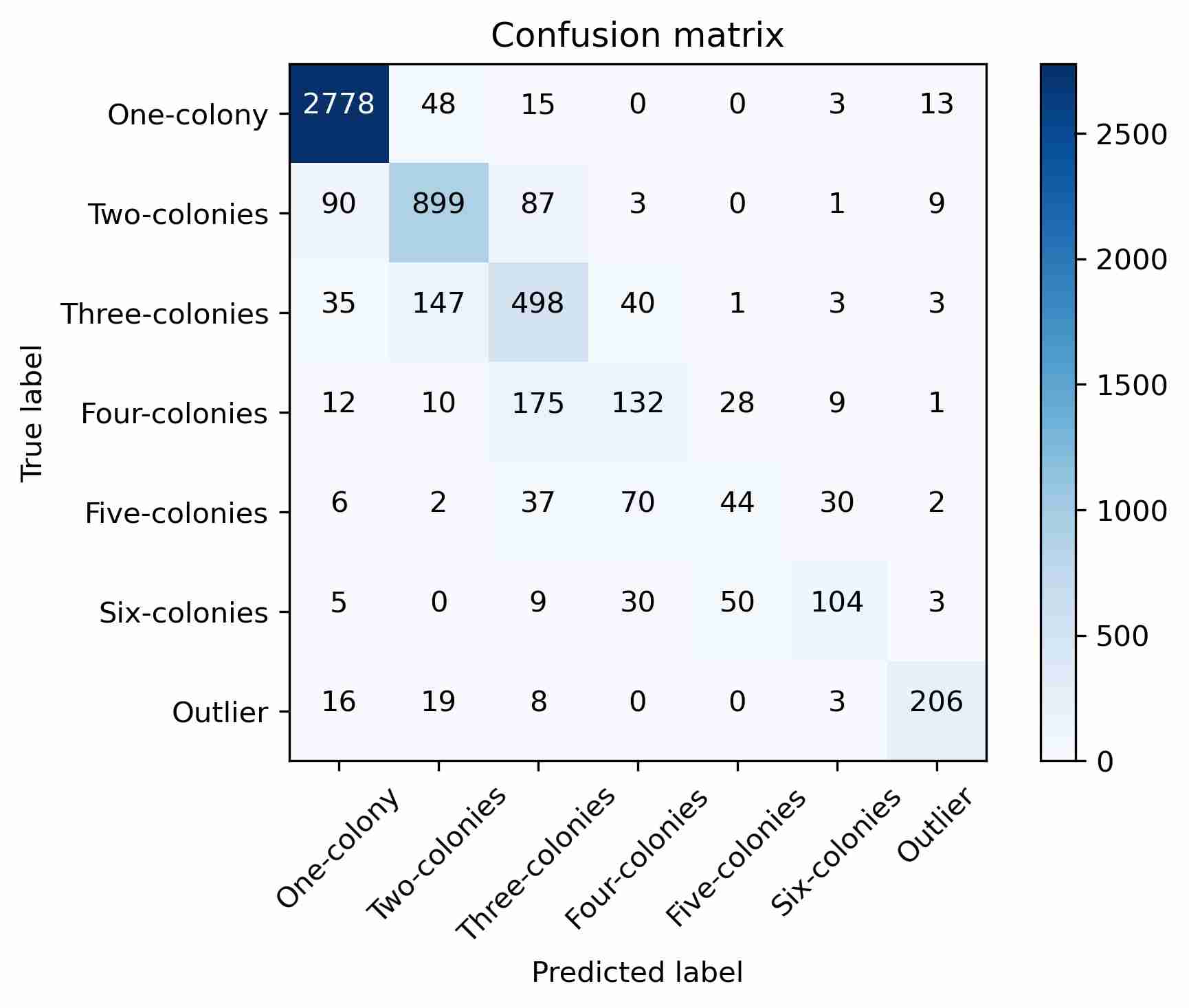}
    \caption{Confusion matrix for MicrobiaS5 validation results. Most misclassifications occur between neighbouring classes.}
    \label{fig:cm_v44_valid}
\end{figure}

\clearpage
\section{Explainability of the colony cardinality classification network}
\label{appendix-explainability}
\subsection{Visualisation of network layer outputs}
This section presents additional experimental results to visualise the penultimate network layer outputs.

\begin{figure}[h!]
\centering
\includegraphics[width=.45\linewidth]{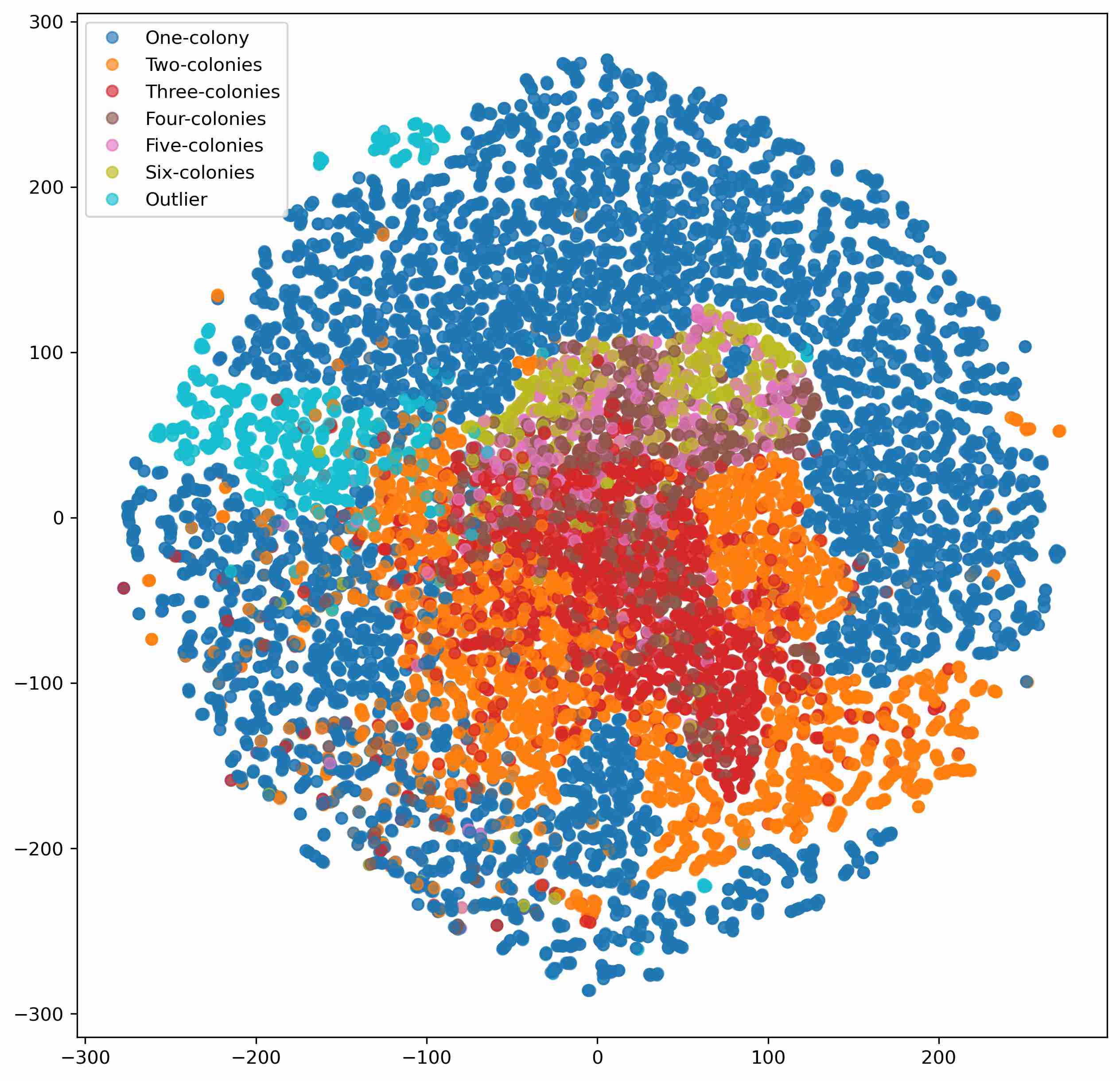}
\includegraphics[width=.45\linewidth]{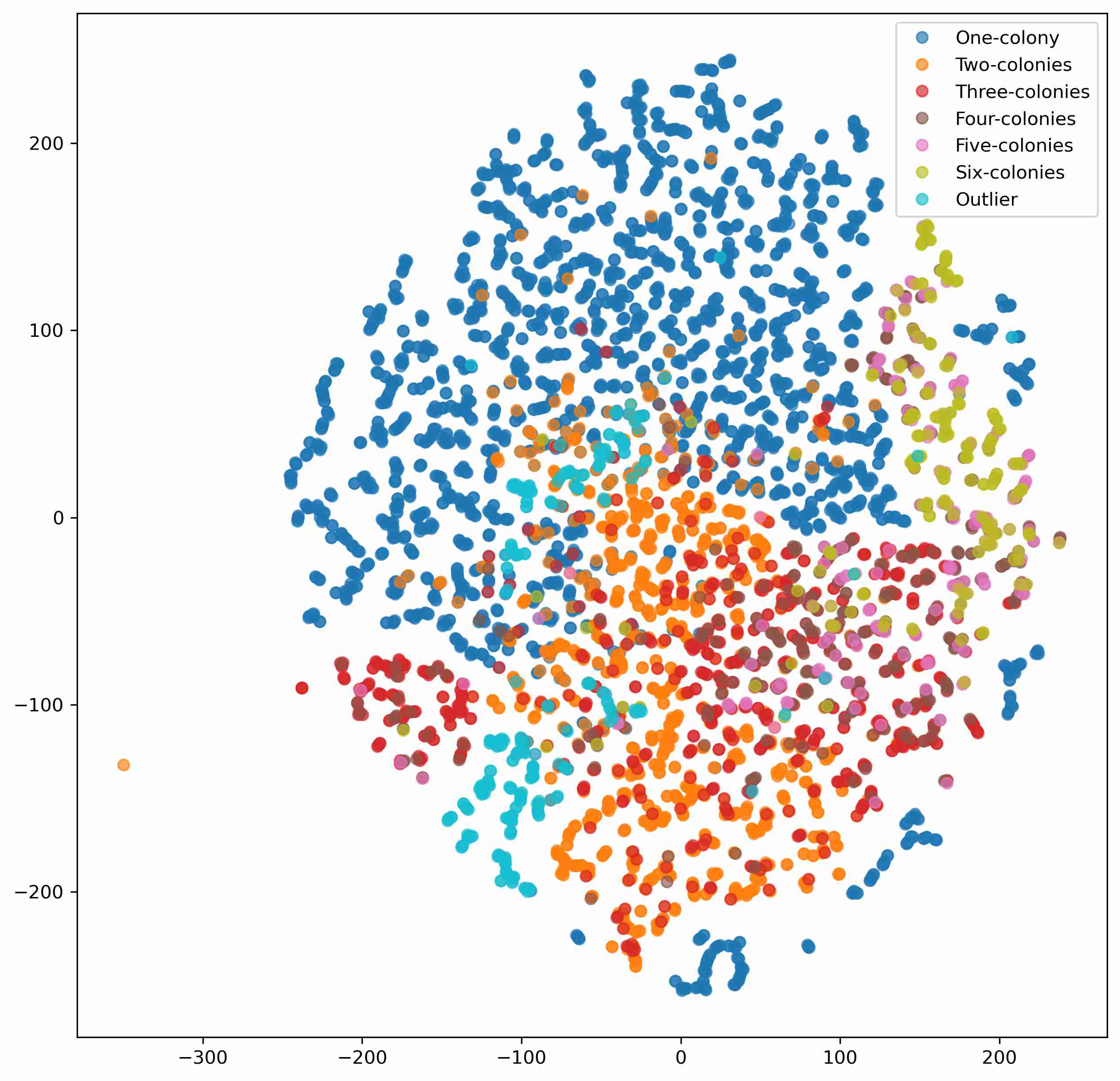}
\caption{The penultimate layer representations from the baseline model on the MicrobiaS1 training set (left) and validation set (right) using t-SNE with a perplexity of 2. Each colour corresponds to one of the seven classes in the dataset. Some classes form distinct clusters (blue, orange, and cyan regions), whereas others (red, brown, pink, and olive regions) are entangled and overlapped extensively.}
\label{fig:visualisation_of_network_layers_with_tsne_2_perplexity_training_val}
\end{figure}

\begin{figure}[h!]
\centering
\includegraphics[width=.45\linewidth]{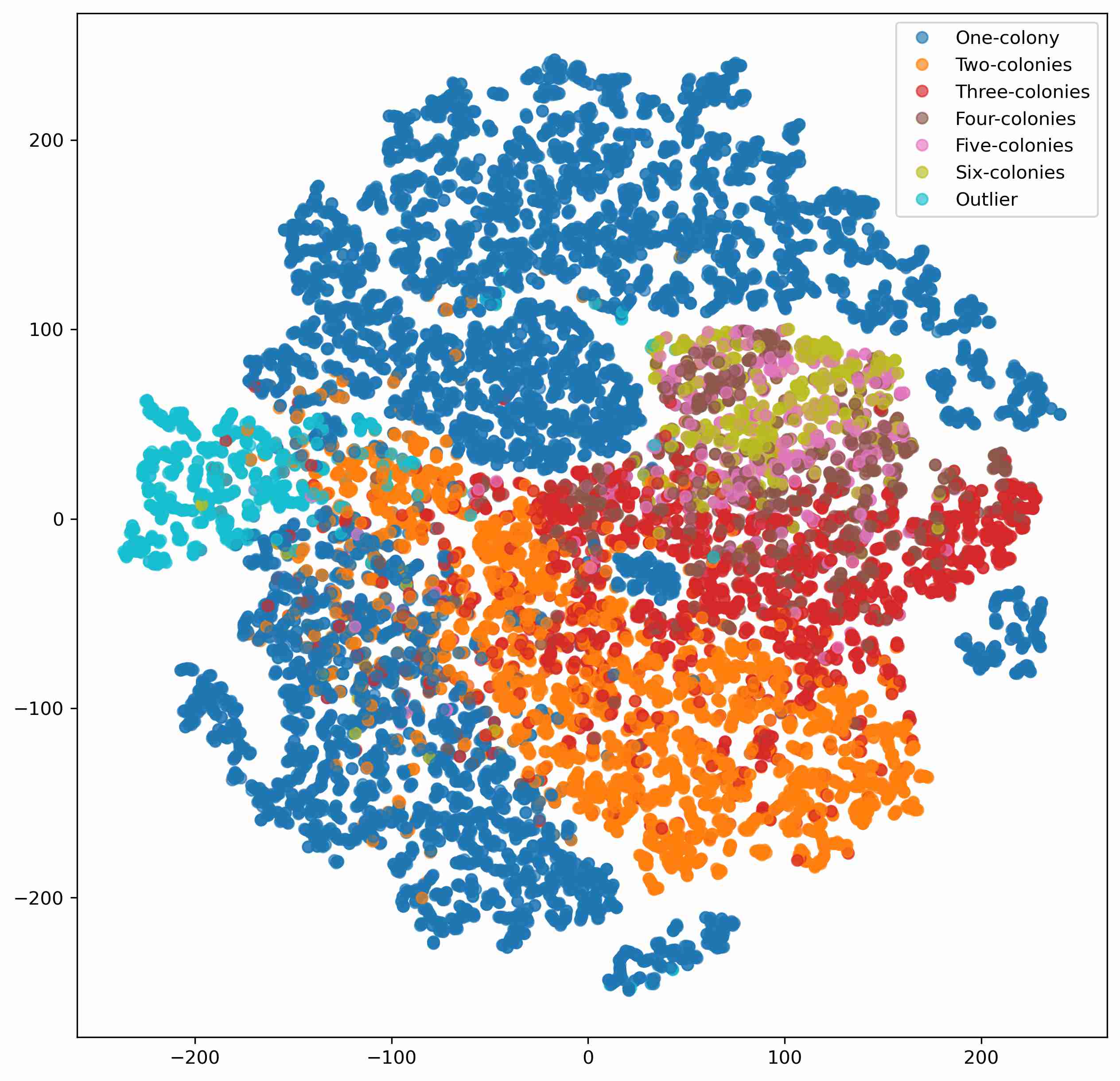}
\includegraphics[width=.45\linewidth]{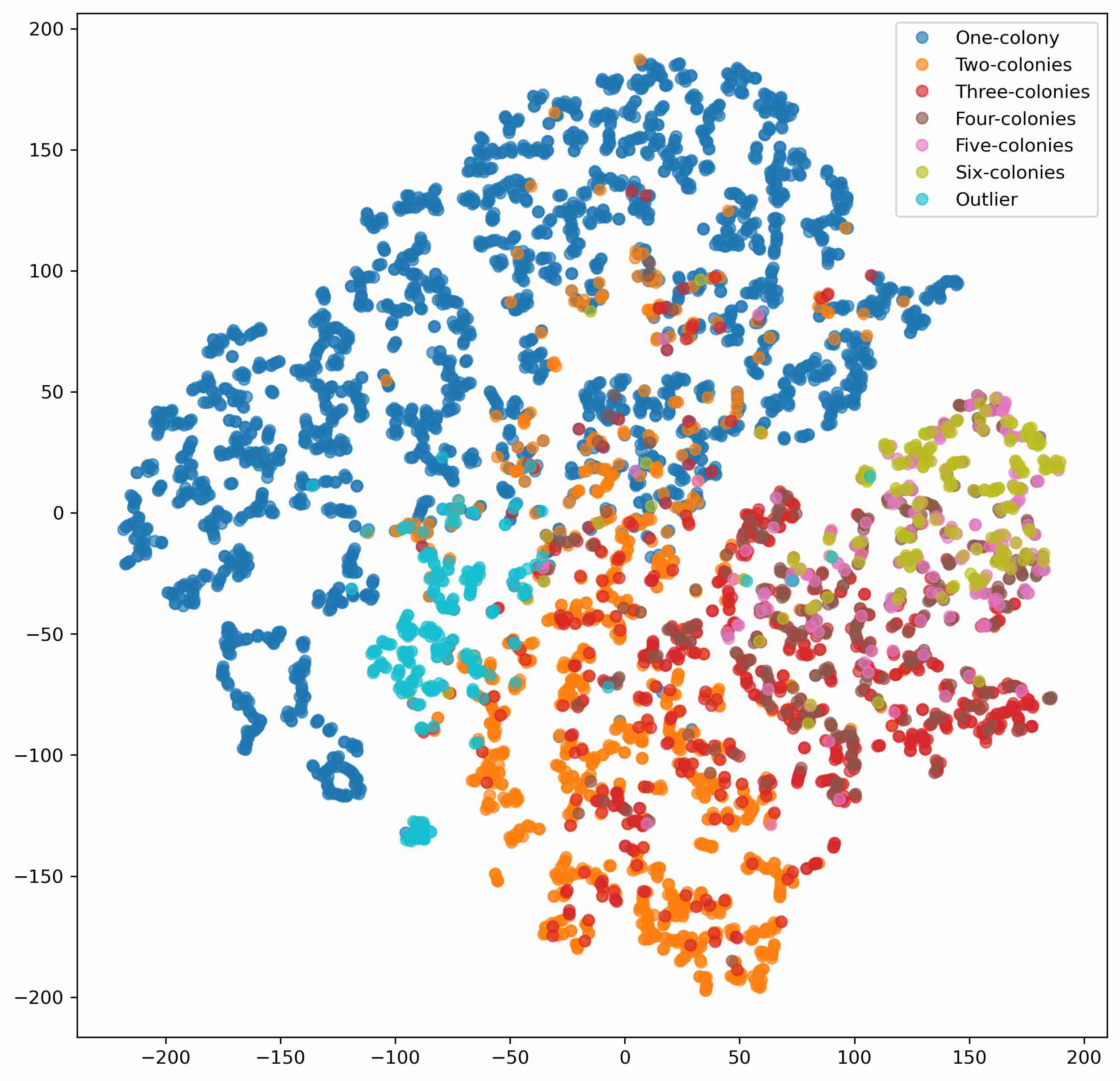}
\caption{Same as Figure \ref{fig:visualisation_of_network_layers_with_tsne_2_perplexity_training_val}, but the perplexity value is 5.}
\label{fig:visualisation_of_network_layers_with_tsne_5_perplexity_training_val}
\end{figure}

\begin{figure}[h!]
\centering
\includegraphics[width=.45\linewidth]{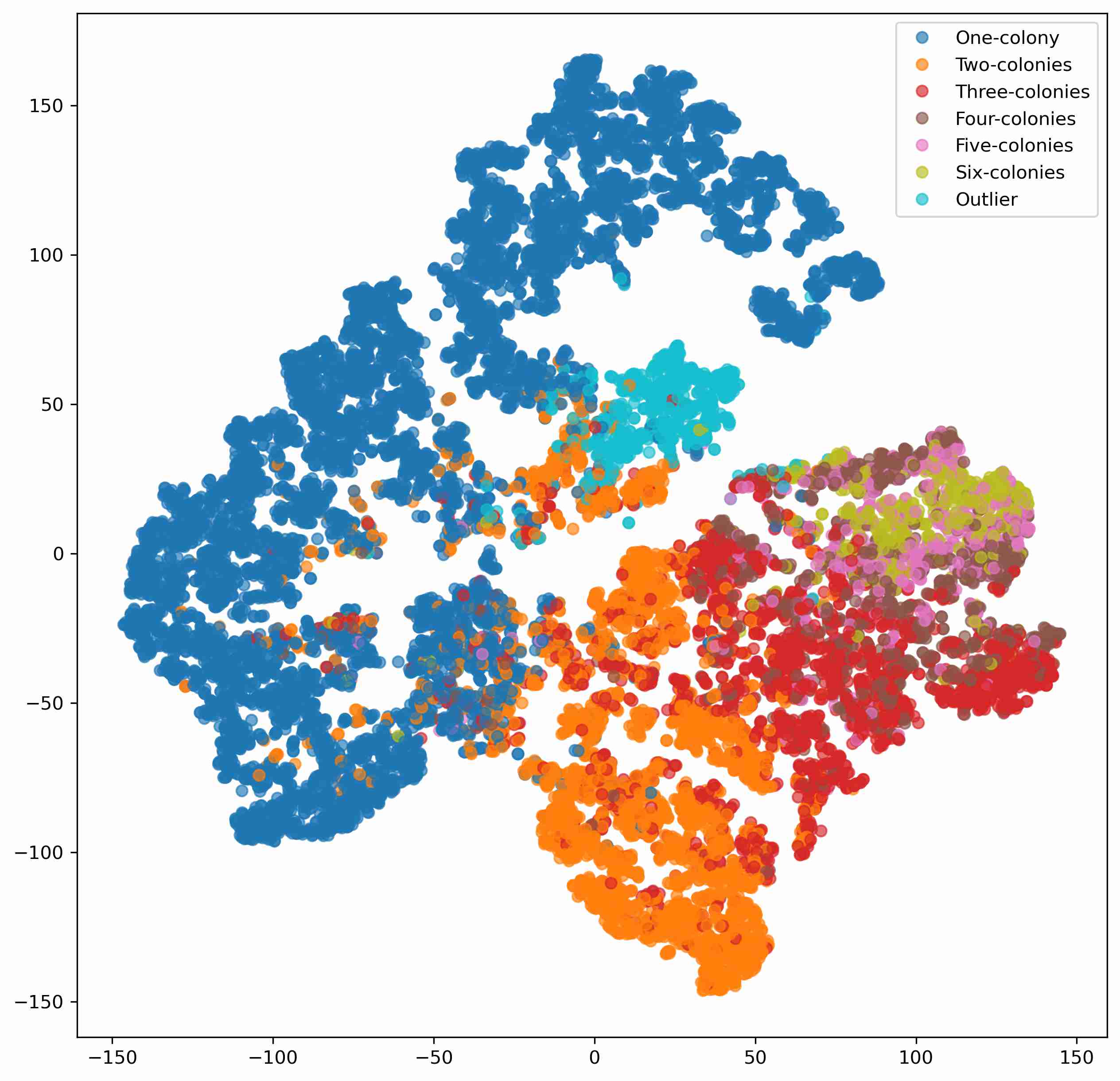}
\includegraphics[width=.45\linewidth]{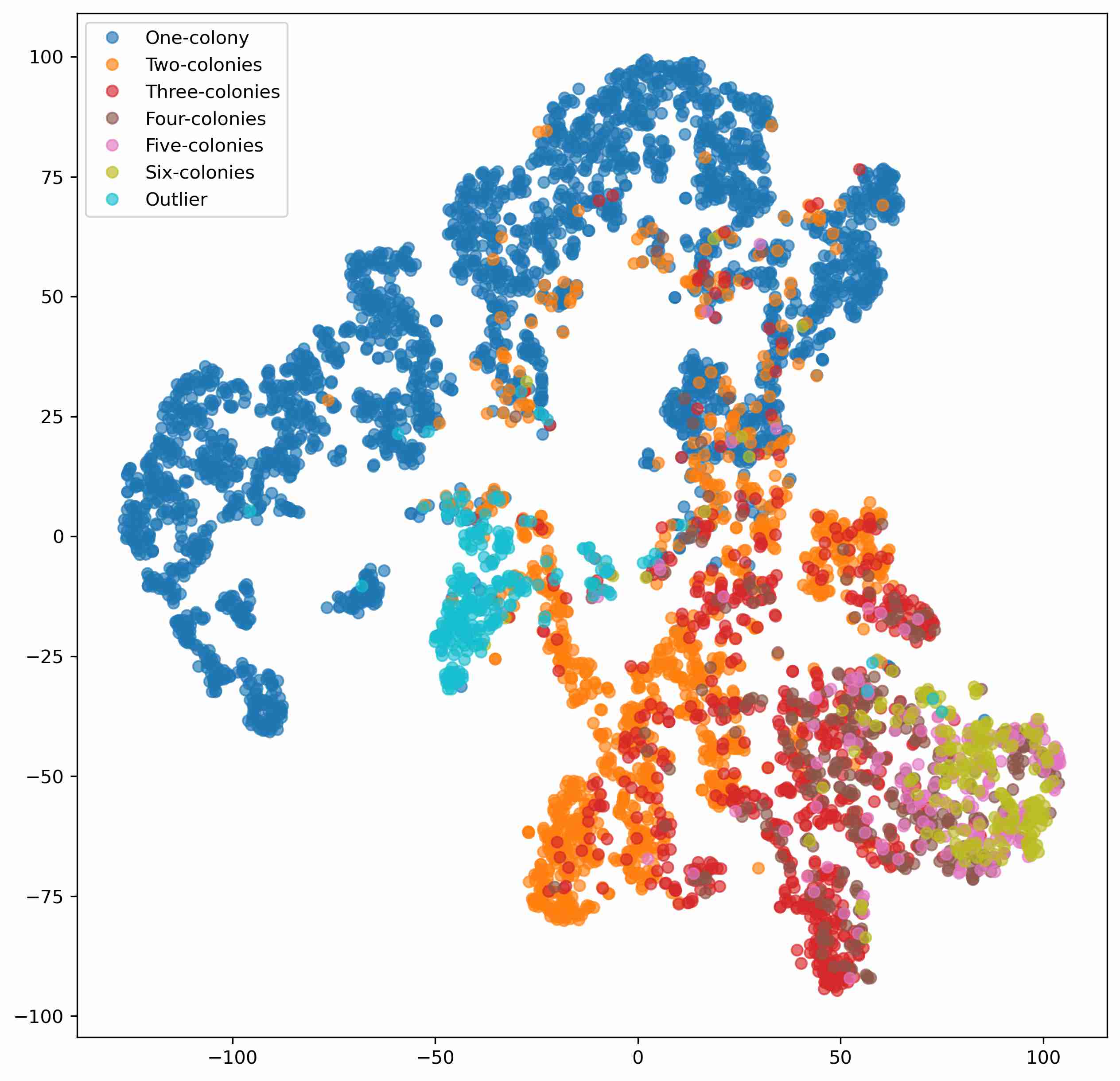}
\caption{Same as Figure \ref{fig:visualisation_of_network_layers_with_tsne_2_perplexity_training_val}, but the perplexity value is 30.}
\label{fig:visualisation_of_network_layers_with_tsne_30_perplexity_training_val}
\end{figure}

\begin{figure}[h!]
\centering
\includegraphics[width=.45\linewidth]{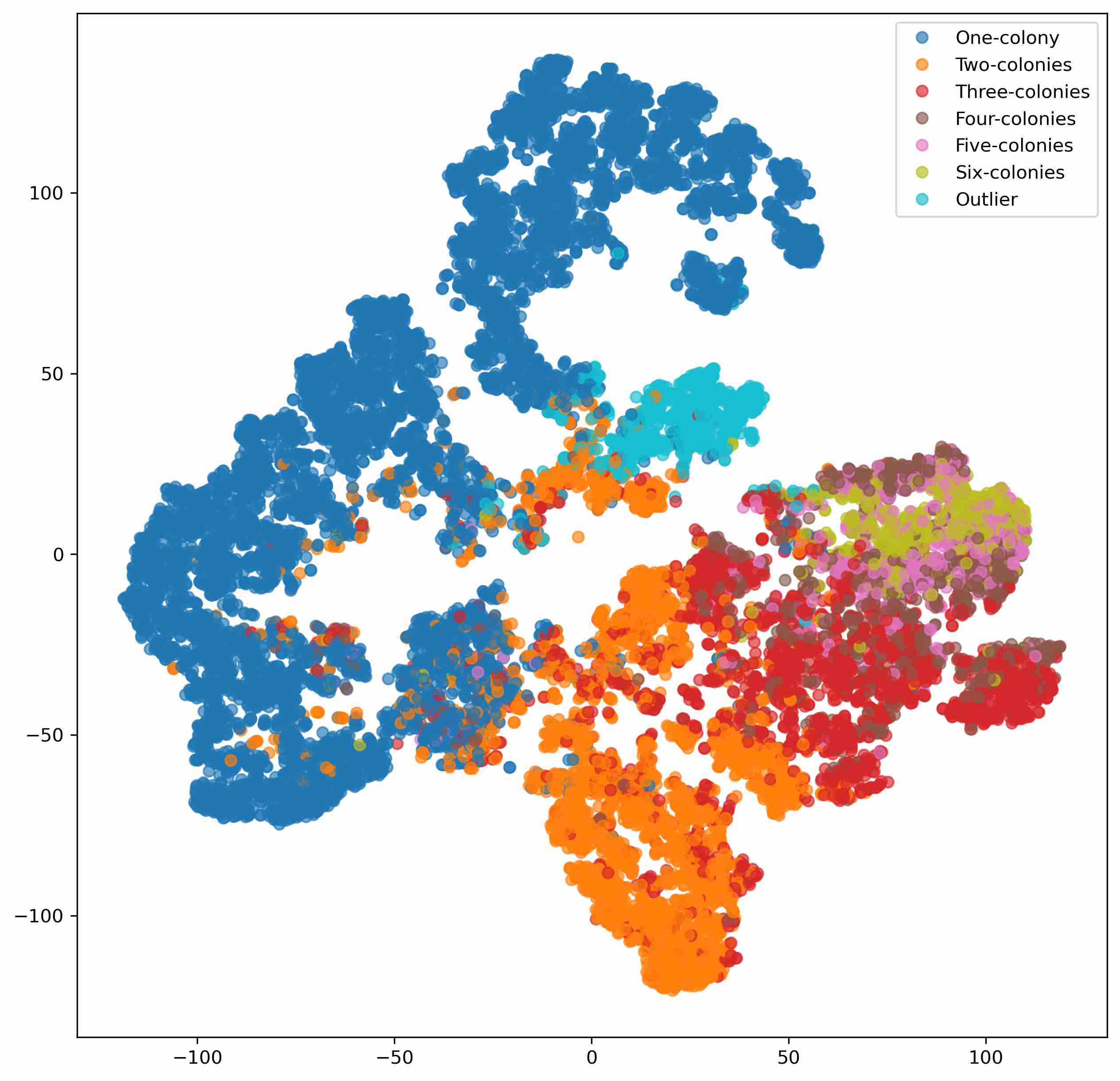}
\includegraphics[width=.45\linewidth]{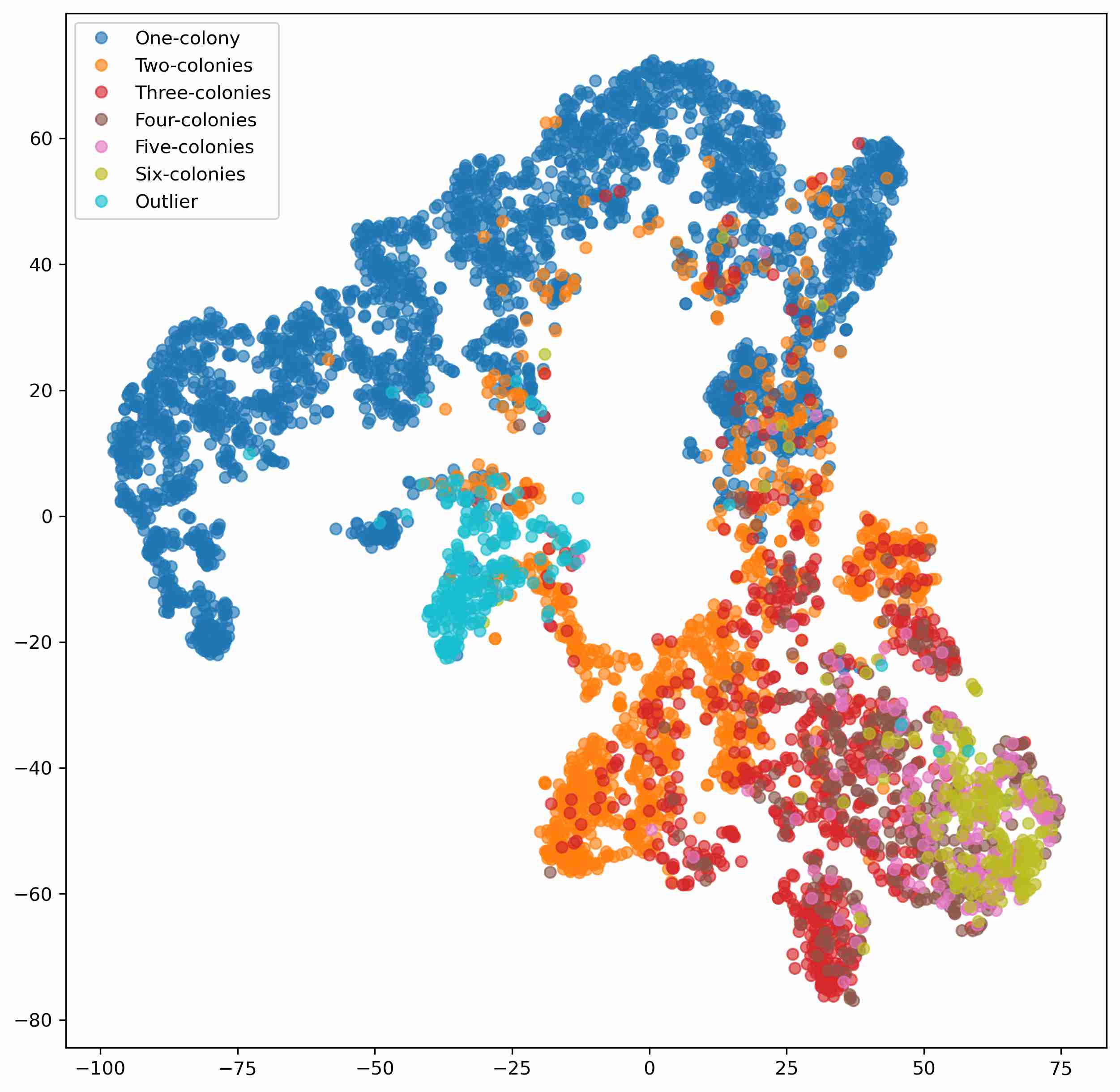}
\caption{Same as Figure \ref{fig:visualisation_of_network_layers_with_tsne_2_perplexity_training_val}, but the perplexity value is 50.}
\label{fig:visualisation_of_network_layers_with_tsne_50_perplexity_training_val}
\end{figure}

\begin{figure}[h!]
\centering
\includegraphics[width=.45\linewidth]{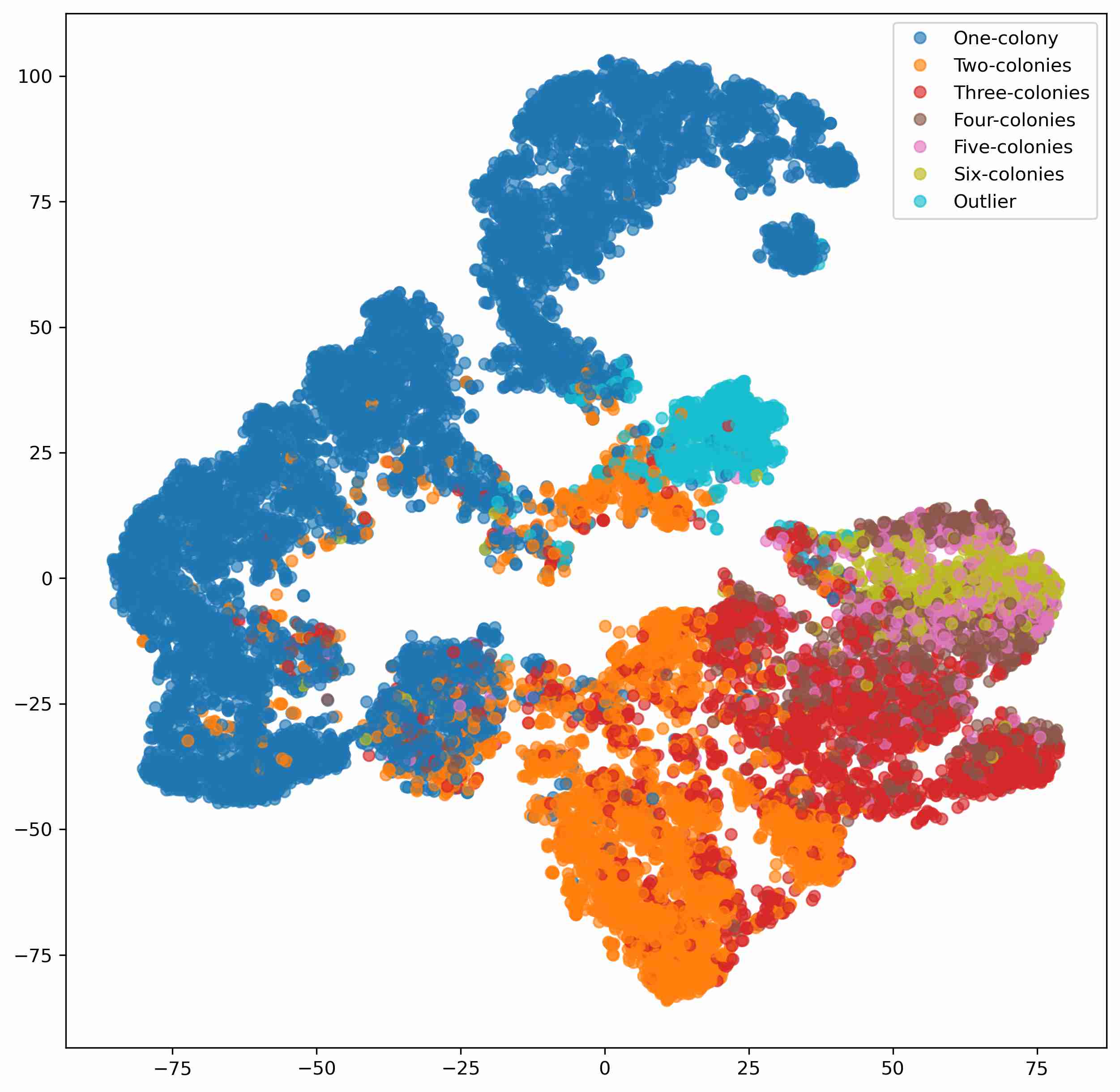}
\includegraphics[width=.45\linewidth]{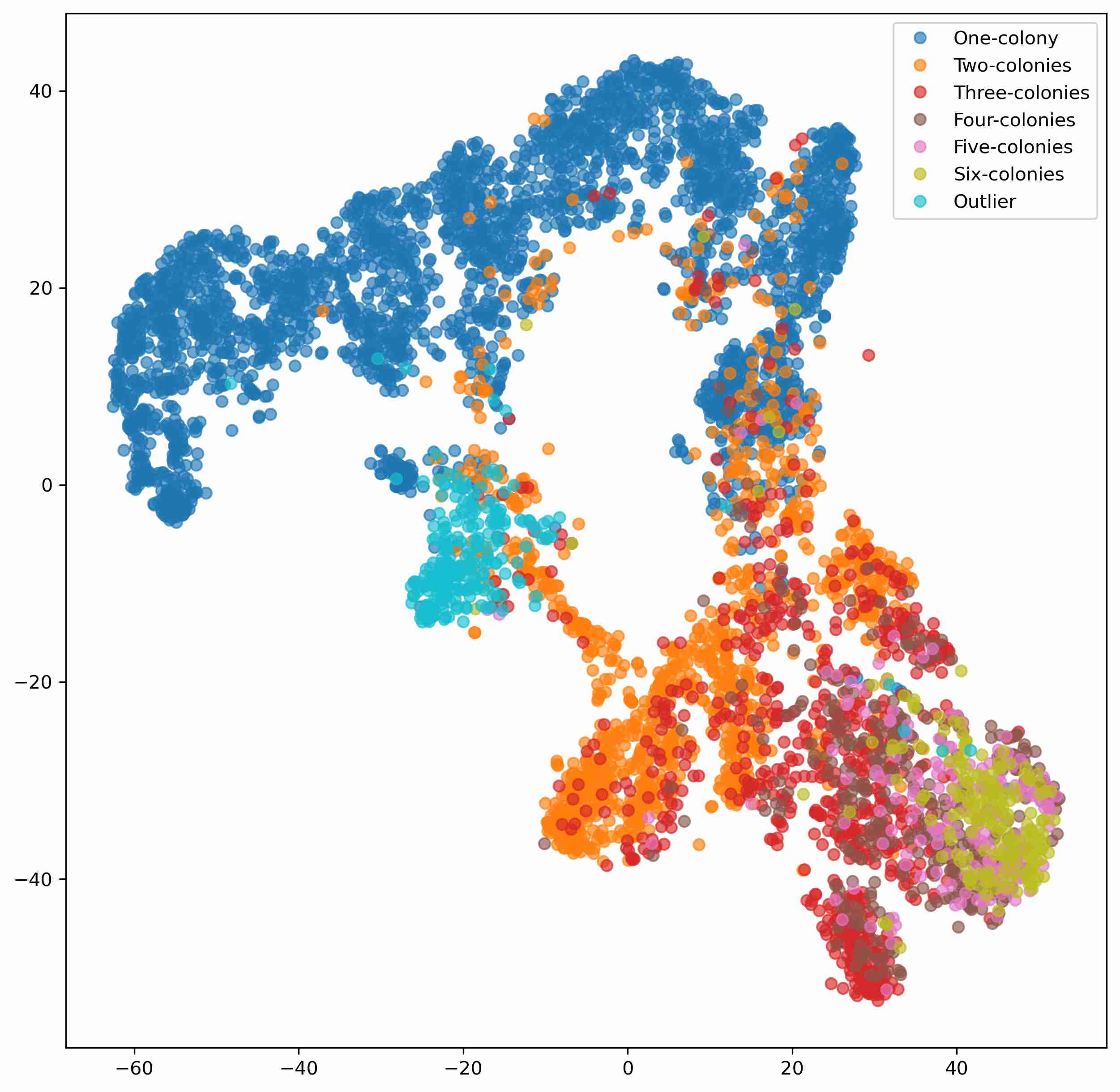}
\caption{Same as Figure \ref{fig:visualisation_of_network_layers_with_tsne_2_perplexity_training_val}, but the perplexity value is 100.}
\label{fig:visualisation_of_network_layers_with_tsne_100_perplexity_training_val}
\end{figure}

\clearpage
\subsection{Visualisation of learned features}
This section includes additional visualisations on the third, fourth, fifth, and sixth kernels in each convolutional layer.

\begin{figure}[h!]
\centering
\includegraphics[width=0.85\linewidth]{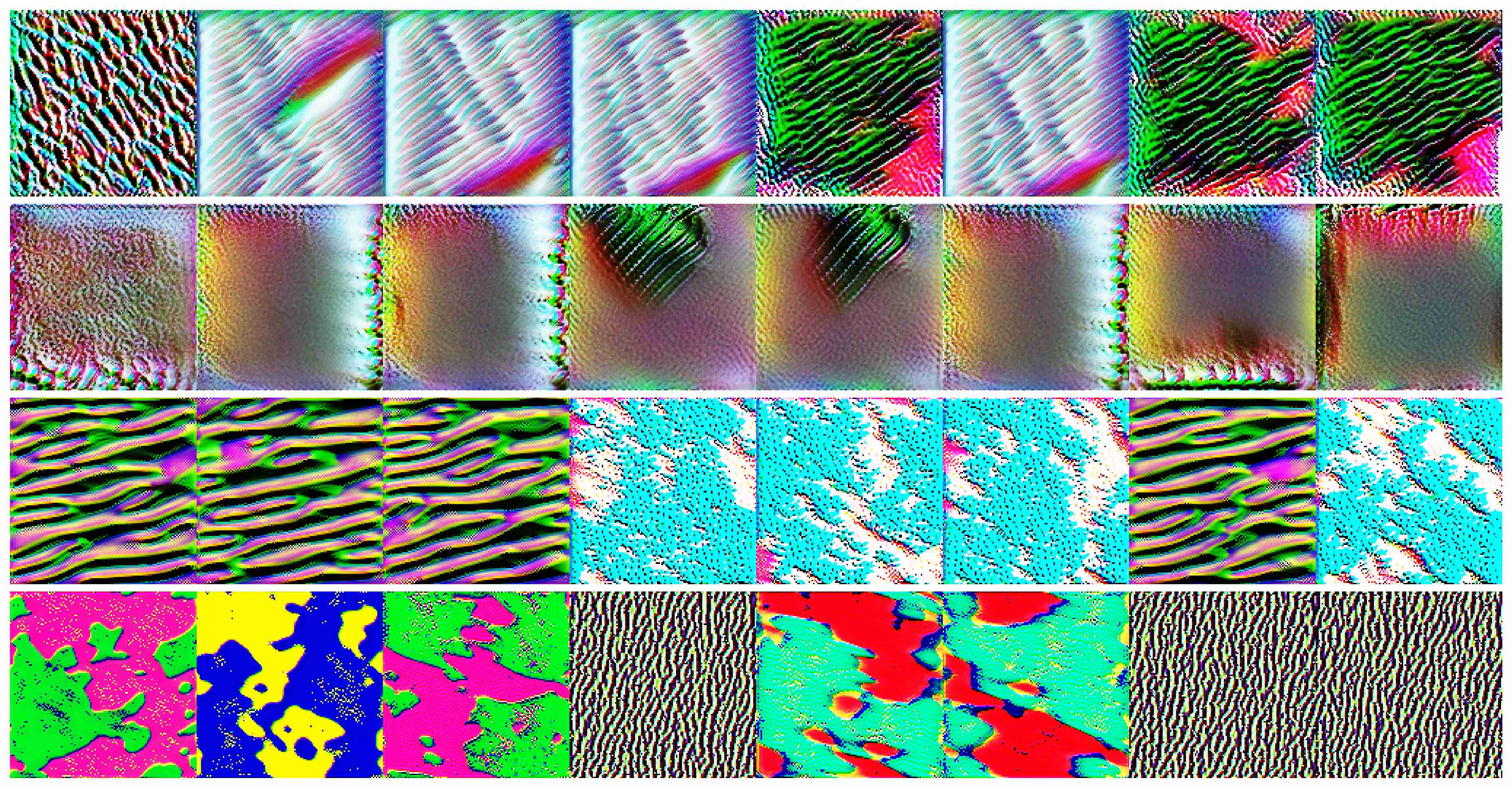}
\caption{Visualisation of feature maps that strongly activate the third convolutional kernel in each of the four convolutional layers. Eight feature maps (shown in the eight columns) are displayed for each of the four convolutional layers (shown in the four rows). The third convolutional kernels in the first, third, and fourth convolutional layers (rows 1, 3, and 4) appear to capture simple texture-based representations. In contrast, the third convolutional kernel in the second convolutional layer (row 2) demonstrates sensitivity to red blob-like shapes that bear a vague resemblance to colonies.}
\label{fig:vis_v4_train_third_conv}
\end{figure}

\begin{figure}[h!]
\centering
\includegraphics[width=0.85\linewidth]{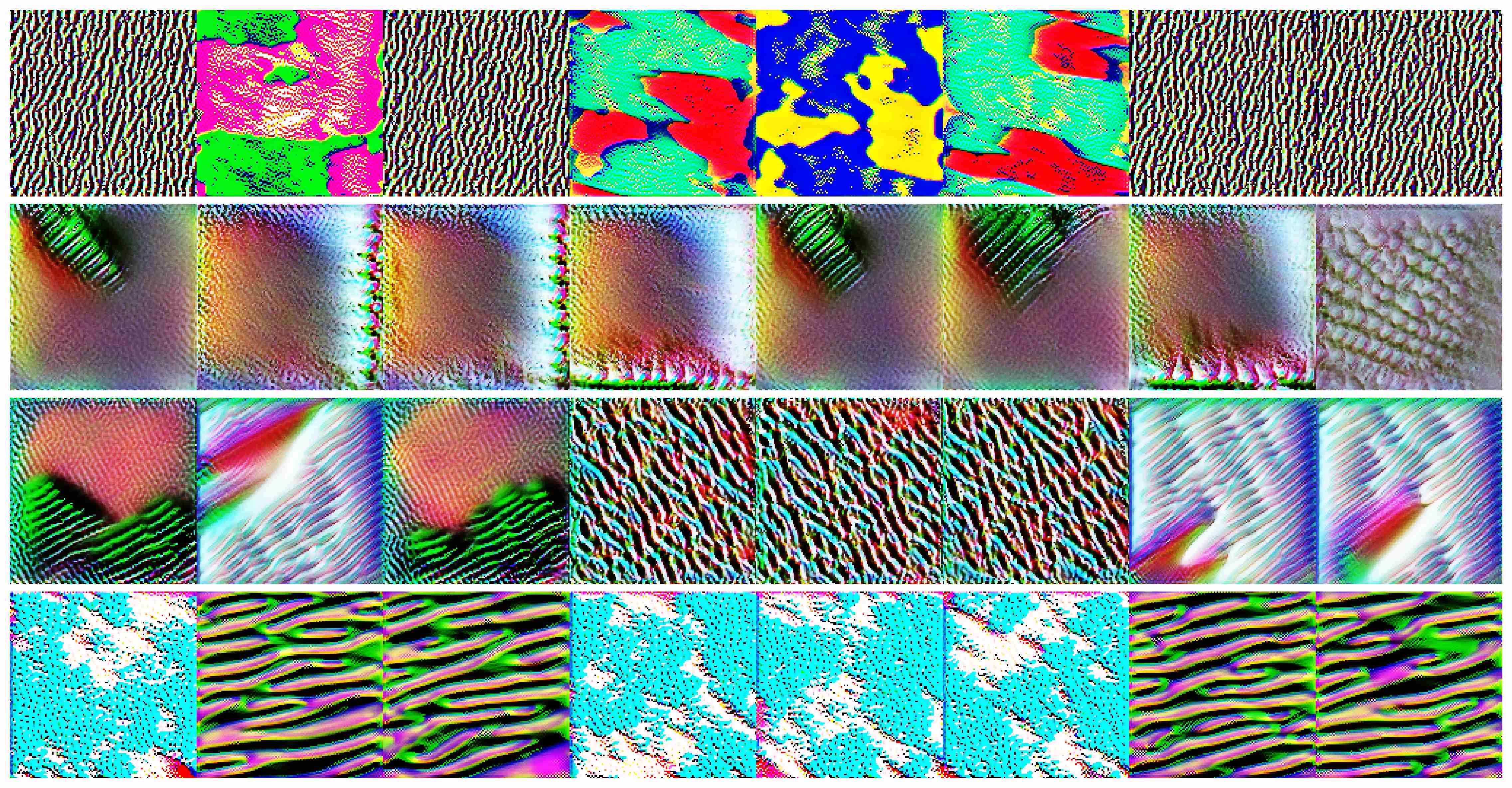}
\caption{Same as Figure \ref{fig:vis_v4_train_third_conv}, but for the fourth convolutional kernel.}
\label{fig:vis_v4_train_fourth_conv}
\end{figure}

\begin{figure}[h!]
\centering
\includegraphics[width=0.85\linewidth]{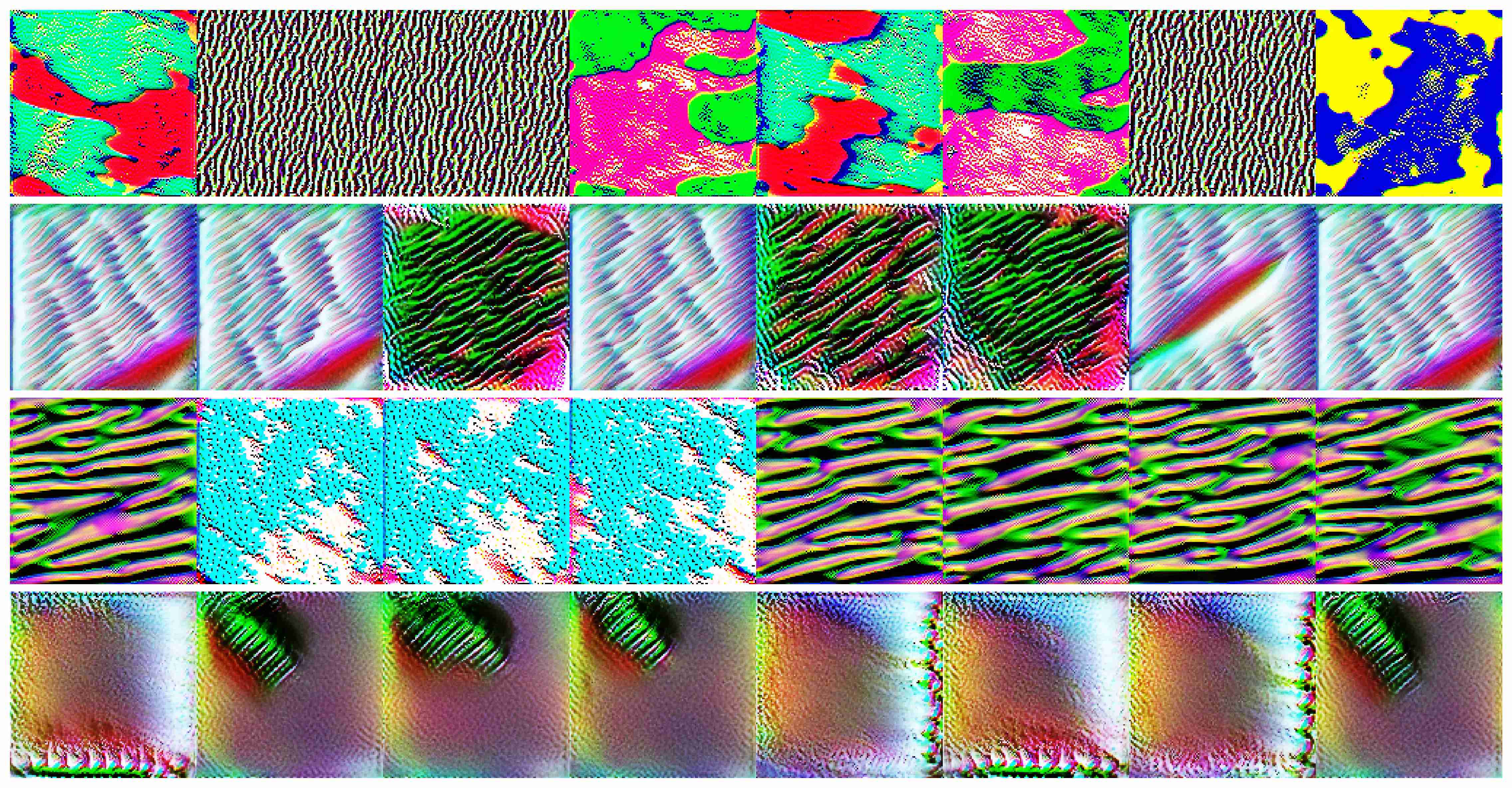}
\caption{Visualisation of feature maps that strongly activate the fifth convolutional kernel in each of the four convolutional layers. Eight feature maps (shown in the eight columns) are displayed for each of the four convolutional layers (shown in the four rows). The fifth convolutional kernels in the first three convolutional layers (rows 1, 2, and 3) appear to capture simple texture-based representations. In contrast, the fifth convolutional kernel in the fourth convolutional layer (row 4) demonstrates sensitivity to red blob-like shapes that bear a vague resemblance to colonies.}
\label{fig:vis_v4_train_fifth_conv}
\end{figure}

\begin{figure}[h!]
\centering
\includegraphics[width=0.85\linewidth]{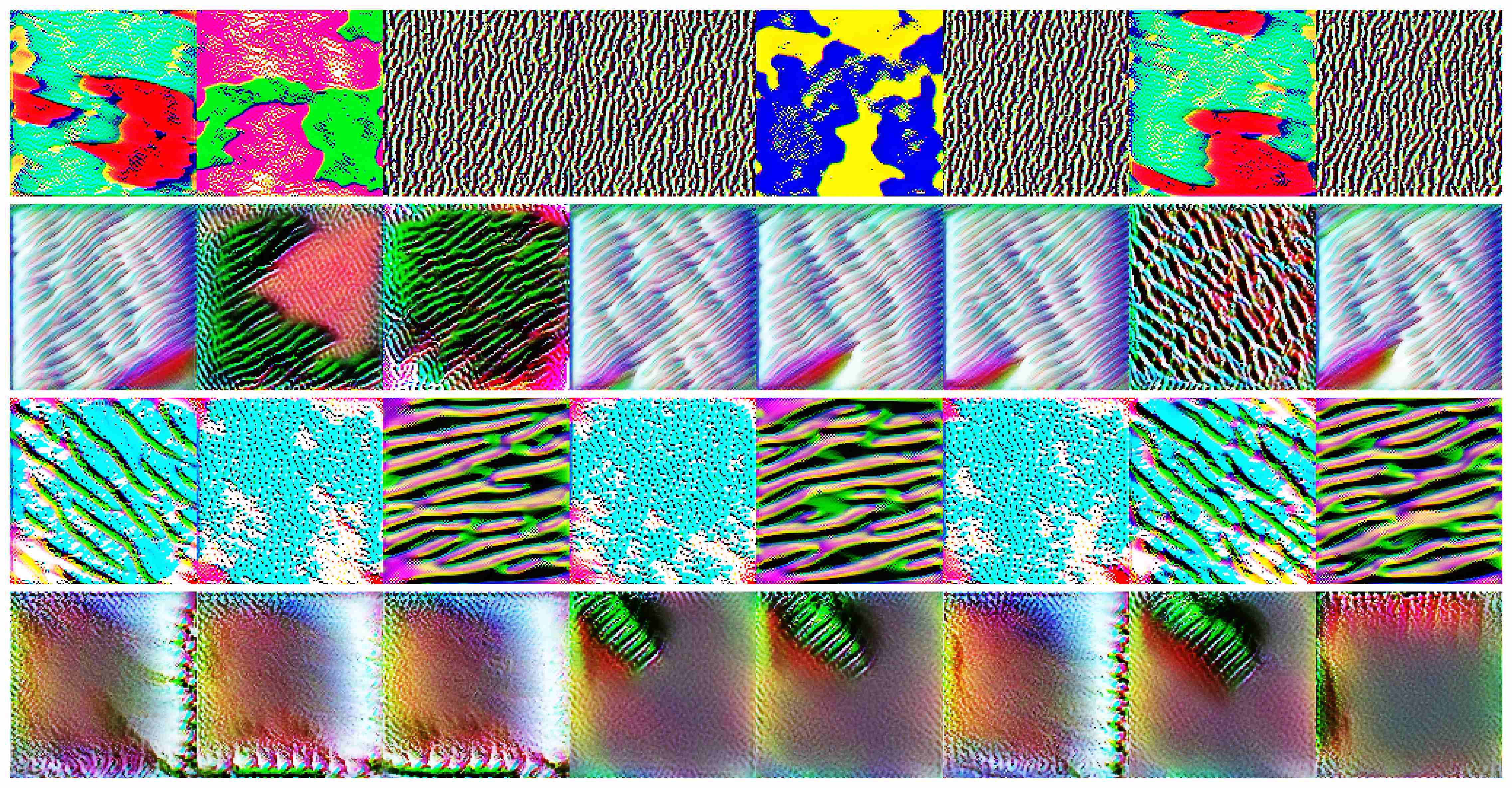}
\caption{Same as Figure \ref{fig:vis_v4_train_fifth_conv}, but for the sixth convolutional kernel.}
\label{fig:vis_v4_train_sixth_conv}
\end{figure}

\clearpage
\subsection{Visualisation of class activation maps}
This section includes additional visualisations on activation maps for all classes except the One-colony class.

\begin{figure}[h!]
\centering
\includegraphics[width=0.85\linewidth]{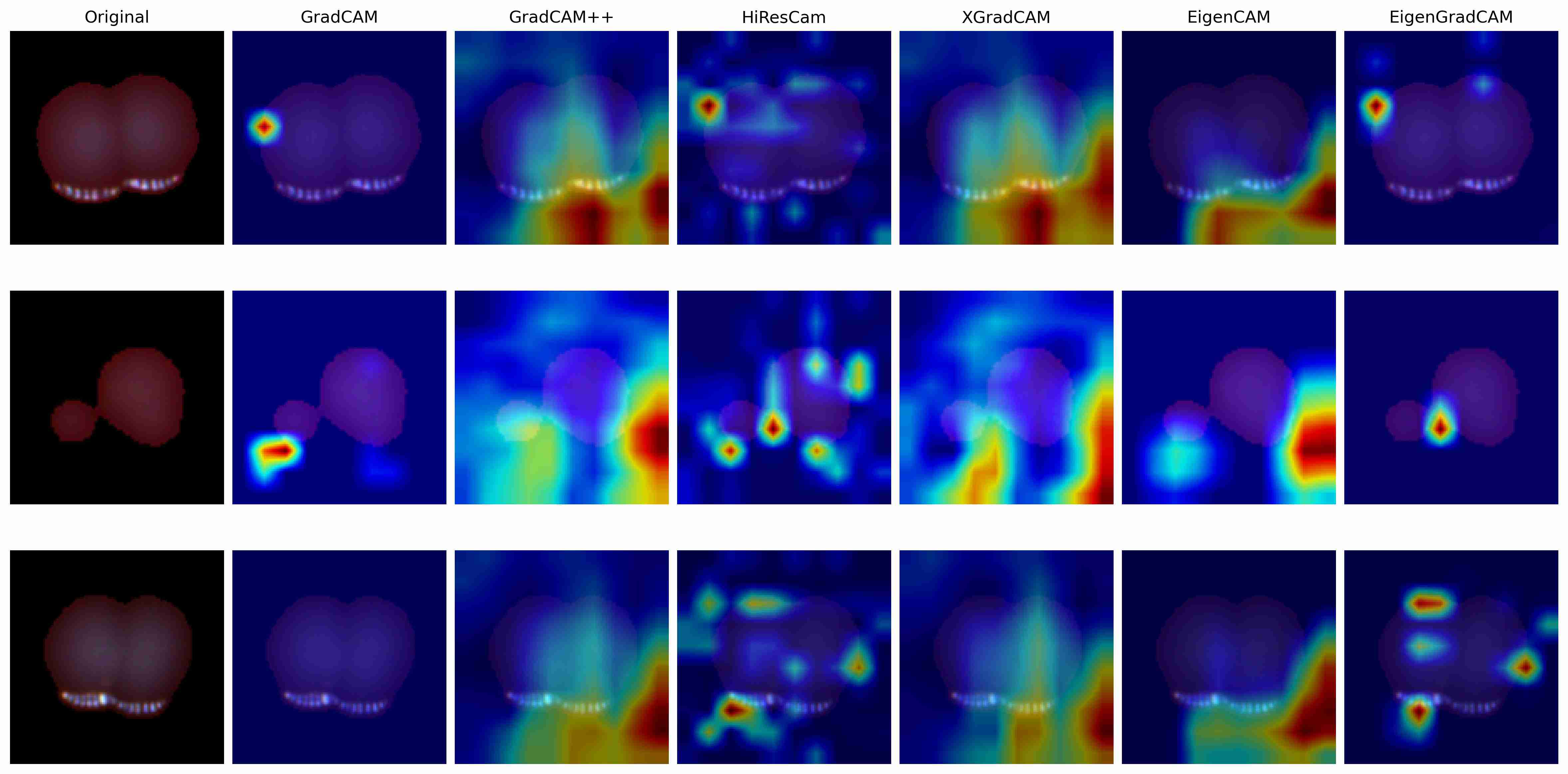}
\caption{Class activation map visualisation for Two-colonies images. Highlighted regions indicate the areas that contribute most strongly to the model’s predictions, with colour intensity representing the degree of contribution.}
\label{fig:cam_vis_v4_two_colonies_conv}
\end{figure}

\begin{figure}[h!]
\centering
\includegraphics[width=0.85\linewidth]{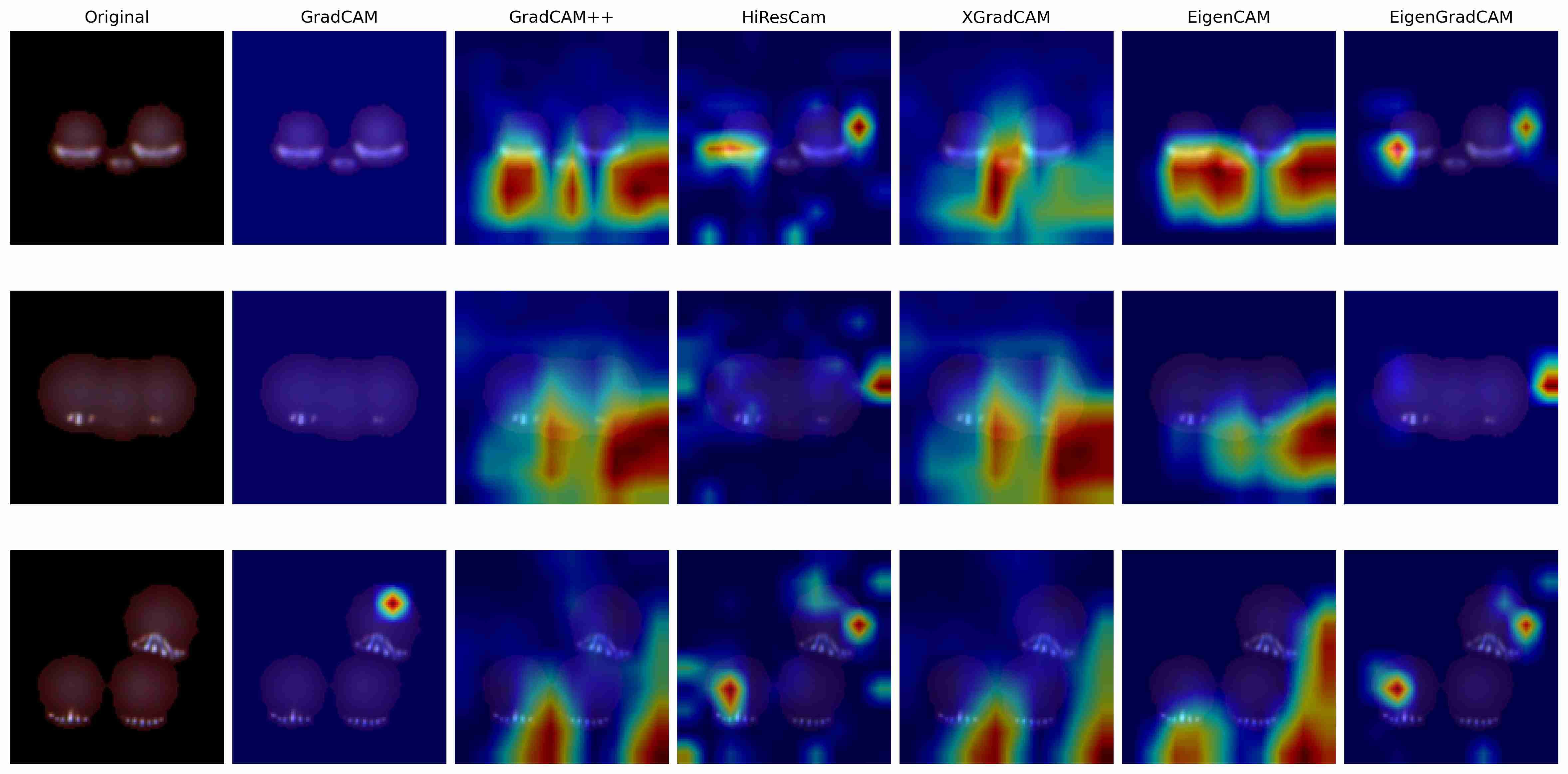}
\caption{Same as Figure \ref{fig:cam_vis_v4_two_colonies_conv}, but for the Three-colonies images.}
\label{fig:cam_vis_v4_three_colonies_conv}
\end{figure}

\begin{figure}[h!]
\centering
\includegraphics[width=0.85\linewidth]{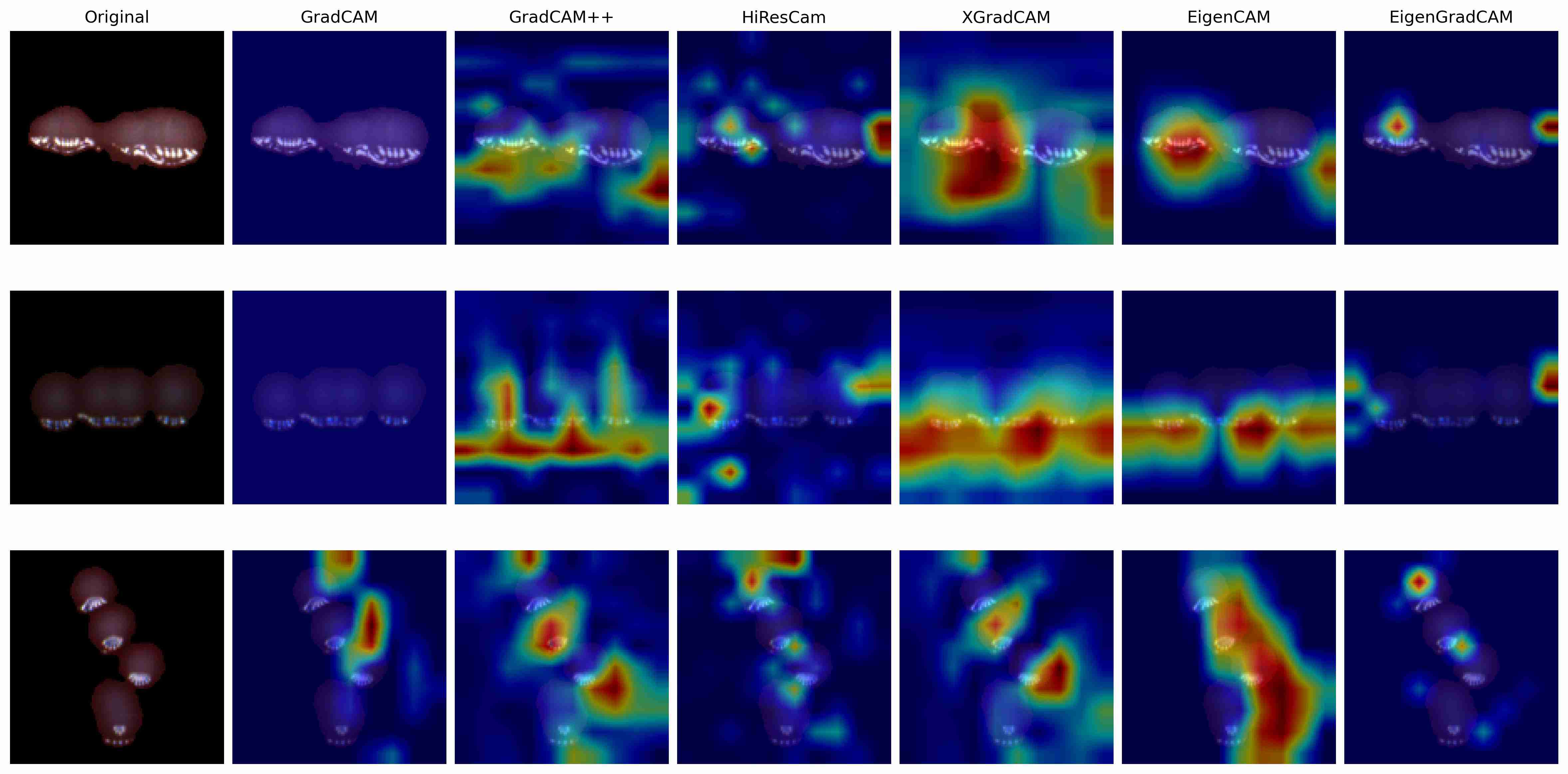}
\caption{Same as Figure \ref{fig:cam_vis_v4_two_colonies_conv}, but for the Four-colonies images.}
\label{fig:cam_vis_v4_four_colonies_conv}
\end{figure}

\begin{figure}[h!]
\centering
\includegraphics[width=0.85\linewidth]{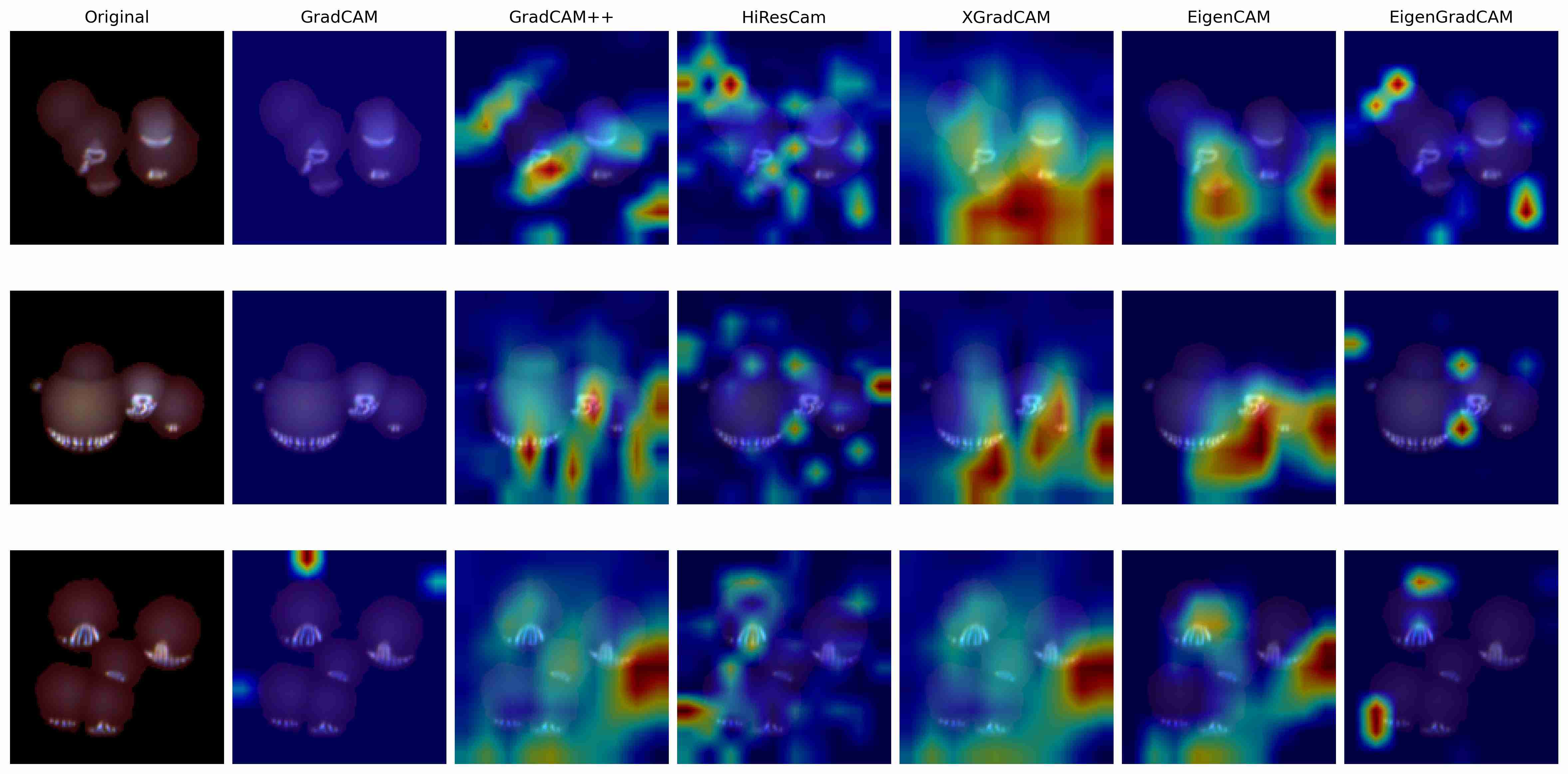}
\caption{Same as Figure \ref{fig:cam_vis_v4_two_colonies_conv}, but for the Five-colonies images.}
\label{fig:cam_vis_v4_five_colonies_conv}
\end{figure}

\begin{figure}[h!]
\centering
\includegraphics[width=0.85\linewidth]{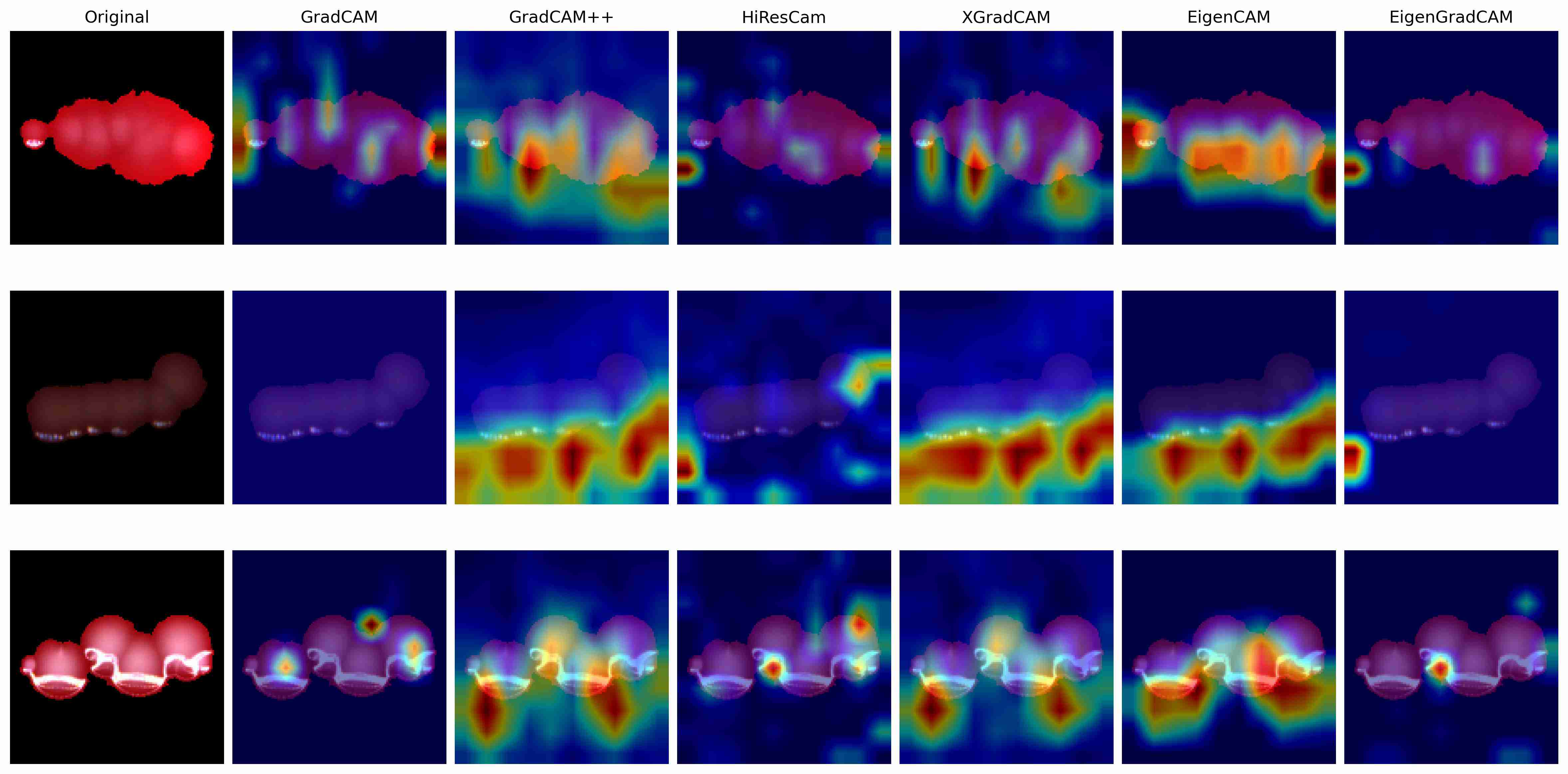}
\caption{Same as Figure \ref{fig:cam_vis_v4_two_colonies_conv}, but for the Six-colonies images.}
\label{fig:cam_vis_v4_six_colonies_conv}
\end{figure}

\begin{figure}[h!]
\centering
\includegraphics[width=0.85\linewidth]{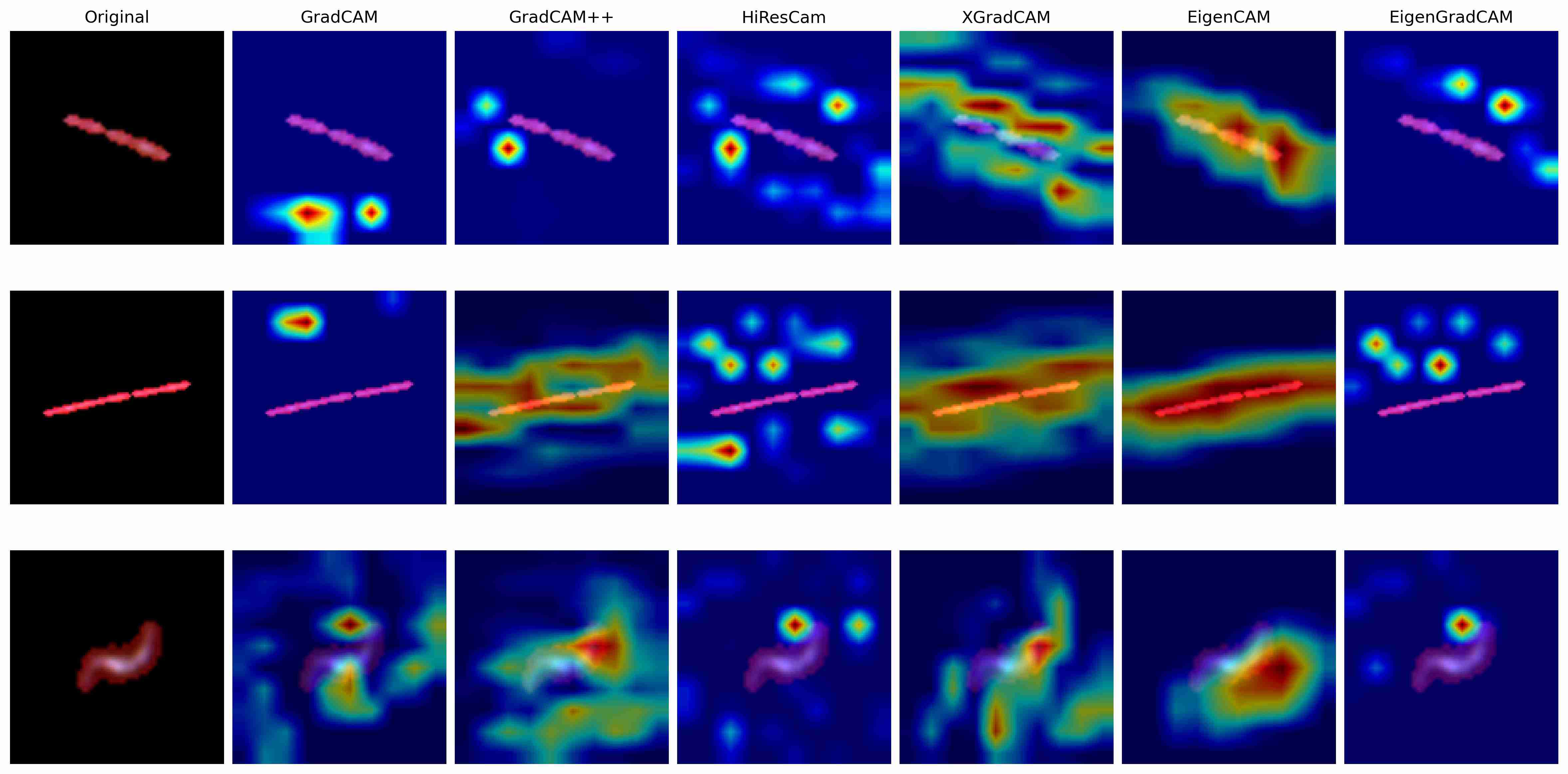}
\caption{Same as Figure \ref{fig:cam_vis_v4_two_colonies_conv}, but for the Outlier images.}
\label{fig:cam_vis_v4_outlier_conv}
\end{figure}

\clearpage
\section{Analysis of the impact of class imbalance on MicrobiaNet}
\label{appendix-impact-of-class-imbalance}
This section includes additional experimental results to visualise the penultimate network layer outputs from the balanced MicrobiaNet model on the MicrobiaS1B1 training and validation sets.

\begin{figure}[h!]
\centering
\includegraphics[width=.45\linewidth]{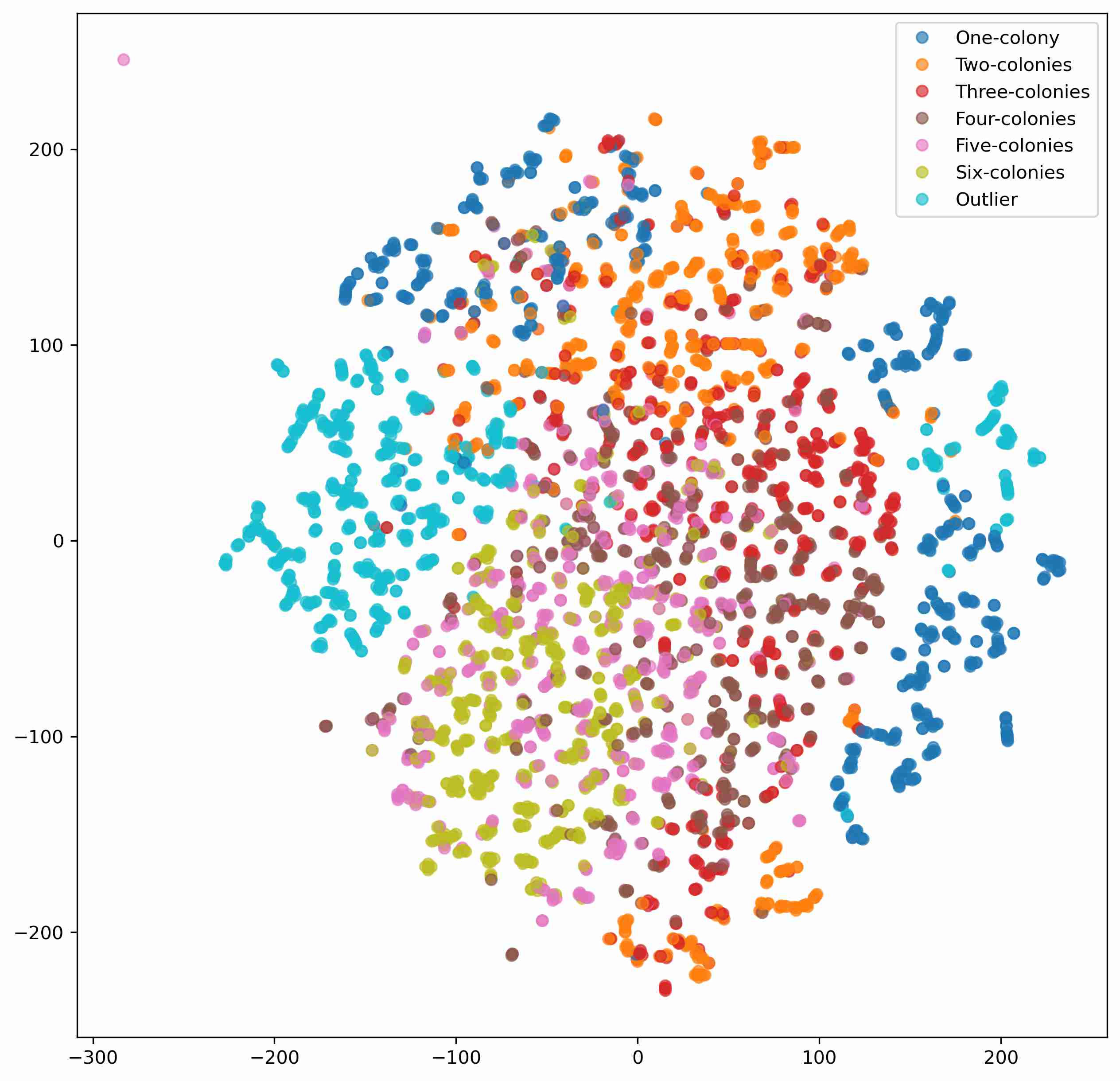}
\includegraphics[width=.45\linewidth]{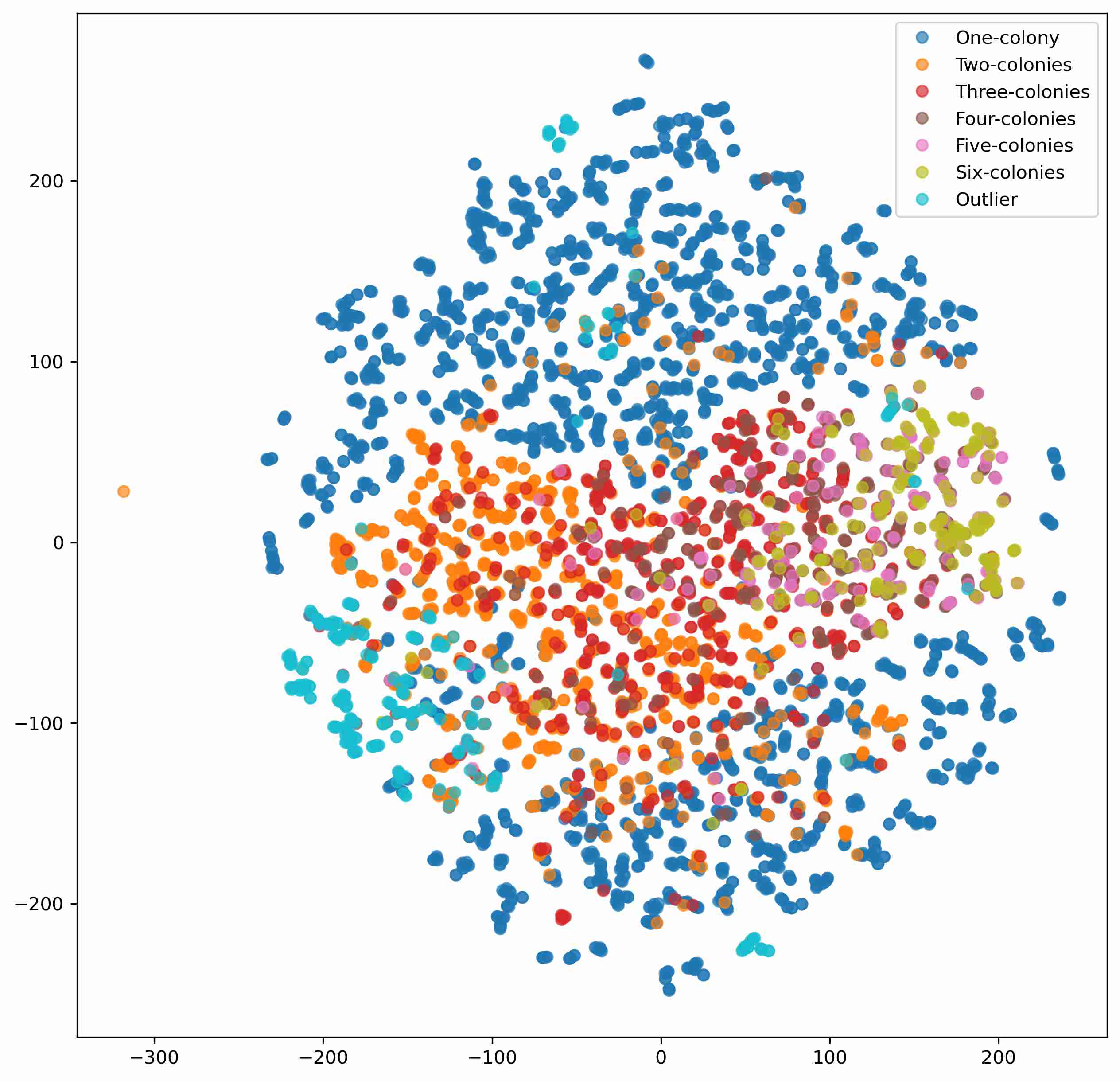}
\caption{The penultimate layer representations from the balanced MicrobiaNet model on the MicrobiaS1B1 training set (left) and validation set (right) using t-SNE with a perplexity of 2. Each colour corresponds to one of the seven classes in the dataset. Some classes form distinct clusters (blue, orange, and cyan regions), whereas others (red, brown, pink, and olive regions) are entangled and overlapped extensively.}
\label{fig:visualisation_of_network_layers_with_tsne_2_perplexity_training_val_v440}
\end{figure}

\begin{figure}[h!]
\centering
\includegraphics[width=.45\linewidth]{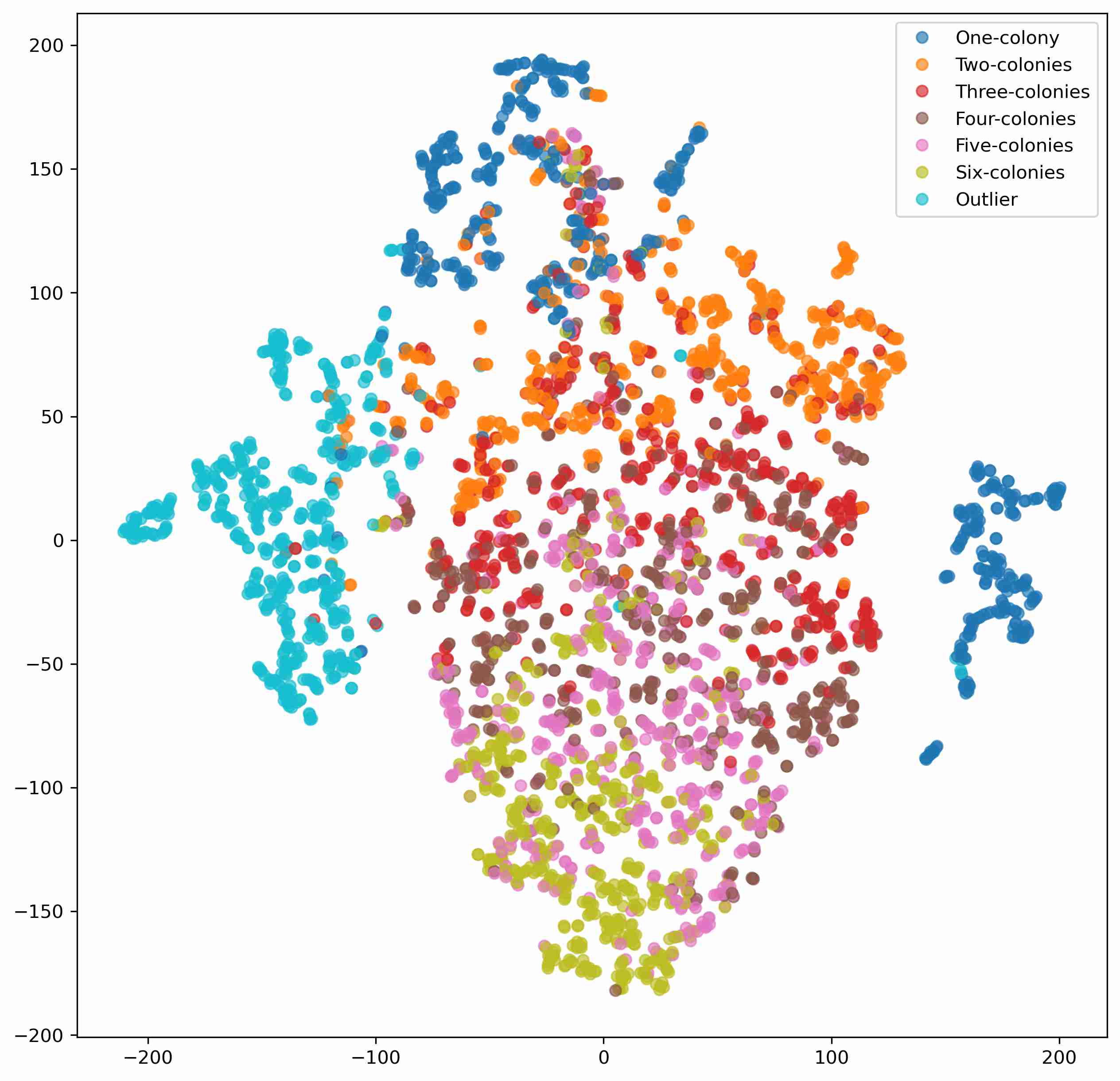}
\includegraphics[width=.45\linewidth]{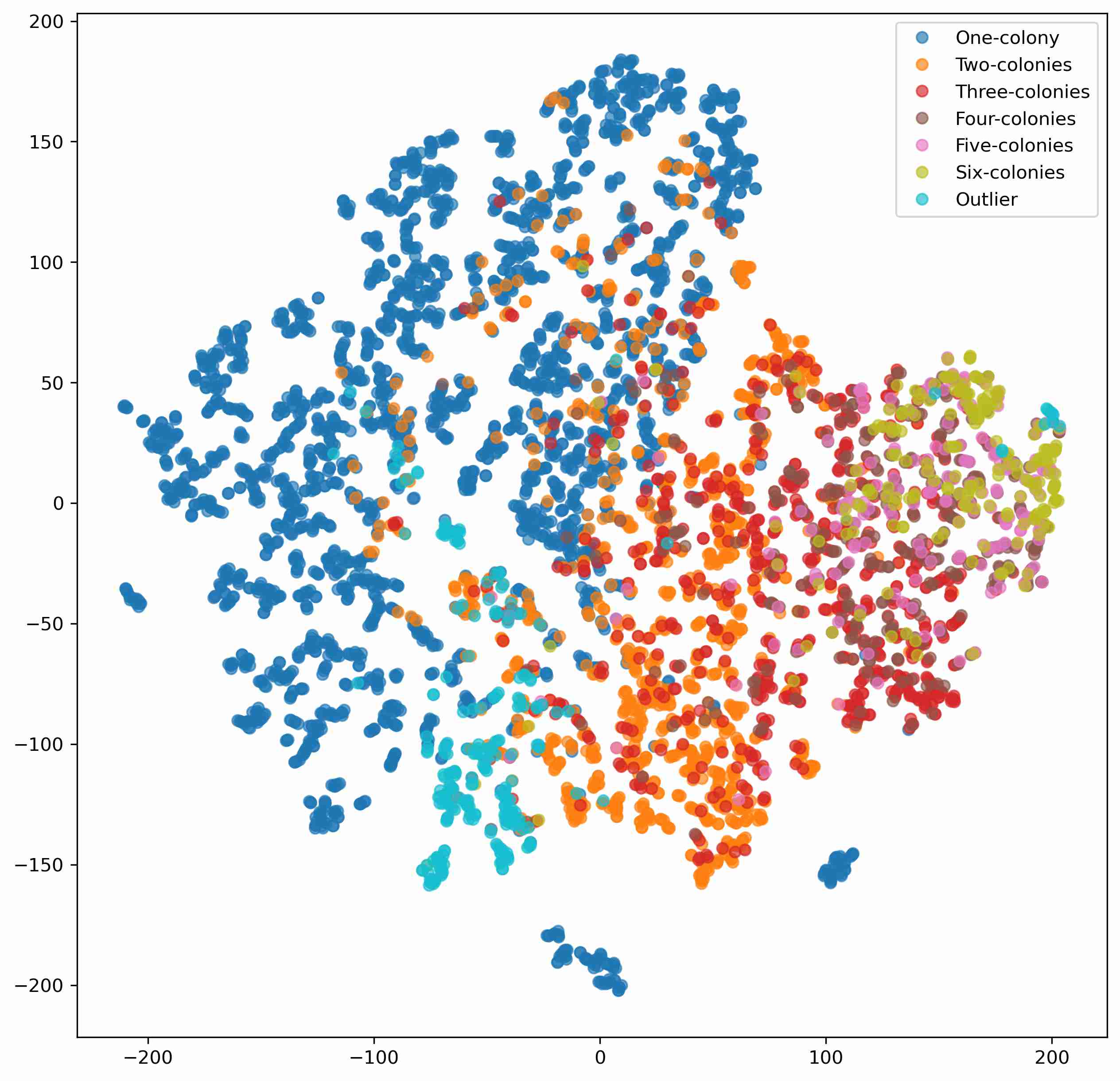}
\caption{Same as Figure \ref{fig:visualisation_of_network_layers_with_tsne_2_perplexity_training_val_v440}, but the perplexity value is 5.}
\label{fig:visualisation_of_network_layers_with_tsne_5_perplexity_training_val_v440}
\end{figure}

\begin{figure}[h!]
\centering
\includegraphics[width=.45\linewidth]{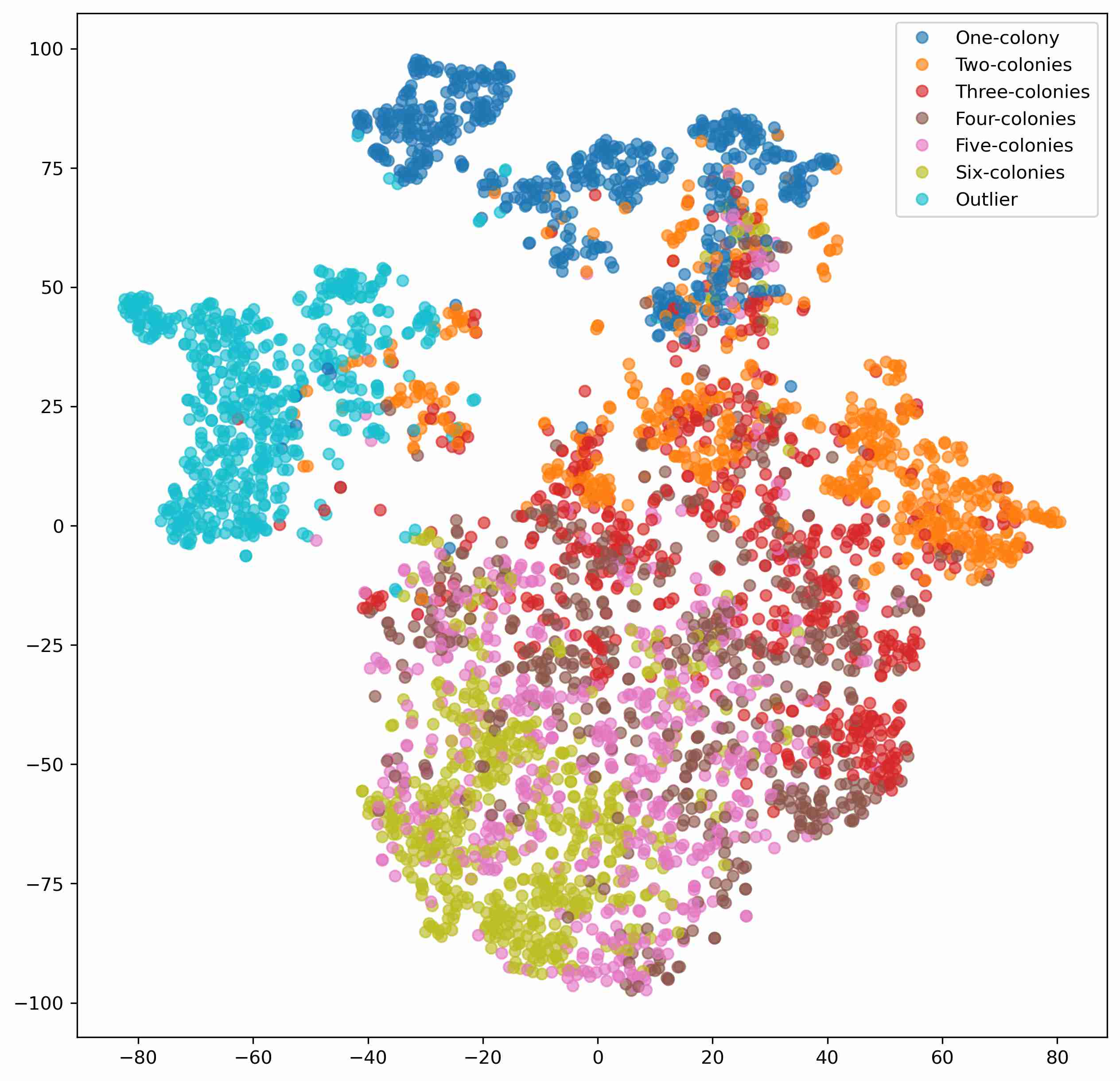}
\includegraphics[width=.45\linewidth]{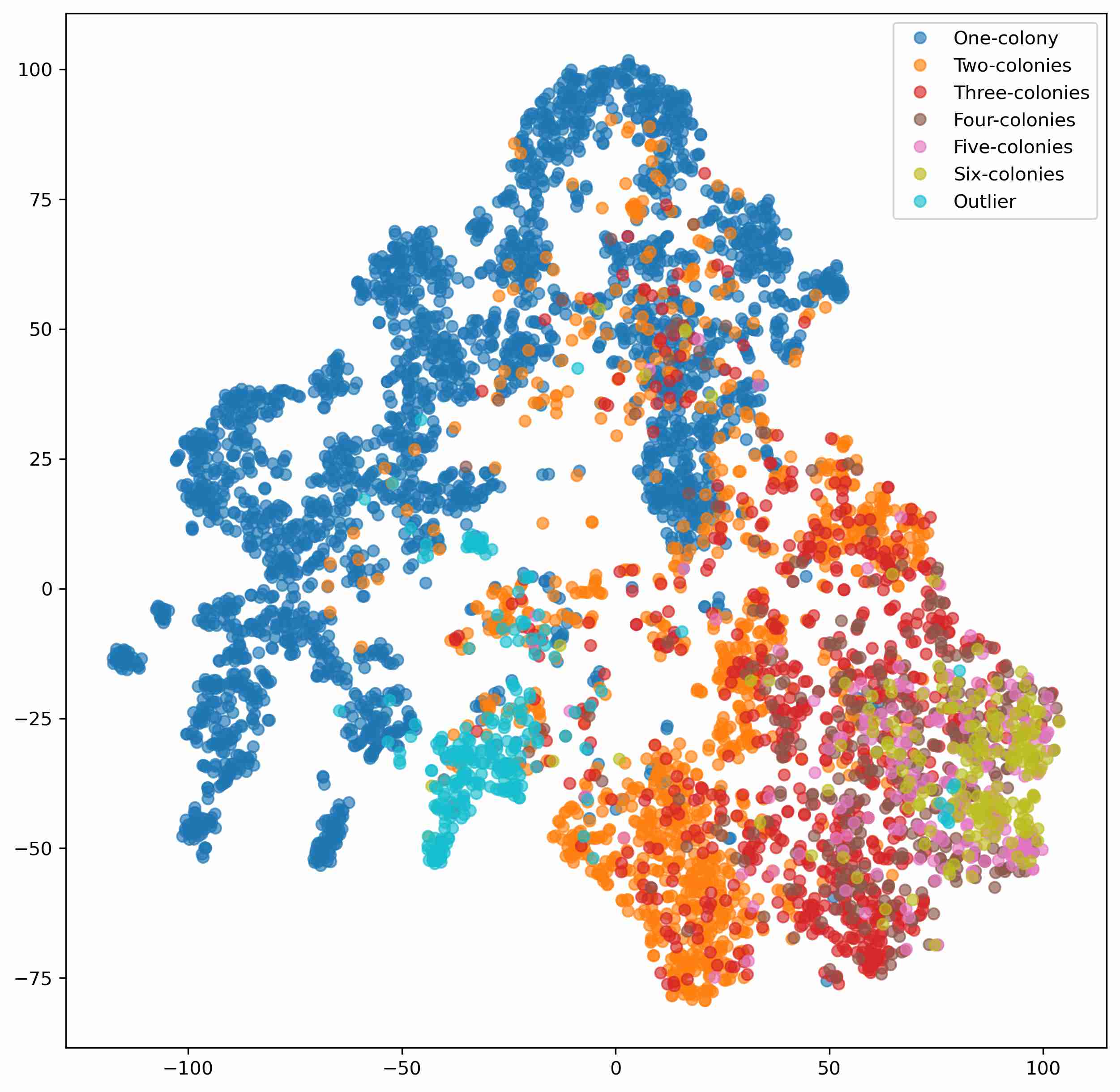}
\caption{Same as Figure \ref{fig:visualisation_of_network_layers_with_tsne_2_perplexity_training_val_v440}, but the perplexity value is 30.}
\label{fig:visualisation_of_network_layers_with_tsne_30_perplexity_training_val_v440}
\end{figure}

\begin{figure}[h!]
\centering
\includegraphics[width=.45\linewidth]{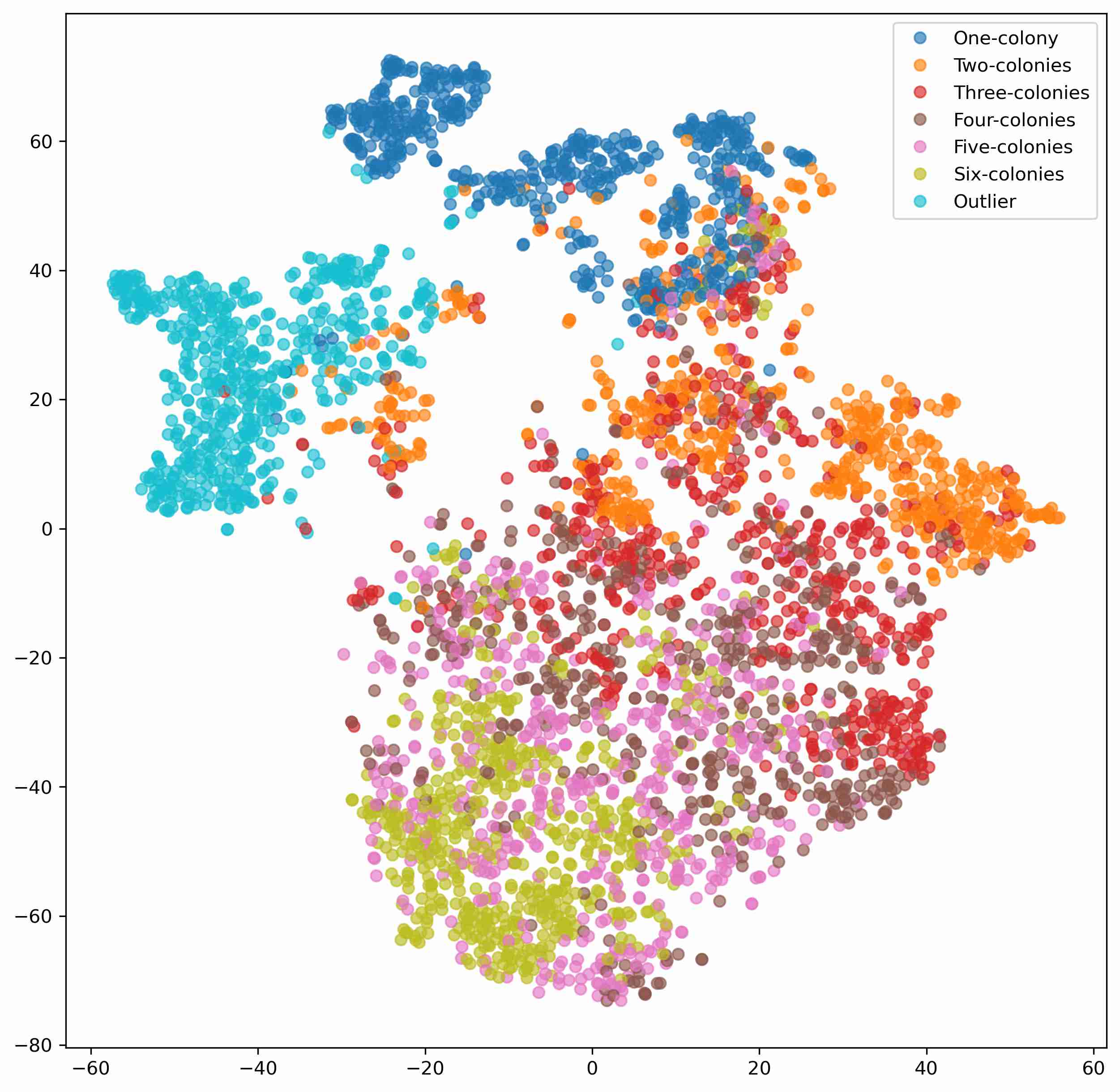}
\includegraphics[width=.45\linewidth]{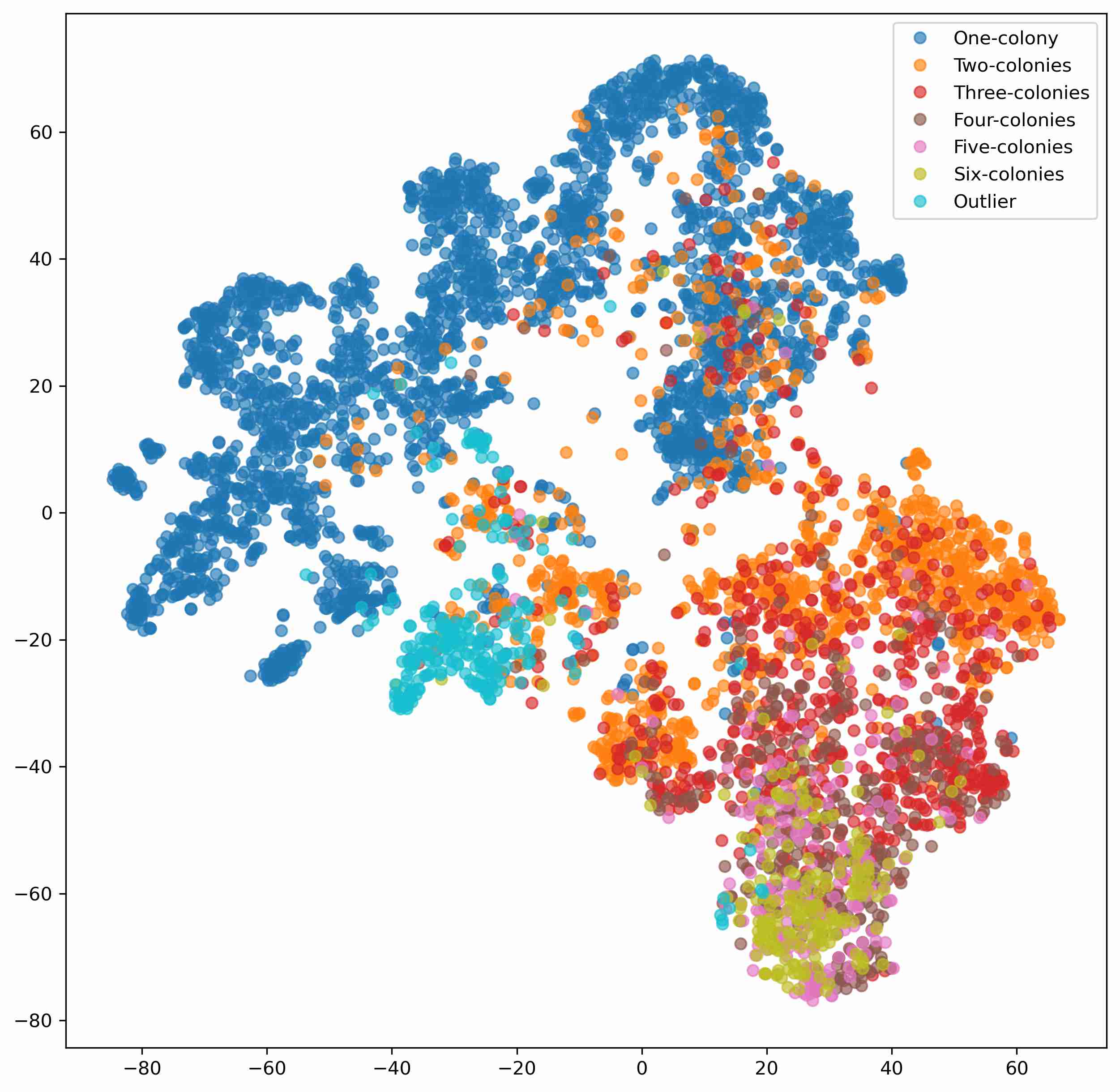}
\caption{Same as Figure \ref{fig:visualisation_of_network_layers_with_tsne_2_perplexity_training_val_v440}, but the perplexity value is 50.}
\label{fig:visualisation_of_network_layers_with_tsne_50_perplexity_training_val_v440}
\end{figure}

\begin{figure}[h!]
\centering
\includegraphics[width=.45\linewidth]{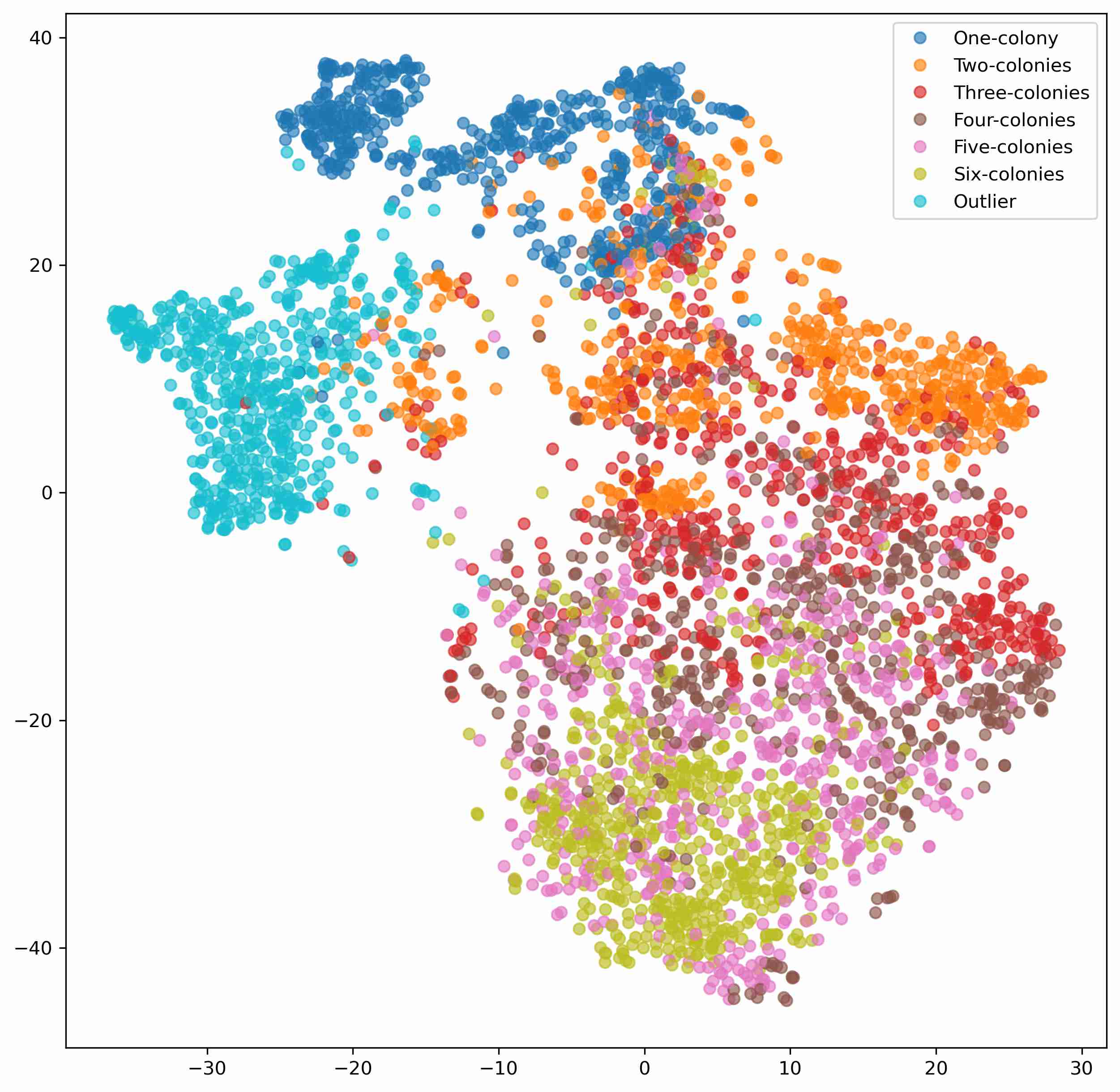}
\includegraphics[width=.45\linewidth]{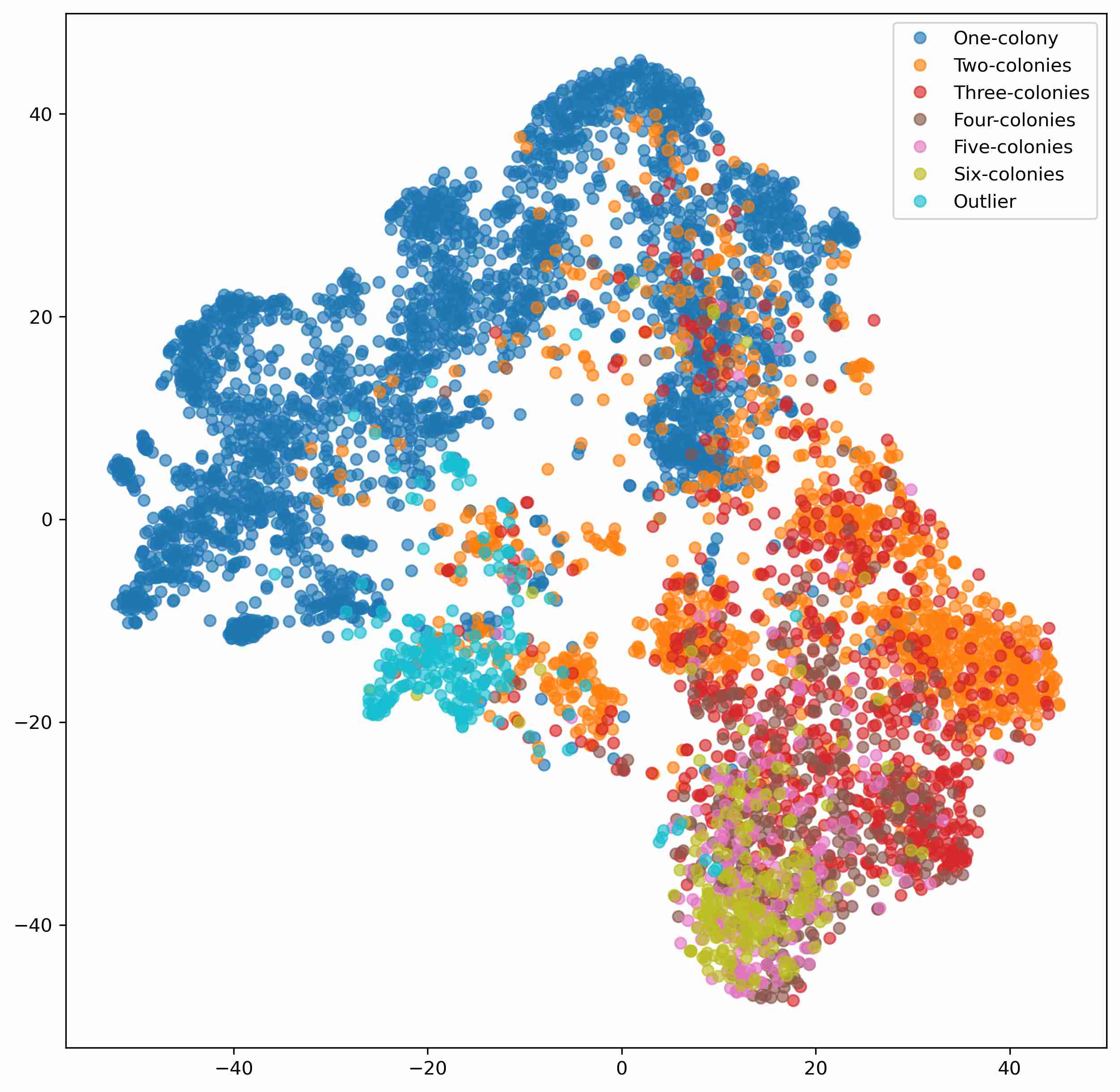}
\caption{Same as Figure \ref{fig:visualisation_of_network_layers_with_tsne_2_perplexity_training_val_v440}, but the perplexity value is 100.}
\label{fig:visualisation_of_network_layers_with_tsne_100_perplexity_training_val_v440}
\end{figure}

\clearpage

\bibliographystyle{elsarticle-harv} 
\bibliography{main.bib}

\end{document}